\documentclass{article}
\usepackage[margin=1in]{geometry}
\usepackage{graphicx}
\usepackage{subcaption}
\usepackage{booktabs}
\usepackage{hyperref}
\usepackage{amsmath}
\usepackage{authblk}
\hypersetup{colorlinks=true, linkcolor=blue, citecolor=blue}

\title{State of AI:\\An Empirical 100 Trillion Token Study with OpenRouter}

\author[$\dagger$]{Malika Aubakirova\footnote{\noindent $\ast$Lead contributors. Please see \textit{Contributions} section for details.}}
\author[$\ddagger$]{Alex Atallah}
\author[$\ddagger$]{Chris Clark}
\author[$\ddagger$]{Justin Summerville}
\author[$\dagger$]{Anjney Midha}

\affil[ ]{%
\textsuperscript{$\ddagger$}OpenRouter Inc. \quad \textsuperscript{$\dagger$}a16z (Andreessen Horowitz)
}

\date{December, 2025}

\begin{document}

\maketitle

\begin{abstract}
The past year has marked a turning point in the evolution and real-world use of large language models (LLMs). With the release of the first widely adopted reasoning model, \textit{o1}, on December 5th, 2024, the field shifted from single-pass pattern generation to multi-step deliberation inference, accelerating deployment, experimentation, and new classes of applications. As this shift unfolded at a rapid pace, our empirical understanding of how these models have actually been used in practice has lagged behind. In this work, we leverage the OpenRouter platform, which is an AI inference provider across a wide variety of LLMs, to analyze over 100 trillion tokens of real-world LLM interactions across tasks, geographies, and time. In our empirical study, we observe substantial adoption of open-weight models, the outsized popularity of creative roleplay (beyond just the productivity tasks many assume dominate) and coding assistance categories, plus the rise of agentic inference.  Furthermore, our retention analysis identifies \textit{foundational cohorts}: early users whose engagement persists far longer than later cohorts. We term this phenomenon the Cinderella \emph{``Glass Slipper"} effect. These findings underscore that the way developers and end-users engage with LLMs “in the wild” is complex and multifaceted. We discuss implications for model builders, AI developers, and infrastructure providers, and outline how a data-driven understanding of usage can inform better design and deployment of LLM systems.
\end{abstract}

\section{Introduction}
Just a year ago, the landscape of large language models looked fundamentally different. Prior to late 2024, state-of-the-art systems were dominated by single-pass, autoregressive predictors optimized to continue text sequences. Several precursor efforts attempted to approximate reasoning through advanced instruction following and tool use. For instance, \emph{Anthropic's Sonnet 2.1 \& 3} models excelled at sophisticated \emph{tool use and Retrieval-Augmented Generation (RAG)}, and \emph{Cohere's Command R} models incorporated structured tool-planning tokens. Separately, open source projects like those done by \emph{Reflection} explored supervised chain-of-thought and self-critique loops during training. Although these advanced techniques produced reasoning-like outputs and superior instruction following, the fundamental inference procedure remained based on a single forward pass, emitting a surface-level trace learned from data rather than performing iterative, internal computation.

This paradigm evolved on \textbf{December 5, 2024}, when OpenAI released the first full version of its \textit{o1} reasoning model (codenamed \textit{Strawberry}) \cite{openai2024o1}. The preview released on September 12, 2024 had already indicated a departure from conventional autoregressive inference. Unlike prior systems, \textit{o1} employed an expanded inference-time computation process involving internal multi-step deliberation, latent planning, and iterative refinement before generating a final output. Empirically, this enabled systematic improvements in mathematical reasoning, logical consistency, and multi-step decision-making, reflecting a shift from pattern completion to structured internal cognition. In retrospect, last year marked the field’s true inflection point: earlier approaches gestured toward reasoning, but \textit{o1} introduced the first generally-deployed architecture that performed reasoning through deliberate multi-stage computation rather than merely \textit{describing} it \cite{wei2022, yao2023}.

While recent advances in LLM capabilities have been widely documented, systematic evidence about how these models are actually used in practice remains limited \cite{zhao2024, chiang2024}. Existing accounts tend to emphasize qualitative demonstrations or benchmark performance rather than large-scale behavioral data. To bridge this gap, we undertake an empirical study of LLM usage, leveraging a 100 trillion token dataset from \textbf{OpenRouter}, a multi-model AI inference platform that serves as a hub for diverse LLM queries.

OpenRouter’s vantage point provides a unique window into fine-grained usage patterns. Because it orchestrates requests across a wide array of models (spanning both closed source APIs and open-weight deployments), OpenRouter captures a representative cross-section of how developers and end-users actually invoke language models for various tasks. By analyzing this rich dataset, we can observe which models are chosen for which tasks, how usage varies across geographic regions and over time, and how external factors like pricing or new model launches influence behavior. 

In this paper, we draw inspiration from prior empirical studies of AI adoption, including Anthropic’s economic impact and usage analyses \cite{anthropic2025} and OpenAI’s report \textit{How People Use ChatGPT} \cite{openai2025}, aiming for a neutral, evidence-driven discussion. We first describe our dataset and methodology, including how we categorize tasks and models. We then delve into a series of analyses that illuminate different facets of usage:

\begin{itemize}
    \item \textbf{Open vs. Closed Source Models:} We examine the adoption patterns of open source models relative to proprietary models, identifying trends and key players in the open source ecosystem.
    \item \textbf{Agentic Inference:} We investigate the emergence of multi-step, tool-assisted inference patterns, capturing how users increasingly employ models as components in larger automated systems rather than for single-turn interactions.
    \item \textbf{Category Taxonomy:} We break down usage by task category (such as programming, roleplay, translation, etc.), revealing which application domains drive the most activity and how these distributions differ by model provider.
    \item \textbf{Geography:} We analyze global usage patterns and compare LLM uptake across continents. This highlights how regional factors and local model offerings shape overall demand.
    \item \textbf{Effective Cost vs Usage Dynamics:} We assess how usage corresponds to effective costs, capturing the economic sensitivity of LLM adoption in practice. The metric is based on average input plus output tokens and accounts for caching effects.
    \item \textbf{Retention Patterns:} We analyze long-term retention for the most widely used models, identifying \textit{foundational cohorts} that define persistent, stickier behaviors. We define this to be a Cinderella \textit{``Glass Slipper" effect}, where early alignment between user needs and model characteristics creates a lasting fit that sustains engagement over time.
\end{itemize}

Finally, we discuss what these findings reveal about real-world LLM usage, highlighting unexpected patterns and correcting some myths. 

\section{Data and Methodology}
\label{sec:data}

\subsection{OpenRouter Platform and Dataset}
\label{subsec:openrouter_dataset}
Our analysis is based on metadata collected from the \textbf{OpenRouter} platform, a unified AI inference layer that connects users and developers to hundreds of large language models. Each user request on OpenRouter is executed against a user-selected model, and structured metadata describing the resulting “generation” event is logged. The dataset used in this study consists of \textbf{anonymized request-level metadata} for billions of prompt–completion pairs from a global user base, spanning approximately two years up to the time of writing. We do zoom in on the last year.

Crucially, we did not have access to the underlying text of prompts or completions. Our analysis relies entirely on \textbf{\emph{metadata}} that capture the structure, timing, and context of each \emph{generation}, without exposing user content. This privacy-preserving design enables large-scale behavioral analysis.

Each generation record includes information on timing, model and provider identifiers, token usage, and system performance metrics. Token counts encompass both prompt (input) and completion (output) tokens, allowing us to measure overall model workload and cost. Metadata also include fields related to geographic routing, latency, and usage context (for example, whether the request was streamed or cancelled, or whether tool-calling features were invoked). Together, these attributes provide a detailed but non-textual view of how models are used in practice.

All analyses, aggregations, and most visualizations based on this metadata were conducted using the \textbf{Hex} analytics platform, which provided a reproducible pipeline forfor versioned SQL queries, transformations, and final figure generation,

We emphasize that this dataset is \textbf{observational}: it reflects real-world activity on the OpenRouter platform, which itself is shaped by model availability, pricing, and user preferences. As of 2025, OpenRouter supports more than 300+ active models from over 60 providers and serves millions of developers and end-users, with over 50\% of usage originating outside the United States. While certain usage patterns outside the platform are not captured, OpenRouter’s global scale and diversity make it a representative lens on large-scale LLM usage dynamics.

\subsection{GoogleTagClassifier for Content Categorization}
\label{subsec:google_tagclassifier}

No direct access to user prompts or model outputs was available for this study. Instead, \textbf{OpenRouter performs internal categorization on a random sample comprising approximately 0.25\% of all prompts} and responses through a non-proprietary module \textbf{GoogleTagClassifier}. While this represents only a fraction of total activity, the underlying dataset remains substantial given the overall query volume processed by OpenRouter. 

To clarify on content categorization, this sampling is not performed automatically for every user, nor is it done without consent. OpenRouter provides an explicit opt-in mechanism. Users who agree to share anonymized prompt metadata for analytics receive discounted pricing, while those who prefer full privacy can keep their data entirely inaccessible for analytics. The classifier is embedded within the model inference pipeline and executes in an isolated service that receives only the minimal information required to produce a category label. All processing occurs transiently in memory. The classifier's outputs are restricted to aggregate counts for internal analytics and reporting.

The design aligns with OpenRouter's published Terms of Service, which explicitly state that classification is performed only on anonymous chunks and used solely for aggregate analytics based on opt-in mechanism. The system does not enable reconstruction of prompt content nor attribution of activity to specific users. 

GoogleTagClassifier interfaces with Google Cloud Natural Language’s \texttt{classifyText} content-classification API.\footnote{\url{https://cloud.google.com/natural-language/docs/classifying-text}}
 The API applies a hierarchical, language-agnostic taxonomy to textual input, returning one or more category paths (e.g., \texttt{/Computers \& Electronics/Programming}, \texttt{/Arts \& Entertainment/Roleplaying Games}) with corresponding confidence scores in the range \$[0,1]\$. The classifier operates directly on prompt data (up to the first 1{,}000 characters). The classifier is deployed within OpenRouter’s infrastructure, ensuring that classifications remain anonymous and are not linked to individual customers. Categories with confidence scores below the default threshold of 0.5 are excluded from further analysis. The classification system itself operates entirely within OpenRouter’s infrastructure and was not part of this study; our analysis relied solely on the resulting categorical outputs (effectively metadata describing prompt classifications) rather than the underlying prompt content.

To make these fine-grained labels useful at scale, we map GoogleTagClassifier’s taxonomy to a compact set of study-defined buckets and assign each request \emph{tags}. Each tag rolls up to higher level \emph{category} in one to one way.  Representative mappings include:
\begin{itemize}
    \item \textbf{Programming}: from \texttt{/Computers \& Electronics/Programming} or \texttt{/Science/Computer Science/*}
    \item \textbf{Roleplay}: from \texttt{/Games/Roleplaying Games}, \texttt{/Adult}, and creative dialogue leaves under \texttt{/Arts \& Entertainment/*}
    \item \textbf{Translation}: from \texttt{/Reference/Language Resources/*}
    \item \textbf{General Q\&A / Knowledge}: from \texttt{/Reference/General Reference/*} and \texttt{/News/*} when the intent appears to be factual lookup
    \item \textbf{Productivity/Writing}: from \texttt{/Computers \& Electronics/Software/Business \& Productivity Software} or \texttt{/Business \& Industrial/Business Services/Writing \& Editing Services}
    \item \textbf{Education}: from \texttt{/Jobs \& Education/Education/*}
    \item \textbf{Literature/Creative Writing}: from \texttt{/Books \& Literature/*} and narrative leaves under \texttt{/Arts \& Entertainment/*}
    \item \textbf{Others}: for the long tail of prompts when no dominant mapping applies. (Note: we omit this category from most analyses below.)
\end{itemize}

There are inherent limitations to this approach, for instance, reliance on a predefined taxonomy constrains how novel or cross-domain behaviors are categorized, and certain interaction types may not yet fit neatly within existing classes. In practice, some prompts receive multiple category labels when their content spans overlapping domains. Nonetheless, the classifier-driven categorization provides us with a lens for downstream analyses (Section~\ref{sec:categories}). This enables us to quantify not just \emph{how much} LLMs are used but \emph{what for}.

\subsection{Model and Token Variants}
A few variants are worth explicitly calling out:
\begin{itemize}
    \item \textit{Open Source vs. Proprietary:} We label models as \textbf{open source (OSS, for simplicity)} if their weights are publicly available, and \textbf{closed source} if access is only via a restricted API (e.g., Anthropic’s Claude). This distinction lets us measure adoption of community-driven models versus proprietary ones.
    \item \textit{Origin (Chinese vs. Rest-of-World):} Given the rise of Chinese LLMs and their distinct ecosystems, we tag models by primary locale of development. \textbf{Chinese models} include those developed by organizations in China, Taiwan, or Hong Kong (e.g., Alibaba’s Qwen, Moonshot AI's Kimi, or DeepSeek). \textbf{RoW (Rest-of-World) models} cover North America, Europe, and other regions. 
       \item \textit{Prompt vs. Completion Tokens:} We distinguish between \textbf{prompt tokens}, which represent the input text provided to a model, and \textbf{completion tokens}, which represent the model’s generated output. \textbf{Total tokens} equal the sum of prompt and completion tokens. \textbf{Reasoning tokens} represent internal reasoning steps in models with native reasoning capabilities and are included within \textbf{completion tokens}.
\end{itemize}
Unless otherwise noted, \textbf{token volume} refers to \textbf{the sum of prompt (input) and completion (output) tokens}.

\subsection{Geographic Segmentation}
\label{subsec:geography}
To understand regional patterns in LLM usage, we segment requests by user geography. Direct request metadata (like IP-based location) is typically imprecise or anonymized. Instead, we determine user region based on the \textbf{billing location} associated with each account. This provides a more reliable proxy for user geography, as billing data reflects the country or region linked to the user’s payment method or account registration. We use this billing-based segmentation in our analysis of regional adoption and model preferences (Section~\ref{sec:geography}).

This method has limitations. Some users employ third-party billing or shared organizational accounts, which may not correspond to their actual location. Enterprise accounts may aggregate activity across multiple regions under one billing entity. Despite these imperfections, billing geography remains the most stable and interpretable indicator available for privacy-preserving geographic analysis given the metadata we had access to.

\subsection{Time Frame and Coverage}
\label{subsec:timeframe}
Our analyses primarily cover a rolling 13-month period ending on November, 2025, but not all underlying metadata spans this full window. Most model-level and pricing analyses were focused on November 3, 2024 – November 30, 2025 time frame. However, category-level analyses (especially those using the GoogleTagClassifier taxonomy, Section~\ref{subsec:google_tagclassifier}) are based on a shorter interval beginning in May 2025, reflecting when consistent tagging became available on OpenRouter. In particular, detailed task classification fields (e.g., tags such as \textit{Programming}, \textit{Roleplay}, or \textit{Technology}) were only added in mid-2025. Consequently, all findings in Section~\ref{sec:categories} should be interpreted as representative of mid-2025 usage rather than the entire prior year.

Unless otherwise specified, all time-series aggregates are computed on a weekly basis using UTC-normalized timestamps, summing prompt and completion tokens. This approach ensures comparability across model families and minimizes bias from transient spikes or regional time-zone effects.

\section{Open vs. Closed Source Models}
\label{sec:oss_vs_closed}

\begin{figure}[htbp]
    \centering
    \includegraphics[width=1\linewidth]{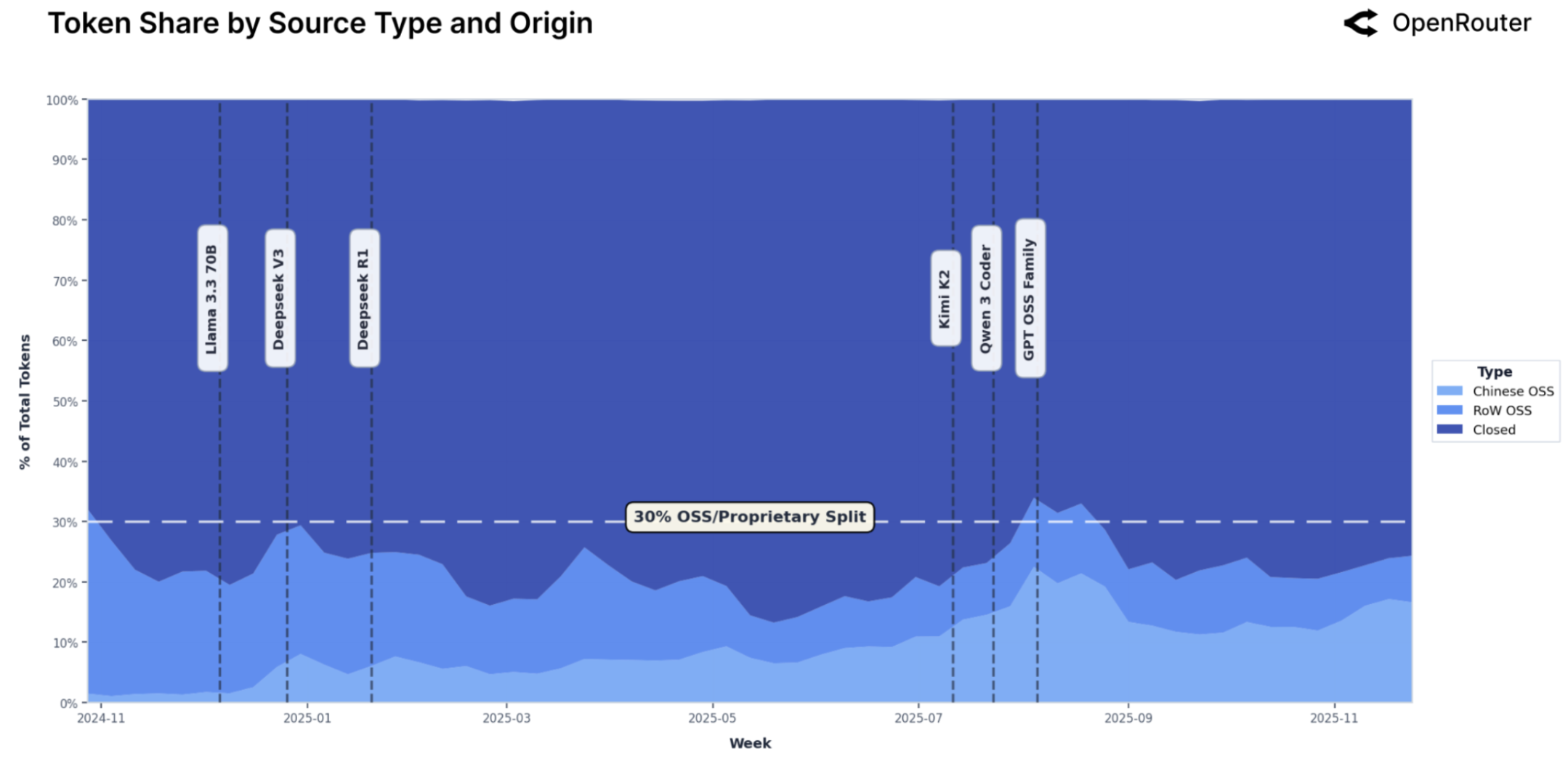}
    \caption{\textbf{Open vs closed source models split.} Weekly share of total token volume by source type. Lighter blue shades represent open-weight models (China vs Rest-of-World), while dark blue corresponds to proprietary (closed) offerings. Vertical dashed lines mark the release of key open weight models including Llama~3.3~70B, DeepSeek~V3, DeepSeek~R1, Kimi K2, GPT OSS family, and Qwen~3~Coder.}
    \label{fig:oss_split}
\end{figure}

\begin{figure}[htbp]
    \centering
    \includegraphics[width=1\linewidth]{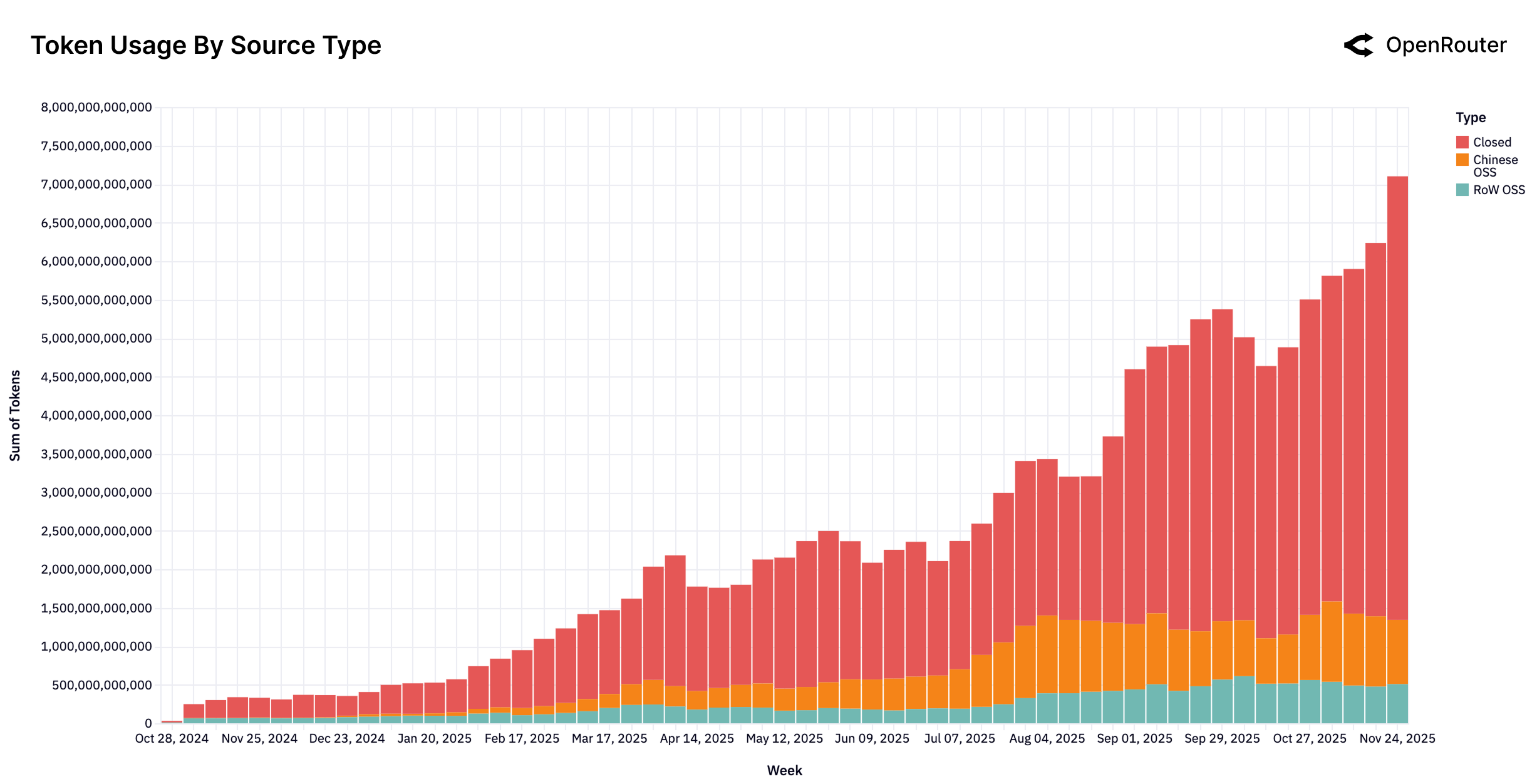}
    \caption{\textbf{Weekly token volume by model type.} Stacked bar chart showing total token usage by model category over time. Dark red corresponds to proprietary models (\emph{Closed}), orange represents Chinese open source models (\emph{Chinese OSS}), and teal indicates open source models developed outside China (\emph{RoW OSS}). The chart highlights a gradual increase in OSS token share through 2025, particularly among Chinese OSS models beginning in mid-year.}
    \label{fig:oss_token_share}
\end{figure}

A central question in the AI ecosystem is the balance between open-weight (that we abbreviate to OSS for simplicity) and proprietary models. Figures~\ref{fig:oss_split} and~\ref{fig:oss_token_share} illustrate how this balance has evolved on OpenRouter over the past year. While proprietary models, especially those from major North American providers, still serve the majority of tokens, OSS models have grown steadily, reaching approximately one-third of usage by late 2025.

This expansion is not incidental. Usage spikes align with major open-model releases such as DeepSeek~V3 and Kimi K2 (indicated by vertical dashed lines), suggesting that competitive OSS launches such as DeepSeek V3 \cite{deepseek2024} and GPT OSS models \cite{dubey2024} are adopted rapidly and sustain their gains. Importantly, these increases persist beyond initial release weeks, implying genuine production use rather than short-term experimentation.

A significant share of this growth has come from \textbf{Chinese-developed models}. Starting from a negligible base in late 2024 (weekly share as low as 1.2\%), Chinese OSS models steadily gained traction, reaching nearly 30\% of total usage among all models in some weeks. Over the one-year window, they averaged approximately 13.0\% of weekly token volume, with strong growth concentrated in the second half of 2025. For comparison, RoW OSS models averaged 13.7\%, while proprietary RoW models retained the largest share (70\% on average). The expansion of Chinese OSS reflects not only competitive quality, but also rapid iteration and dense release cycles. Models like Qwen and DeepSeek maintained regular model releases that enabled fast adaptation to emerging workloads. This pattern has materially reshaped the open source segment and progressed global competition across the LLM landscape.

These trends indicate a durable dual structure in the LLM ecosystem. Proprietary systems continue to define the upper bound of reliability and performance, particularly for regulated or enterprise workloads. OSS models, by contrast, offer cost efficiency, transparency, and customization, making them an attractive option for certain workloads. \textbf{The equilibrium is currently reached at roughly 30\%.} These models are not mutually exclusive; rather, they complement each other within a multi-model stack that developers and infrastructure providers increasingly favor.

\subsection{Key Open Source Players}
\label{subsec:oss_players}

Table~\ref{tab:top_oss_model_authors} ranks the top model families in our dataset by total token volume served. The landscape of OSS models has shifted significantly over the last year: while DeepSeek remains the single largest OSS contributor by volume, its dominance has waned as new entrants rapidly gain ground. Today, multiple open source families each sustain substantial usage, pointing to a diversified ecosystem.

\begin{table}[htbp]
\centering
\caption{\textbf{Total token volume by model author (Nov 2024–Nov 2025).} Token counts reflect aggregate usage across all model variants on OpenRouter.}
\label{tab:top_oss_model_authors}
\begin{tabular}{l r}
\toprule
\textbf{Model Author} & \textbf{Total Tokens (Trillions)} \\
\midrule
DeepSeek & 14.37 \\
Qwen & 5.59 \\
Meta LLaMA & 3.96 \\
Mistral AI & 2.92 \\
OpenAI & 1.65 \\
Minimax & 1.26 \\
Z-AI & 1.18 \\
TNGTech & 1.13 \\
MoonshotAI & 0.92 \\
Google & 0.82 \\
\bottomrule
\end{tabular}
\end{table}

\begin{figure}[htbp]
\centering
\includegraphics[width=\textwidth]{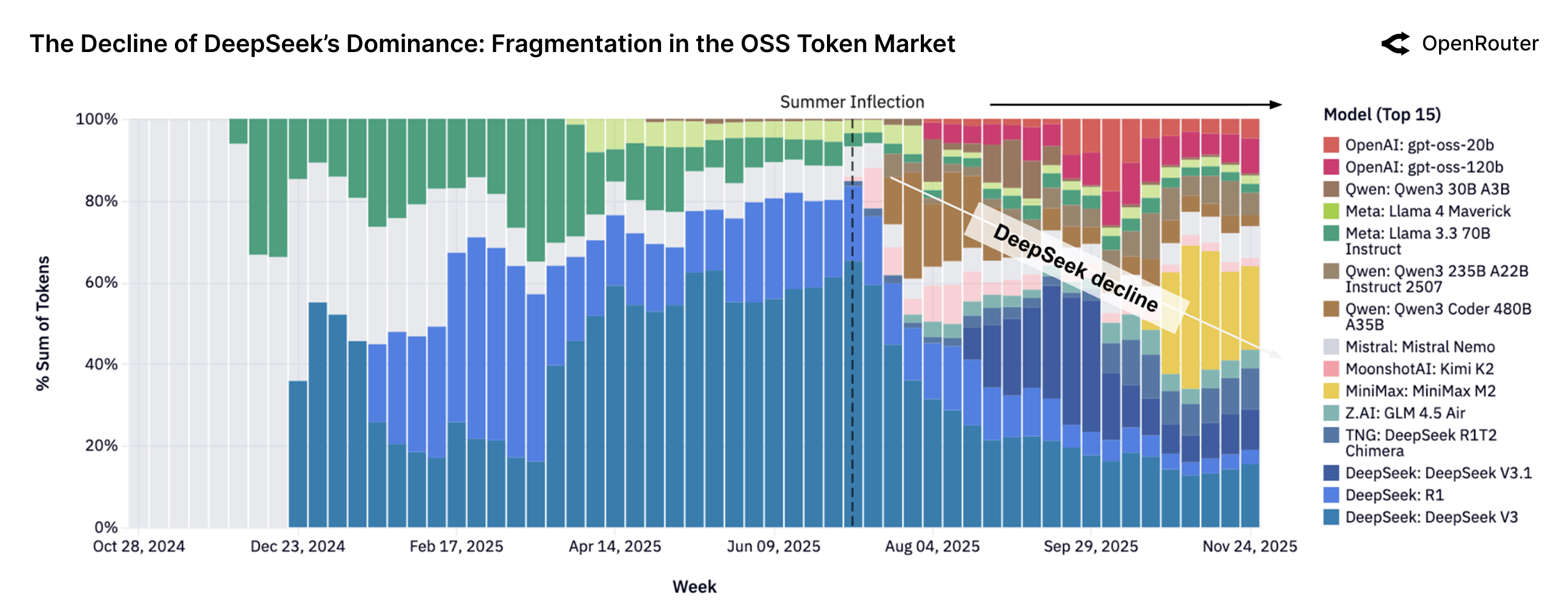}
\caption{\textbf{Top 15 OSS models over time.} Weekly relative token share for leading open source models (stacked area chart). Each colored band represents one model’s contribution to total OSS tokens. The broadening palette over time indicates a more competitive distribution without a single dominant model in recent months.}
\label{fig:top_15_oss_models_over_time}
\end{figure}

Figure~\ref{fig:top_15_oss_models_over_time} illustrates the dramatic evolution of market share among the top individual open source models week by week. Early in the period (late 2024), the market was highly consolidated: two models from the DeepSeek family (V3 and R1) consistently accounted for over half of all OSS token usage, forming the large, dark blue bands at the bottom of the chart.

This near-monopoly structure shattered following the Summer Inflection (mid-2025). The market has since become both broader and deeper, with usage diversifying significantly. New entrants like Qwen's models, Minimax's M2, MoonshotAI's Kimi K2, and OpenAI’s GPT-OSS series all grew rapidly to serve significant portions of requests, often achieving production-scale adoption within weeks of release. This signals that the open source community and AI startups can achieve quick adoption by introducing models with novel capabilities or superior efficiency.

By late 2025, the competitive balance had shifted from near-monopoly to a pluralistic mix. No single model exceeds 25\% of OSS tokens, and the token share is now distributed more evenly across five to seven models. The practical implication is that users are finding value in a wider array of options, rather than defaulting to one ``best" choice. Although this figure visualizes relative share among OSS models (not absolute volume), the clear trend is a decisive shift toward market fragmentation and increased competition within the open source ecosystem.

\paragraph{Overall, the open source model ecosystem is now highly dynamic.} Key insights include:
\begin{itemize}
    \item \textbf{Top-tier diversity:} Where one family (DeepSeek) once dominated OSS usage, we now increasingly see half a dozen models each sustaining meaningful share. No single open model holds more than $\approx$20--25\% of OSS tokens consistently.
    \item \textbf{Rapid scaling of new entrants:} Capable new open models can capture significant usage within weeks. For example, MoonshotAI’s models quickly grew to rival older OSS leaders, and even a newcomer like MiniMax went from zero to substantial traffic in a single quarter. This indicates low switching friction and a user base eager to experiment.
    \item \textbf{Iterative advantage:} The longevity of DeepSeek’s presence at the top underscores that continuous improvement is critical. DeepSeek’s successive releases (Chat-V3, R1, etc.) kept it competitive even as challengers emerged. OSS models that stagnate in development tend to lose share to those with frequent updates at the frontier or domain-specific finetunes.
\end{itemize}

Today the open source LLM arena in 2025 resembles a competitive ecosystem where innovation cycles are rapid and leadership is not guaranteed. For model builders, this means that releasing an open model with state-of-the-art performance can yield immediate uptake, but maintaining usage share requires ongoing investment in further development. For users and application developers, the trend is positive: there is a richer selection of open models to choose from, often with comparable or sometimes superior capabilities to proprietary systems in specific areas (like roleplay). 

\subsection{The Model Size vs. Market Fit: Medium is the New Small}

\begin{figure}[htbp]
    \centering
    \includegraphics[width=\linewidth]{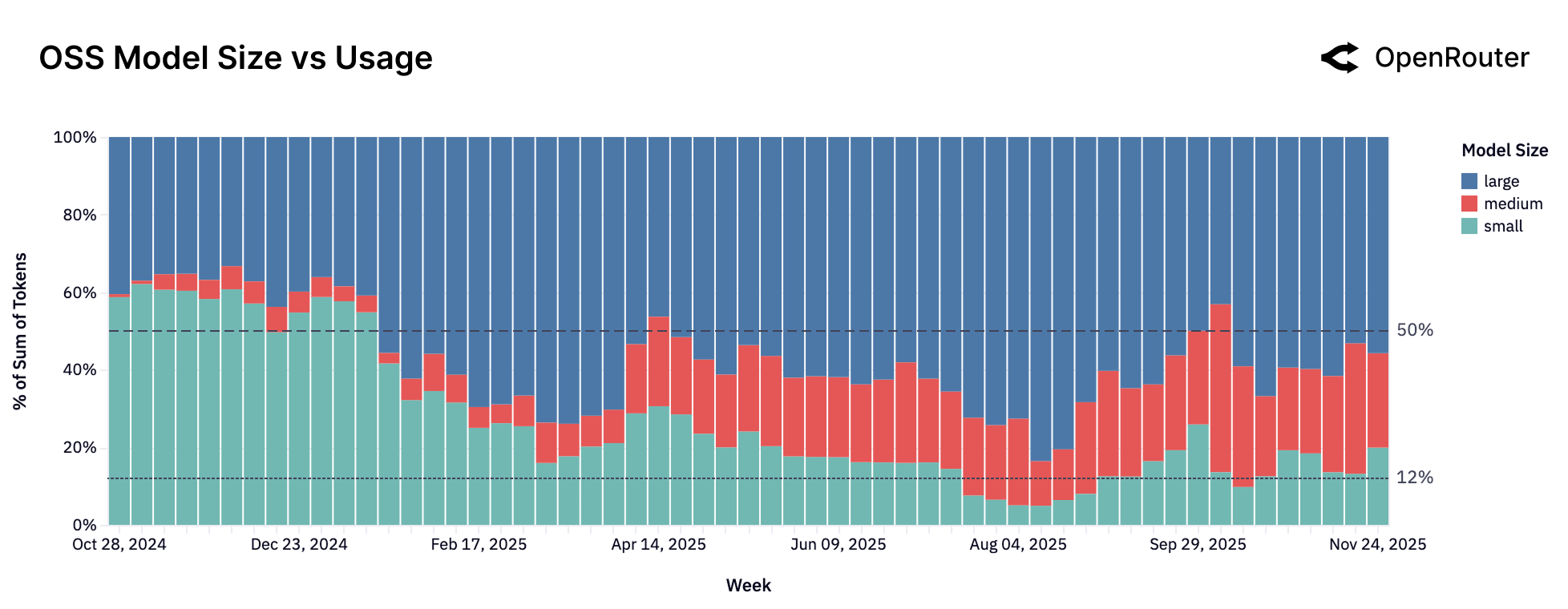}
    \caption{\textbf{OSS model size vs. usage.} Weekly share of total OSS token volume served by small, medium, and large models. Percentages are normalized by total OSS usage per week.}
    \label{fig:param_size_vs_usage}
\end{figure}

A year ago, the open source model ecosystem was largely a story of trade-offs between two extremes: a vast number of small, fast models and a handful of powerful, large-scale models. However, a review of the past year reveals a significant maturation of the market and the emergence of a new, growing category: the medium-sized model. Please note that we categorize models by their parameter count as follows:
\begin{itemize}
    \item \textbf{Small:} Models with fewer than 15 billion parameters.
    \item \textbf{Medium:} Models with 15 billion to 70 billion parameters.
    \item \textbf{Large:} Models with 70 billion or more parameters.
\end{itemize}

\begin{figure}[htbp]
    \centering
    \includegraphics[width=\linewidth]{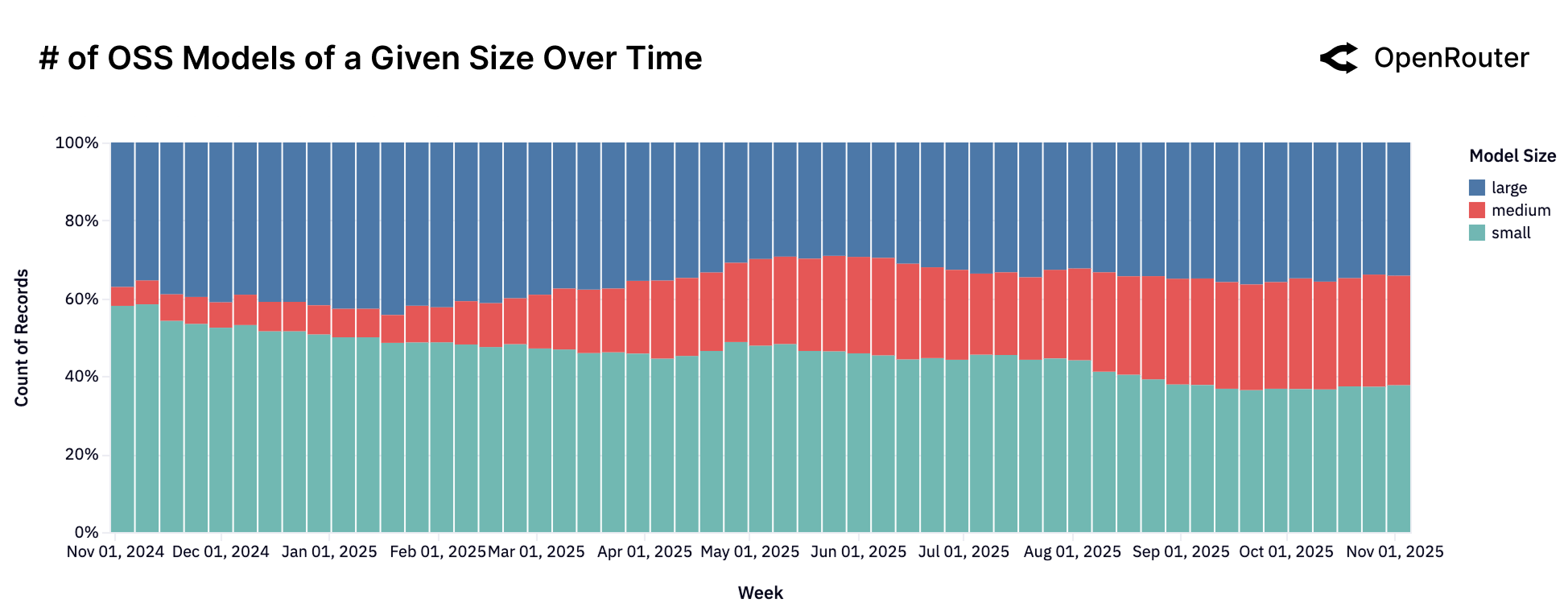}
    \caption{\textbf{Number of OSS models by size over time.} Weekly counts of available open source models, grouped by parameter size category.}
    \label{fig:param_size_vs_num_available}
\end{figure}

The data on developer and user behavior tells us a nuanced story. Figures \ref{fig:param_size_vs_usage} and \ref{fig:param_size_vs_num_available} show that while the \textit{number} of models across all categories has grown, the \textit{usage} has shifted notably. Small models are losing favor while medium and large models are capturing that value.

A deeper look at the models driving these trends reveals distinct market dynamics:

\begin{itemize}
    \item \textbf{The ``Small" Market: Overall Decline in Usage.} Despite a steady supply of new models, the small model category as a whole is seeing its share of usage decline, as seen in Figure \ref{fig:param_size_vs_usage}. This category is characterized by high fragmentation. No single model holds a dominant position for long, and it sees a constant churn of new entrants from a diverse set of providers like Meta, Google, Mistral, and DeepSeek. For example, \texttt{Google Gemma 3.12B} (released August 2025) saw a rapid adoption but competes in a crowded field where users continually seek the next best alternative.
    
    \item \textbf{The ``Medium" Market: Finding ``Model-Market Fit."} The medium model category tells a clear story of market creation. The segment itself was negligible until the release of \texttt{Qwen2.5 Coder 32B} in November 2024, which effectively established this category. This segment then matured into a competitive ecosystem with the arrival of other strong contenders like \texttt{Mistral Small 3} (January 2025) and \texttt{GPT-OSS 20B} (August 2025), which carved out user mind share. This segment demonstrates that users are seeking a balance of capability and efficiency.

    \item \textbf{The ``Large" Model Segment: A Pluralistic Landscape.} The ``flight to quality" has not led to consolidation but to diversification. The large model category now features a range of high-performing contenders from \texttt{Qwen3 235B A22B Instruct} (released in July 2025) and \texttt{Z.AI GLM 4.5 Air} to \texttt{OpenAI: GPT-OSS-120B} (August 5th): each capturing meaningful and sustained usage. This pluralism suggests users are actively benchmarking across multiple open large models rather than converging on a single standard.
\end{itemize}

The era of small models dominating the open source ecosystem might be behind. The market is now bifurcating, with users either gravitating toward a new, robust class of medium models, or consolidating their workloads onto the single most capable large model.

Please note that the overall decline in small model usage here reflects OpenRouter’s vantage point as an API service, not the entire ecosystem. Small models are precisely the ones most commonly self-hosted, which means a meaningful portion of their real usage is invisible to any centralized API provider. It is entirely possible that total small model usage has grown overall even as their share within OpenRouter declines. Our findings should therefore be interpreted as platform-level observations rather than ecosystem-wide conclusions.

\subsection{What Are Open Source Models Used For?}
\label{sec:oss_categories}

\begin{figure}[htbp]
    \centering
    \includegraphics[width=\linewidth]{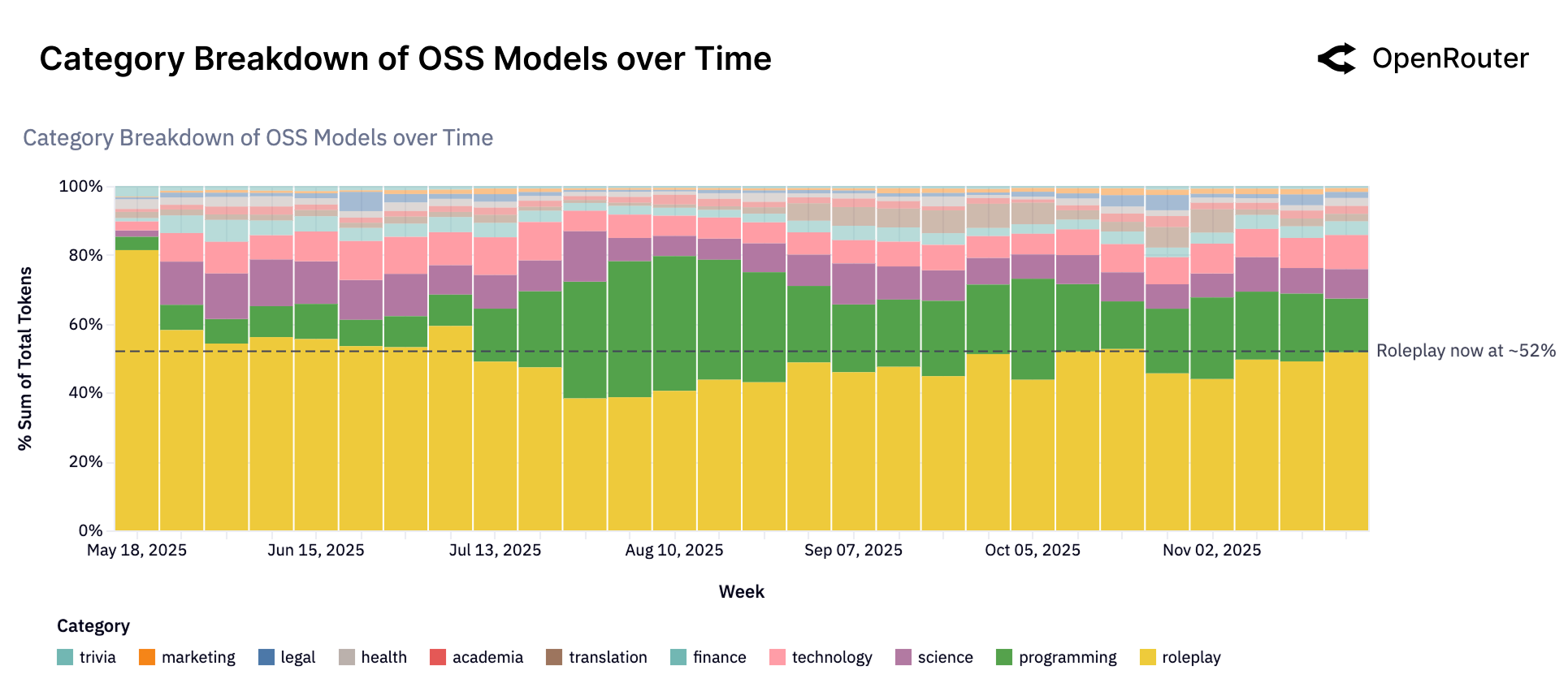}
    \caption{\textbf{Category Trends of OSS Models.} Distribution of open source model usage across high-level task categories. Roleplay (about 52\%) and programming consistently dominate the OSS workload mix, together accounting for the majority of OSS tokens. Smaller segments include translation, general knowledge Q\&A, and others.}
    \label{fig:oss_category_trends}
\end{figure}

Open source models today are employed for a remarkably broad range of tasks, spanning creative, technical, and informational domains. While proprietary models still dominate in structured business tasks, OSS models have carved out leadership in two particular areas: \textbf{creative roleplay} and \textbf{programming assistance}. Together, these categories account for the majority of OSS token usage (Figure~\ref{fig:oss_category_trends}).

\textbf{Figure~\ref{fig:oss_category_trends}} highlights that more than half of all OSS model usage falls under \emph{Roleplay}, with \emph{Programming} being the second-largest category. This indicates that users turn to open models primarily for creative interactive dialogues (such as storytelling, character roleplay, and gaming scenarios) and for coding-related tasks. The dominance of roleplay (hovering at more than 50\% of all OSS tokens) underscores a use case where open models have an edge: they can be utilized for creativity and are often less constrained by content filters, making them attractive for fantasy or entertainment applications. Roleplay tasks require flexible responses, context retention, and emotional nuance - attributes that open models can deliver effectively without being heavily restricted by commercial safety or moderation layers. This makes them particularly appealing for communities experimenting with character-driven experiences, fan fiction, interactive games, and simulation environments.
\begin{figure}[htbp]
    \centering
    \includegraphics[width=\linewidth]{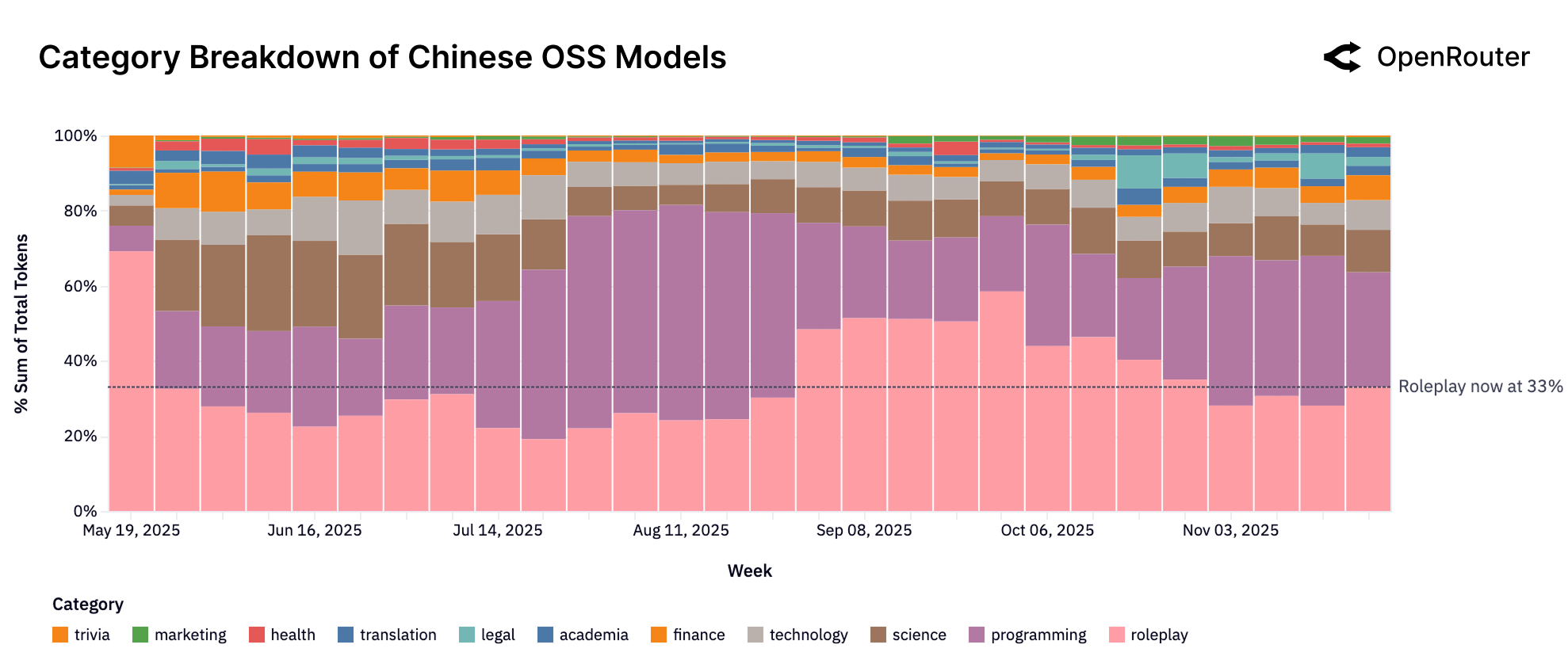}
    \caption{\textbf{Chinese OSS Category Trends.} Category composition among open source models developed in China. Roleplay remains the single largest use case, though programming and technology collectively make up a larger fraction here than in the overall OSS mix (33\% compared to 38\%).}
    \label{fig:chinese_oss_category_trends}
\end{figure}

The programming slice (roughly 15-20\%) shows that many developers leverage OSS models for code generation and debugging, likely due to very capable code models like \texttt{Qwen-Coder}, \texttt{GPT-OSS} family, and \texttt{GLM-4.6}. Other categories like \textit{Translation}, \textit{Knowledge Q\&A}, and \textit{Education} occupy smaller shares but are non-negligible, each catering to specific needs (multilingual support, factual lookups, tutoring, etc.). One limitation is that the classification might conflate some overlapping uses (e.g., an interactive coding tutorial could be tagged as education or programming depending on prompt framing) but overall the chart gives a clear indication of where OSS models excel in practice.

\textbf{Figure~\ref{fig:chinese_oss_category_trends}} shows category breakdown over time if we zoom in on Chinese OSS models only. These models are no longer used primarily for creative tasks. Roleplay remains the largest category at around 33\%, but programming and technology now account for a combined majority of usage (39\%). This shift suggests that models like \texttt{Qwen} and \texttt{DeepSeek} are increasingly used for code generation and infrastructure-related workloads. While high-volume enterprise users may influence specific segments, the overall trend points to Chinese OSS models competing directly in technical and productivity domains.

\begin{figure}[htbp]
    \centering
    \includegraphics[width=\linewidth]{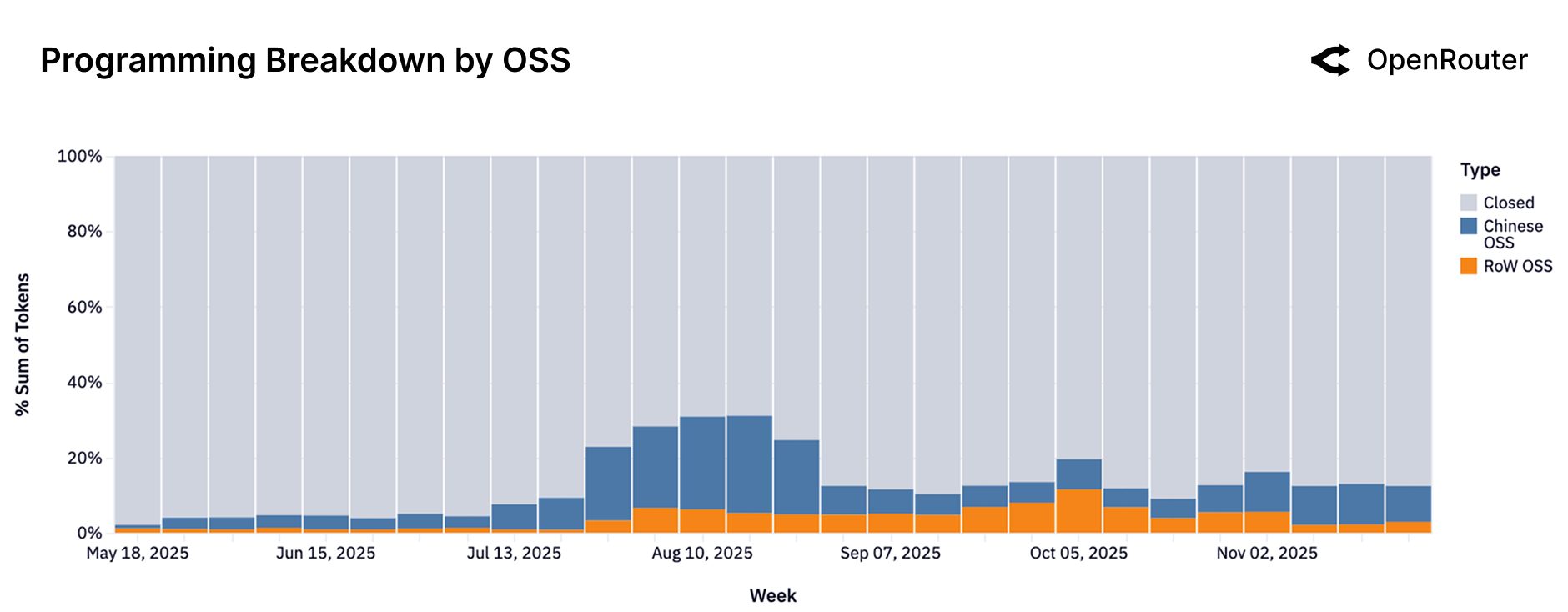}
    \caption{\textbf{Programming Queries by Model Source.} Share of programming-related token volume handled by proprietary models vs. Chinese OSS vs. non-Chinese (RoW) OSS models. Within the OSS segment, the balance shifted markedly toward RoW OSS in late 2025, which now accounts for over half of all open source coding tokens (after an earlier period where Chinese OSS dominated OSS coding usage).}
    \label{fig:programming_oss_split}
\end{figure}

If we zoom in just on the programming category,  we observe in \textbf{Figure~\ref{fig:programming_oss_split}} that proprietary models still handle the bulk of coding assistance overall (the gray region), reflecting strong offerings like Anthropic’s Claude. However, within the OSS portion, there was a notable transition: in mid-2025, Chinese OSS models (blue) delivered the majority of open source coding help (driven by early successes like \texttt{Qwen 3 Coder}). By Q4 2025, Western OSS models (orange) such as Meta’s LLaMA-2 Code and OpenAI’s GPT-OSS series had surged, but decreased in overall share in recent weeks. This oscillation suggests a very competitive environment. The practical takeaway is that open source code assistant usage is dynamic and highly responsive to new model quality: developers are open to whichever OSS model currently provides the best coding support. As a limitation, this figure doesn’t show absolute volumes: open source coding usage grew overall so a shrinking blue band doesn’t mean Chinese OSS lost users, only relative share.

\begin{figure}[htbp]
    \centering
    \includegraphics[width=\linewidth]{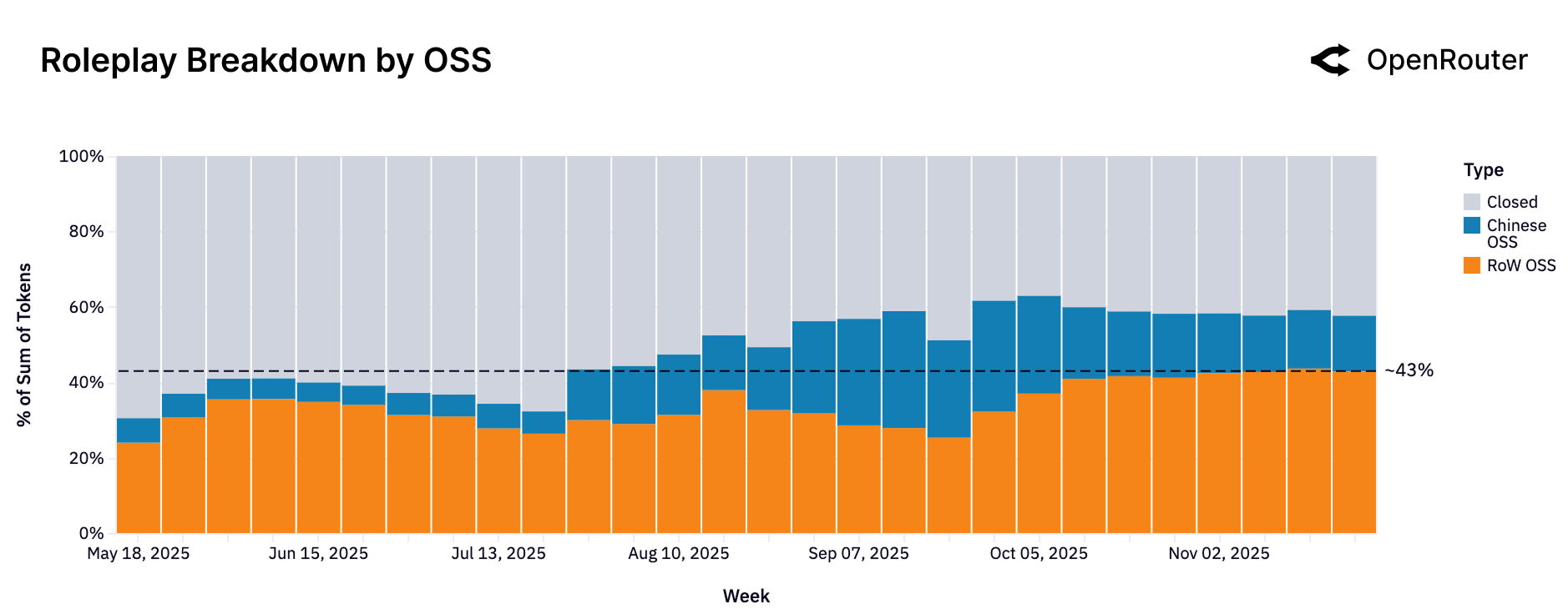}
    \caption{\textbf{Roleplay Queries by Model Source.} Token volume for roleplay use cases, split between Chinese OSS and RoW OSS models. Roleplay remains the largest category for both groups; by late 2025, traffic is roughly evenly divided between Chinese and non-Chinese open models.}
    \label{fig:roleplay_oss_split}
\end{figure}

Now if we examine just the roleplay traffic, we see in Figure~\ref{fig:roleplay_oss_split} that it is now almost equally served by Rest-of-World OSS (orange, 43\% in recent weeks) and Closed (gray, at ~42\% most recently) models. This represents a significant shift from earlier in 2025, when the category was dominated by proprietary (gray) models, which held approximately 70\% of the token share. At that time (May 2025), Western OSS models accounted for only ~22\% of traffic, and Chinese OSS (blue) models held a small share of ~8\%. Throughout the year, the proprietary share steadily eroded. By the end of October 2025, this trend accelerated as both Western and Chinese open source models gained significant ground.

The resulting convergence indicates a healthy competition; users have viable choices from both open and proprietary offerings for creative chats and storytelling. This reflects that developers recognize the demand for roleplay/chat models and have tailored their releases to that end (e.g., fine-tuning on dialogues, adding alignment for character consistency). A point to note is that “roleplay” covers a range of subgenres (from casual chatting to complex game scenarios). Yet from a macro perspective, it is clear OSS models have an edge in this creative arena.

\paragraph{Interpretation.} Broadly, across the OSS ecosystem, the key use cases are: \textbf{Roleplay and creative dialogue:} the top category, likely because open models can be uncensored or more easily customized for fictional persona and story tasks. \textbf{Programming assistance:} second-largest, and growing, as open models become more competent at code. Many developers leverage OSS models locally for coding to avoid API costs. \textbf{Translation and multilingual support:} a steady use case, especially with strong bilingual models available (Chinese OSS models have an edge here). \textbf{General knowledge Q\&A and education:} moderate usage; while open models can answer questions, users may prefer closed models like GPT-5 for highest factual accuracy.

It is worth noting that the OSS usage pattern (heavy on roleplay) mirrors what many might consider for “enthusiasts” or "indie developers" - areas where customization and cost-efficiency trump absolute accuracy. The lines are blurring, though: OSS models are rapidly improving in technical domains, and proprietary models are being used creatively too.

\section{The Rise of Agentic Inference}
\label{sec:agentic-inference}

Building on the previous section’s view of the evolving model landscape (open vs closed source), we now turn to the fundamental \textit{shape} of LLM usage itself. A foundational shift is underway in how language models are used in production: from single-turn text completion toward multi-step, tool-integrated, and reasoning-intensive workflows. We refer to this shift as the rise of \textbf{agentic inference}, where models are deployed not just to generate text, but to act through planning, calling tools, or interacting across extended contexts. This section traces that shift through five proxies: the rise of reasoning models, the expansion of tool-calling behavior, the changing sequence length profile, and how programming use drives complexity.

\begin{figure}[htbp]
    \centering
    \includegraphics[width=\linewidth]{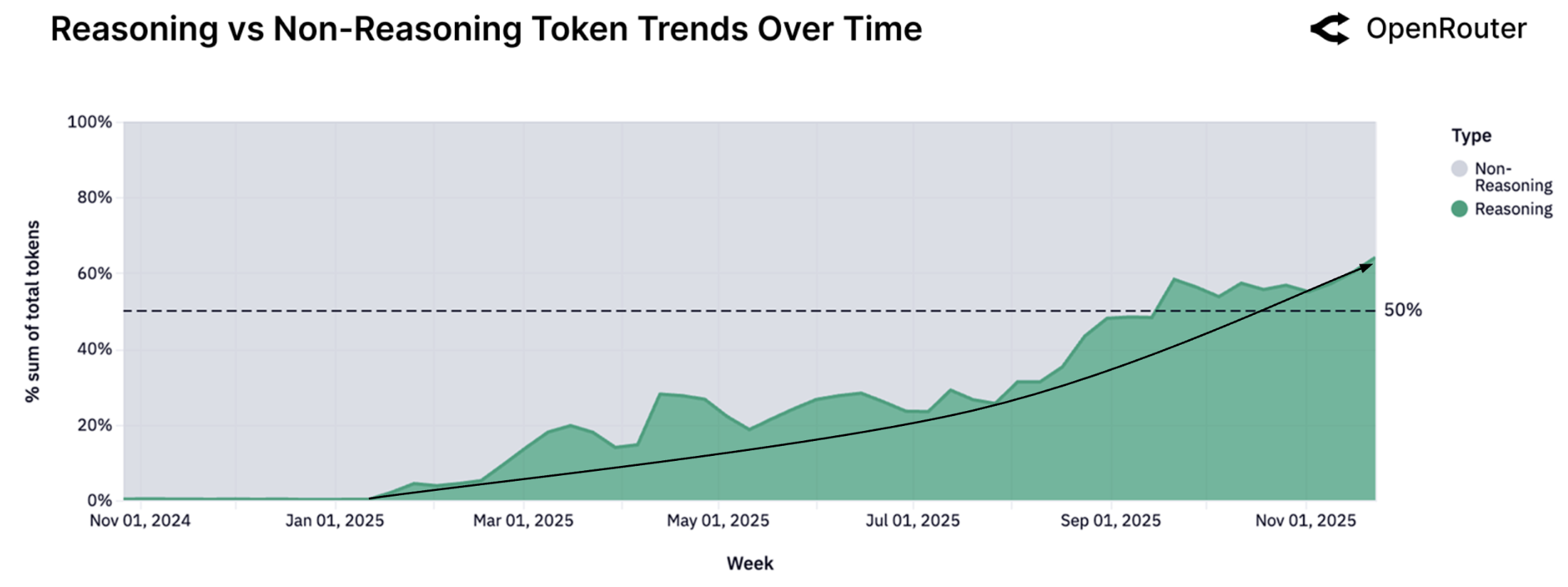}
    \caption{\textbf{Reasoning vs. Non-Reasoning Token Trends.} Share of all tokens routed through reasoning-optimized models has risen steadily since early 2025. The metric reflects the proportion of all tokens served by reasoning models, \emph{not} the share of “reasoning tokens” within model outputs.}
    \label{fig:reasoning_vs_nonreasoning}
\end{figure}

\subsection{Reasoning Models Now Represent Half of All Usage}
\label{subsec:reasoning_usage}

As shown in Figure~\ref{fig:reasoning_vs_nonreasoning}, the share of total tokens routed through reasoning-optimized models climbed sharply in 2025. What was effectively a negligible slice of usage in early Q1 now exceeds fifty percent. This shift reflects both sides of the market. On the supply side, the release of higher-capability systems like GPT-5, Claude 4.5, and Gemini 3 expanded what users could expect from stepwise reasoning. On the demand side, users increasingly prefer models that can manage task state, follow multi-step logic, and support agent-style workflows rather than simply generate text.

\begin{figure}[htbp]
    \centering
    \includegraphics[width=\linewidth]{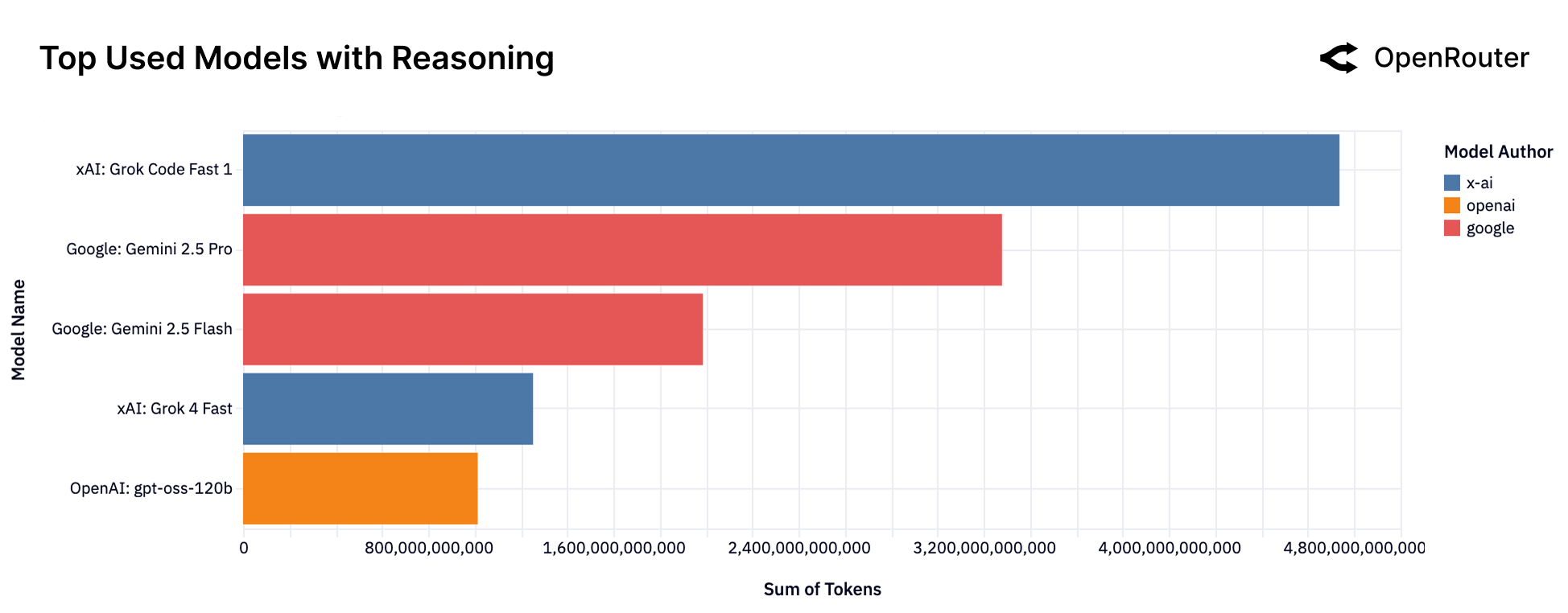}
    \caption{\textbf{Top Reasoning Models by Token Volume.} Among reasoning models, xAI’s Grok Code Fast 1 currently processes the largest share of reasoning-related token traffic, followed by Google’s Gemini 2.5 Pro and Gemini 2.5 Flash. xAI’s Grok 4 Fast and OpenAI’s gpt-oss-120b complete the top group.}
    \label{fig:reasoning_top_models}
\end{figure}

Figure~\ref{fig:reasoning_top_models} shows the top models driving this shift. In the most recent data, xAI’s Grok Code Fast 1 now drives the largest share of reasoning traffic (excluding free launch access), ahead of Google’s Gemini 2.5 Pro and Gemini 2.5 Flash. This is a notable change from only a few weeks ago, when Gemini 2.5 Pro led the category and DeepSeek R1 and Qwen3 were also in the top tier. Grok Code Fast 1 and Grok 4 Fast have gained share quickly, supported by xAI’s aggressive rollout, competitive pricing, and developer attention around its code-oriented variants. At the same time, the continued presence of open models like OpenAI’s gpt-oss-120b underscores that developers still reach for OSS when possible. The mix overall highlights how dynamic the reasoning landscape has become, with rapid model turnover shaping which systems dominate real workloads.

The data points to a clear conclusion: reasoning-oriented models are becoming the default path for real workloads, and the share of tokens flowing through them is now a leading indicator of how users want to interact with AI systems.

\subsection{Rising Adoption of Tool-Calling}
\label{subsec:tool_use}

\begin{figure}[htbp]
    \centering
    \includegraphics[width=\linewidth]{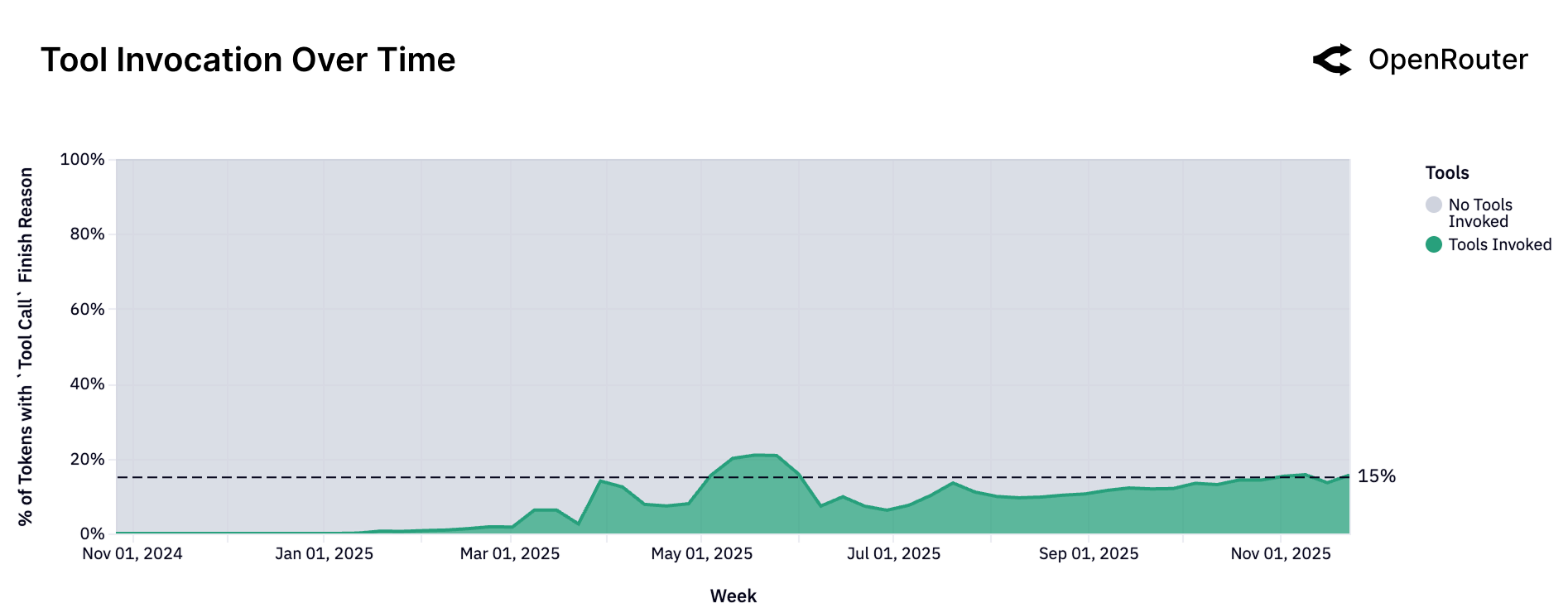}
    \caption{\textbf{Tool Invocations.} Share of total tokens normalized to requests whose finish reason was classified as a \emph{Tool Call}, meaning a tool was actually invoked during the request. This metric reflects successful tool call invocations; the number of requests that contain tool definitions is proportionally higher.}
    \label{fig:tool_calling_vs_not}
\end{figure}

In Figure~\ref{fig:tool_calling_vs_not}, we report the share of total tokens originating from requests whose finish reason was a \emph{Tool Call}. This metric is normalized and captures only those interactions in which a tool was actually invoked. 

This is in contrast to the \emph{Input Tool} signal that records whether a tool was provided to the model during a request (regardless of invocation). Input Tool counts are, by definition, higher than Tool Call finish reasons, since provision is a superset of successful execution. Whereas the finish-reason metric measures realized tool use, Input Tool reflects potential availability rather than actual invocation. Because this metric was introduced only in September 2025, we are not reporting it in this paper.

\begin{figure}[htbp]
    \centering
    \includegraphics[width=\linewidth]{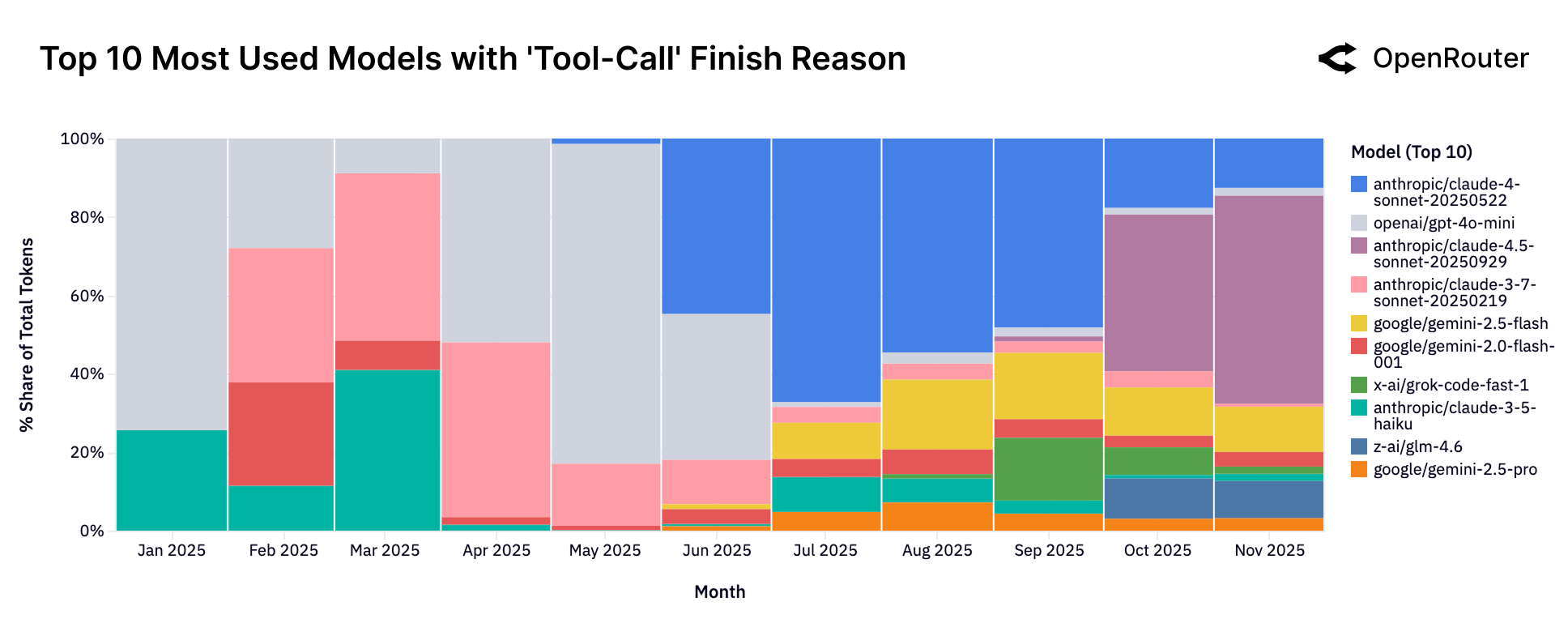}
    \caption{\textbf{Top Models by Tools Provided Volume.} Tool provision is concentrated among models explicitly optimized for agentic inference, such as Claude Sonnet, Gemini Flash.}
    \label{fig:tool_use_models}
\end{figure}

As shown in Figure~\ref{fig:tool_use_models}, tool invocation was initially concentrated among a small group of models: OpenAI’s \texttt{gpt-4o-mini} and Anthropic’s Claude 3.5 and 3.7 series, which together accounted for most tool-enabled tokens in early 2025. By mid-year, however, a broader set of models began supporting tool provision, reflecting a more competitive and diversified ecosystem. From end of September onward, newer Claude 4.5 Sonnet model rapidly gained share. Meanwhile, newer entries like \texttt{Grok Code Fast} and \texttt{GLM 4.5} have made visible inroads, reflecting broader experimentation and diversification in tool-capable deployments.

For operators, the implication is clear: enabling tool use is on the rise for high-value workflows. Models without reliable tool formats risk falling behind in enterprise adoption and orchestration environments.

\subsection{The Anatomy of Prompt-Completion Shapes}

\begin{figure}[htbp]
\centering
\includegraphics[width=\linewidth]{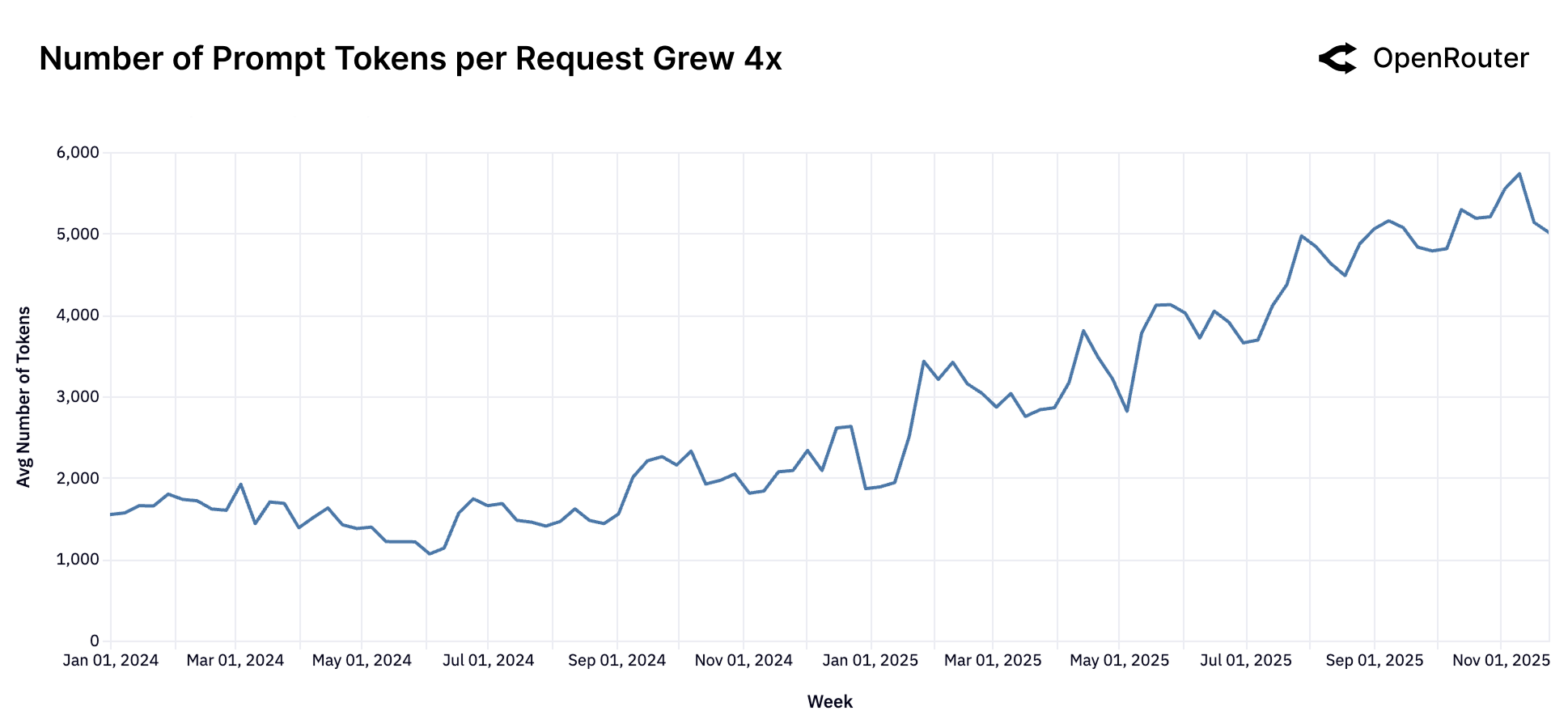}
\caption{\textbf{Number of Prompt Tokens is on the Rise.} Average prompt token length has grown nearly fourfold since early 2024, reflecting increasingly context-heavy workloads.}
\label{fig:prompt_tokens_growth}
\end{figure}

The shape of model workloads has evolved markedly over the past year. Both prompt (input) and completion (output) token volumes have risen sharply, though at different scales and rates. Average prompt tokens per request have increased roughly fourfold from around 1.5K to over 6K while completions have nearly tripled from about 150 to 400 tokens. The relative magnitude of growth highlights a decisive shift toward more complex, context-rich workloads.

\begin{figure}[htbp]
\centering
\includegraphics[width=\linewidth]{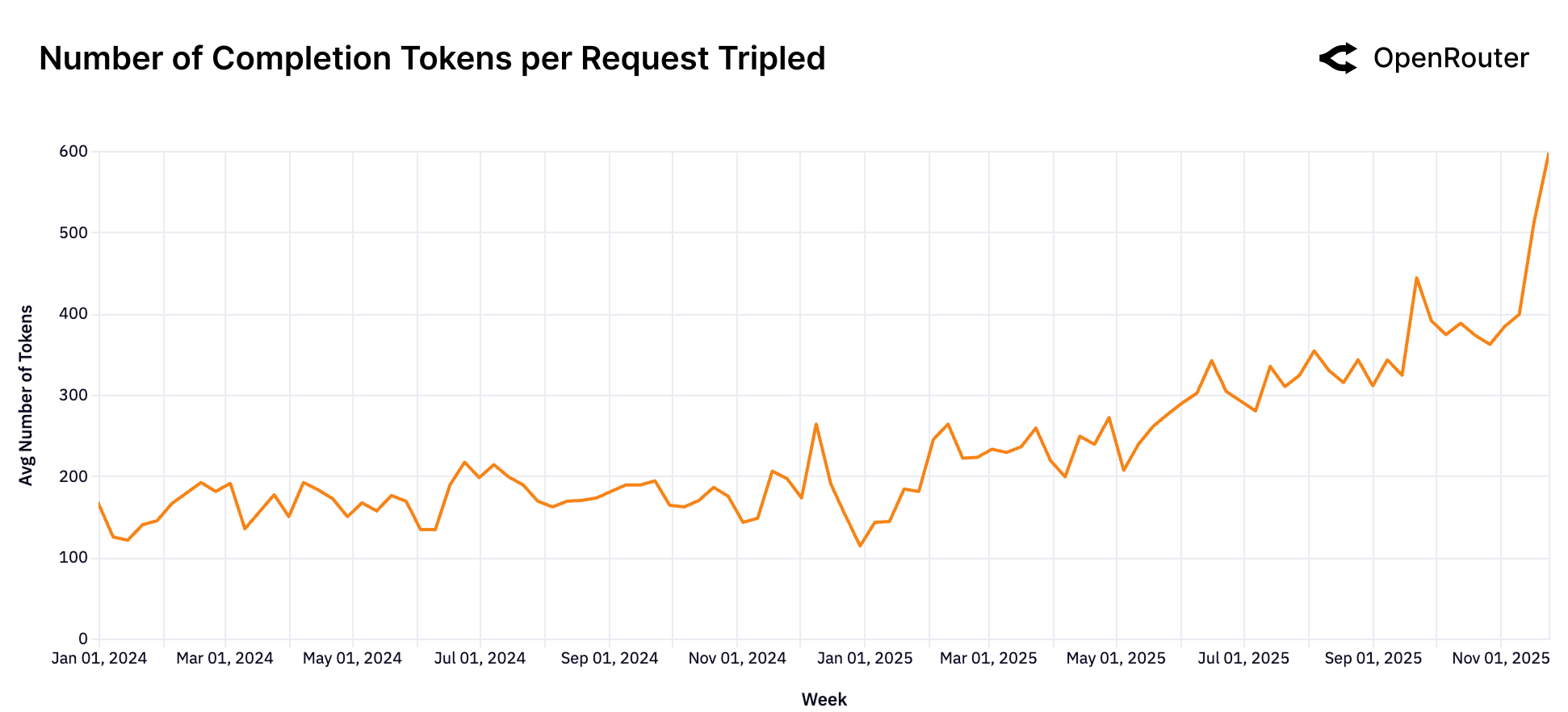}
\caption{\textbf{Number of Completion Tokens Almost Tripled.} Output lengths have also increased, though from a smaller baseline, suggesting richer, more detailed responses mostly due to reasoning tokens.}
\label{fig:completion_tokens_growth}
\end{figure}

This pattern reflects a new equilibrium in model usage. The typical request today is less about open-ended generation (“write me an essay”) and more about reasoning over substantial user-provided material such as codebases, documents, transcripts, or long conversations, and producing concise, high-value insights. Models are increasingly acting as analytical engines rather than creative generators.

\begin{figure}[htbp]
\centering
\includegraphics[width=\linewidth]{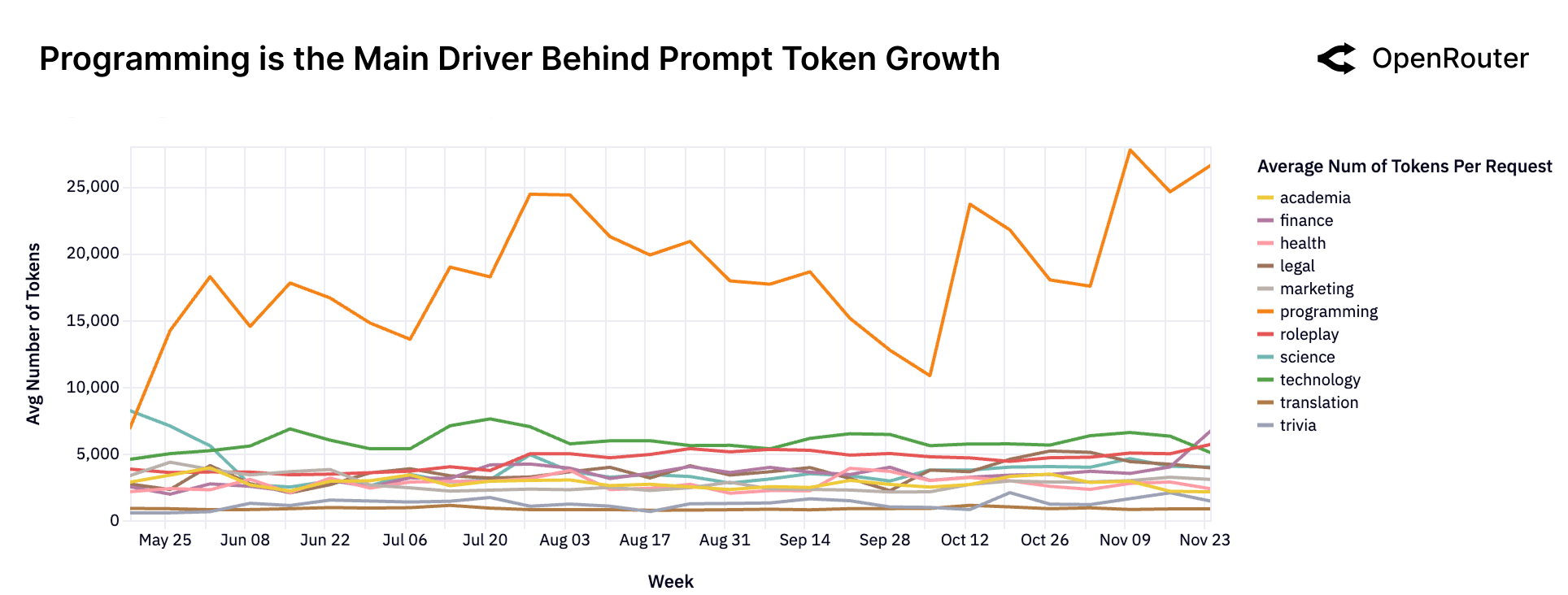}
\caption{\textbf{Programming as the Main Driver Behind Prompt Token Growth.} Since tags are available starting in Spring 2025, programming-related tasks have consistently required the largest input contexts.}
\label{fig:prompt_tokens_growth_per_category}
\end{figure}

Category-level data (available only since Spring 2025 as per Section~\ref{subsec:timeframe}) provides a more nuanced picture: programming workloads are the dominant driver of prompt token growth. Requests involving code understanding, debugging, and code generation routinely exceed 20K input tokens, while all other categories remain relatively flat and low-volume. This asymmetric contribution suggests that the recent expansion in prompt size is not a uniform trend across tasks but rather a concentrated surge tied to software development and technical reasoning use cases.

\subsection{Longer Sequences, More Complex Interactions}
\label{subsec:seq_length_growth}

\begin{figure}[htbp]
    \centering
    \includegraphics[width=\linewidth]{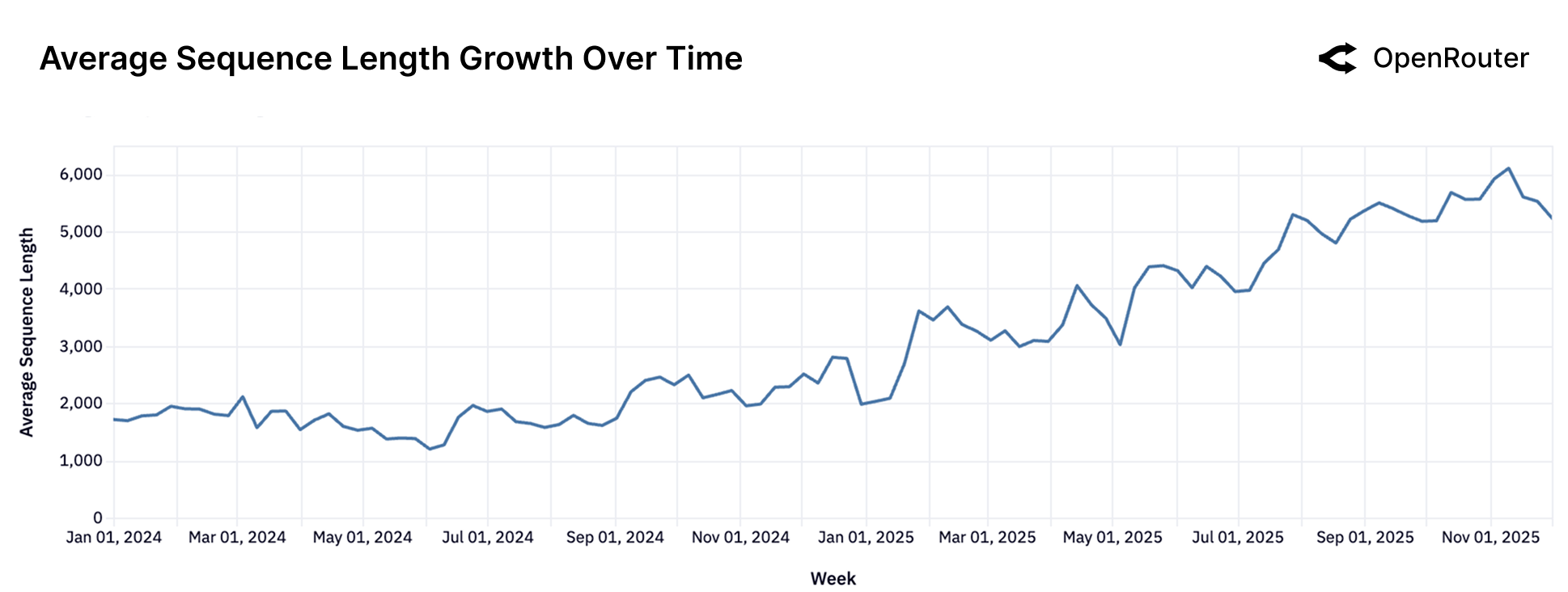}
    \caption{\textbf{Average Sequence Length Over Time.} Mean number of tokens per generation (prompt + completion).}
    \label{fig:average_seq_length}
\end{figure}

\begin{figure}[htbp]
    \centering
    \includegraphics[width=\linewidth]{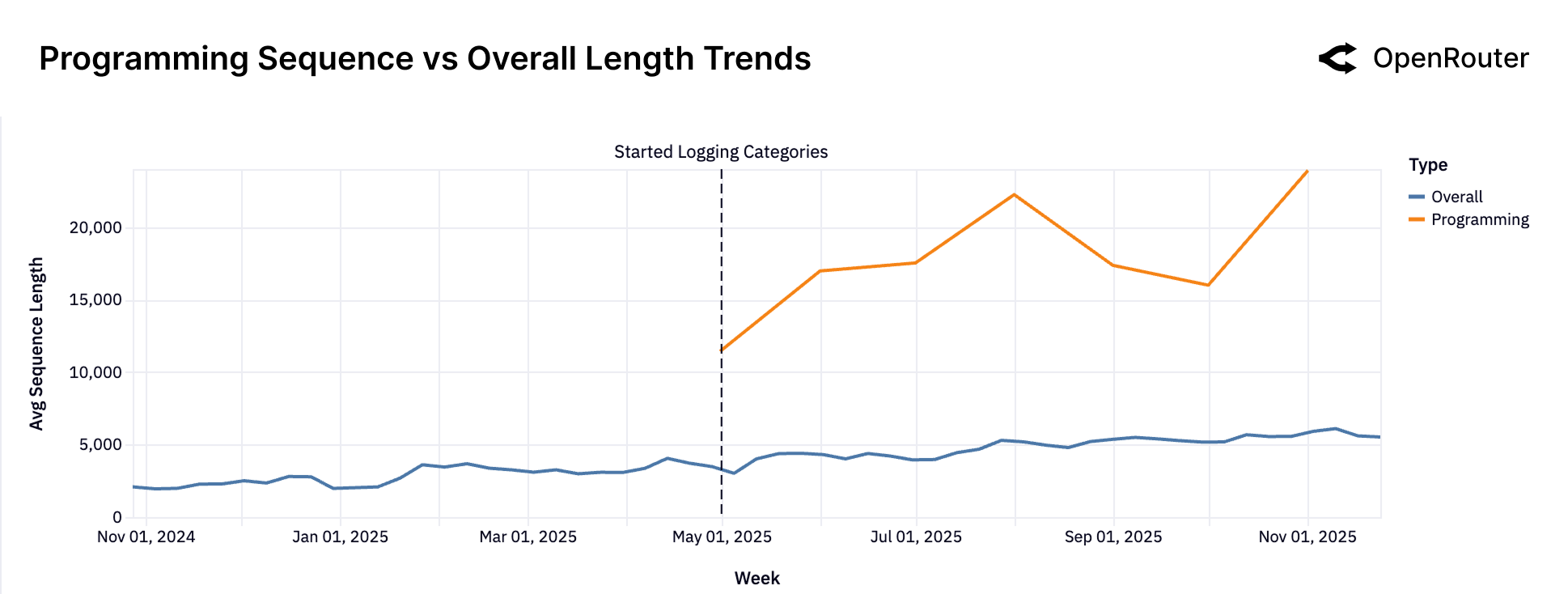}
    \caption{\textbf{Sequence Length in Programming vs Overall.} Programming prompts are systematically longer and growing faster.}
    \label{fig:seq_length_programming}
\end{figure}

Sequence length is a proxy for task complexity and interaction depth. Figure~\ref{fig:average_seq_length} shows that average sequence length has more than tripled over the past 20 months from under 2,000 tokens in late 2023 to over 5,400 by late 2025. This growth reflects a structural shift toward longer context windows, deeper task history, and more elaborate completions.

As per previous section, Figure~\ref{fig:seq_length_programming} adds further clarity: programming-related prompts now average 3–4 times the token length of general-purpose prompts. The divergence indicates that software development workflows are the primary driver of longer interactions. Long sequences are not just user verbosity: they are a signature of embedded, more sophisticated agentic workflows.

\subsection{Implications: Agentic Inference Is the New Default}

Together, these trends (rising reasoning share, expanded tool use, longer sequences, and programming’s outsize complexity) suggest that the center of gravity in LLM usage has shifted. The median LLM request is no longer a simple question or isolated instruction. Instead, it is part of a structured, agent-like loop, invoking external tools, reasoning over state, and persisting across longer contexts.

For model providers, this raises the bar for default capabilities. Latency, tool handling, context support, and robustness to malformed or adversarial tool chains are increasingly critical. For infra operators, inference platforms must now manage not just stateless requests but long-running conversations, execution traces, and permission-sensitive tool integrations. \textbf{Soon enough, if not already, agentic inference will be taking over the majority of the inference.}

\section{Categories: How Are People Using LLMs?}
\label{sec:categories}
Understanding the distribution of tasks that users perform with LLMs is central to assessing real-world demand and \textit{model–market fit}. As described in  Section~\ref{subsec:google_tagclassifier}, we categorized billions of model interactions into high-level application categories. In Section~\ref{sec:oss_categories}, we focused on open source models to see community-driven usage. Here, we broaden the lens to \emph{all} LLM usage on OpenRouter (both closed and open models) to get a comprehensive picture of what people use LLMs for in practice.

\subsection{Dominant Categories}

\begin{figure}[htbp] 
    \centering 
    \includegraphics[width=1\linewidth]{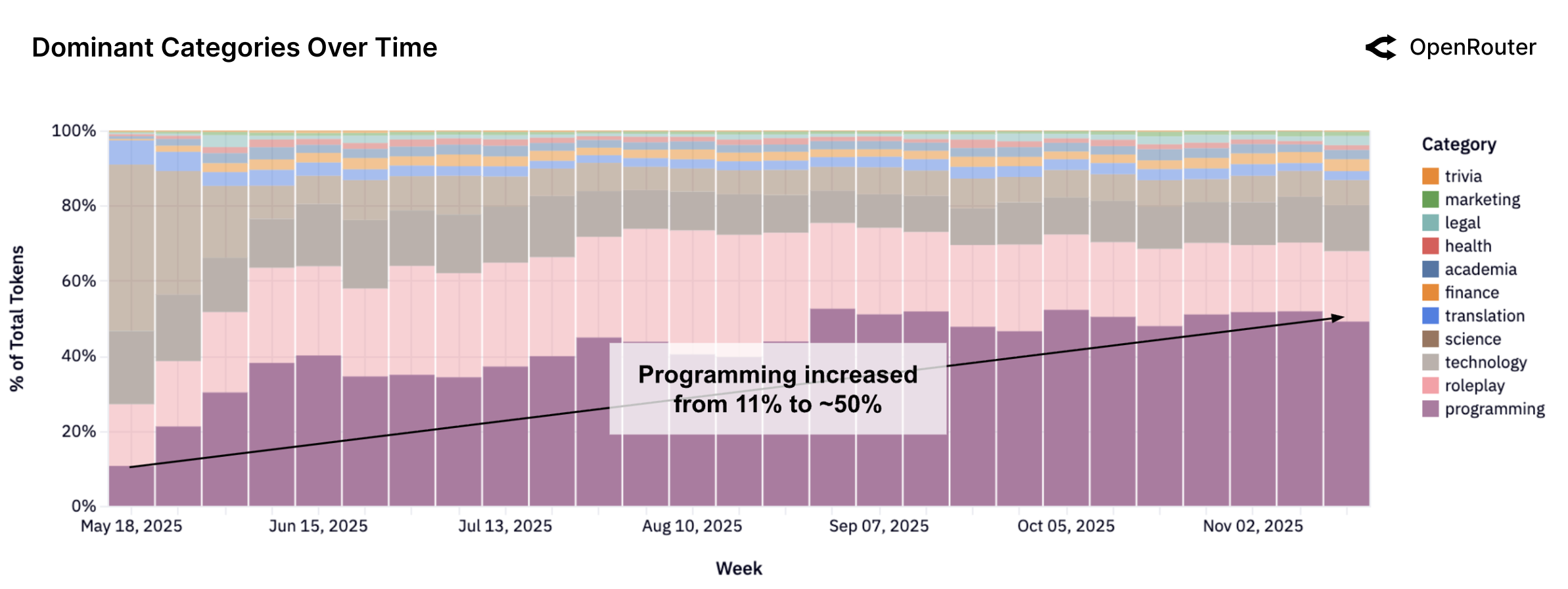} 
    \caption{\textbf{Programming as a dominant and growing category.} The share of all LLM queries classified under programming has increased steadily, reflecting the rise of AI-assisted development workflows.} 
    \label{fig:programming_domnant_tag} 
\end{figure}

Programming has become the most consistently expanding category across all models. The share of programming-related requests has grown steadily through 2025, paralleling the rise of LLM-assisted development environments and tool integrations. As shown in Figure~\ref{fig:programming_domnant_tag}, programming queries accounted for roughly 11\% of total token volume in early 2025 and exceeded 50\% in recent weeks. This trend reflects a shift from exploratory or conversational use toward applied tasks such as code generation, debugging, and data scripting. As LLMs become embedded in developer workflows, their role as programming tools is being normalized. This evolution has implications for model development, including increased emphasis on code-centric training data, improved reasoning depth for multi-step programming tasks, and tighter feedback loops between models and integrated development environments. 

\begin{figure}[htbp]
    \centering 
    \includegraphics[width=1\linewidth]{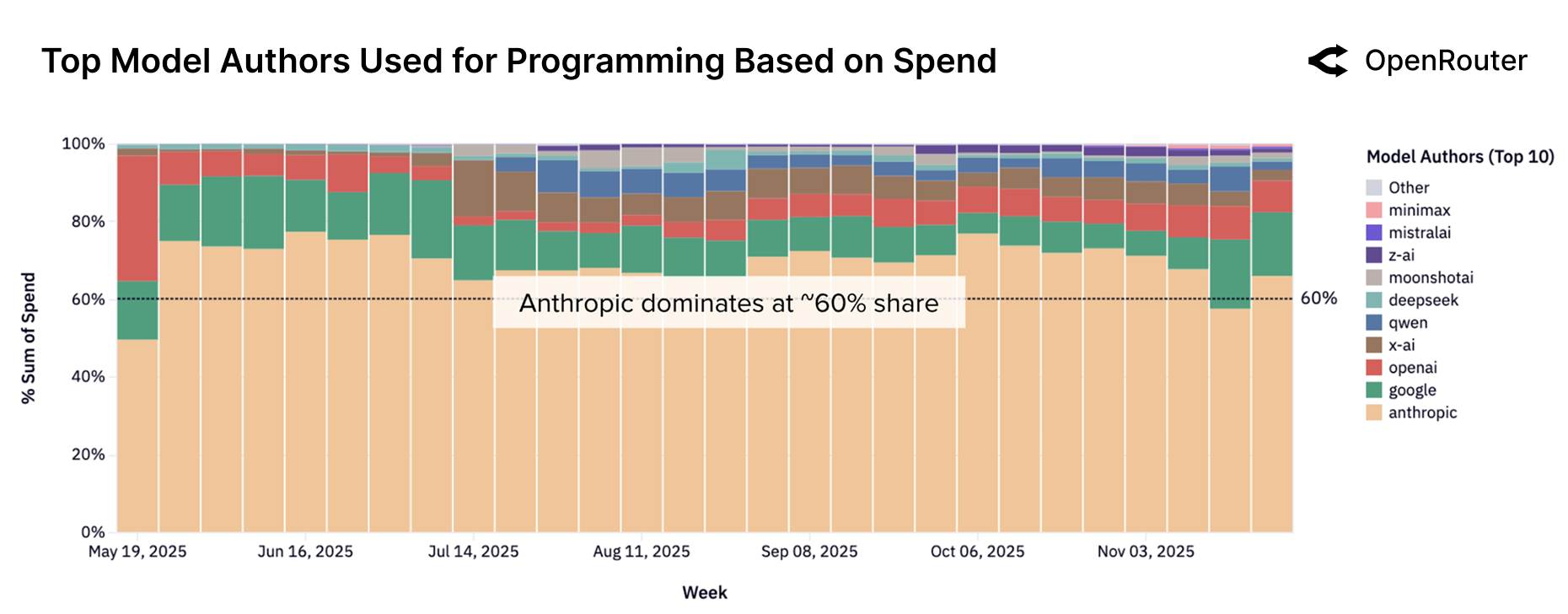} 
    \caption{\textbf{Share of programming requests by model provider.} The programming workload is highly concentrated: Anthropic’s models serve the largest share of coding queries, followed by OpenAI and Google, with a growing slice taken by MiniMax. Other providers collectively account for only a small fraction. This graph omits xAI that had substantial usage but was given away for free for a period of time.} 
    \label{fig:top_models_programming_tag} 
\end{figure}

This growing demand for programming support is reshaping competitive dynamics across model providers. As shown in Figure~\ref{fig:top_models_programming_tag}, Anthropic’s Claude series has consistently dominated the category, accounting for more than 60\% of programming-related spend for most of the observed period. The landscape has nevertheless evolved meaningfully. During the week of November 17, Anthropic’s share fell below the 60\% threshold for the first time. Since July, OpenAI has expanded its share from roughly 2\% to about 8\% in recent weeks, likely reflecting a renewed emphasis on developer-centric workloads. Over the same interval, Google’s share has remained stable at approximately 15\%. The mid-tier segment is also in motion. Open-source providers including Z.AI, Qwen, and Mistral AI are steadily gaining mindshare. MiniMax, in particular, has emerged as a fast-rising entrant, showing notable gains in recent weeks.

Overall, \textbf{programming has become one of the most contested and strategically important model categories}. It attracts sustained attention from top labs, and even modest changes in model quality or latency can shift share week to week. For infrastructure providers and developers, this highlights the need for continual benchmarking and evals, especially as the frontier is constantly evolving.

\subsection{Tag Composition Within Categories}
\label{subsec:category_composition}

\begin{figure}[htbp]
    \centering
    \begin{subfigure}[t]{1\linewidth}
        \centering
        \includegraphics[width=\linewidth]{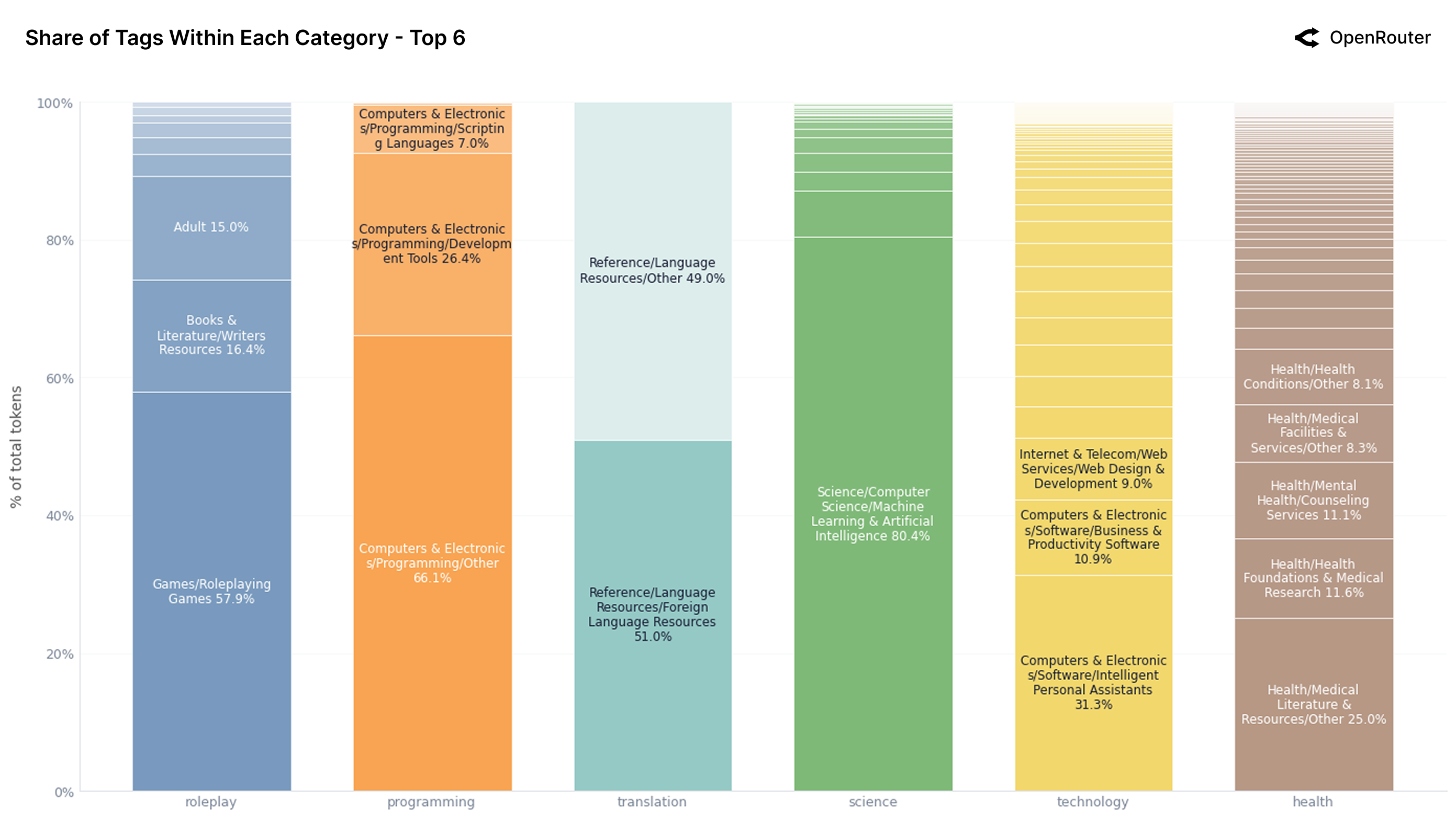}
        \caption{\textbf{Top 6 categories by total token share.} Each bar shows the breakdown of dominant sub-tags within that category. Labels indicate sub-tags contributing at least 7\% of tokens for the category.}
        \label{fig:category_tags_top6}
    \end{subfigure}
    \hfill
    \begin{subfigure}[t]{1\linewidth}
        \centering
        \includegraphics[width=\linewidth]{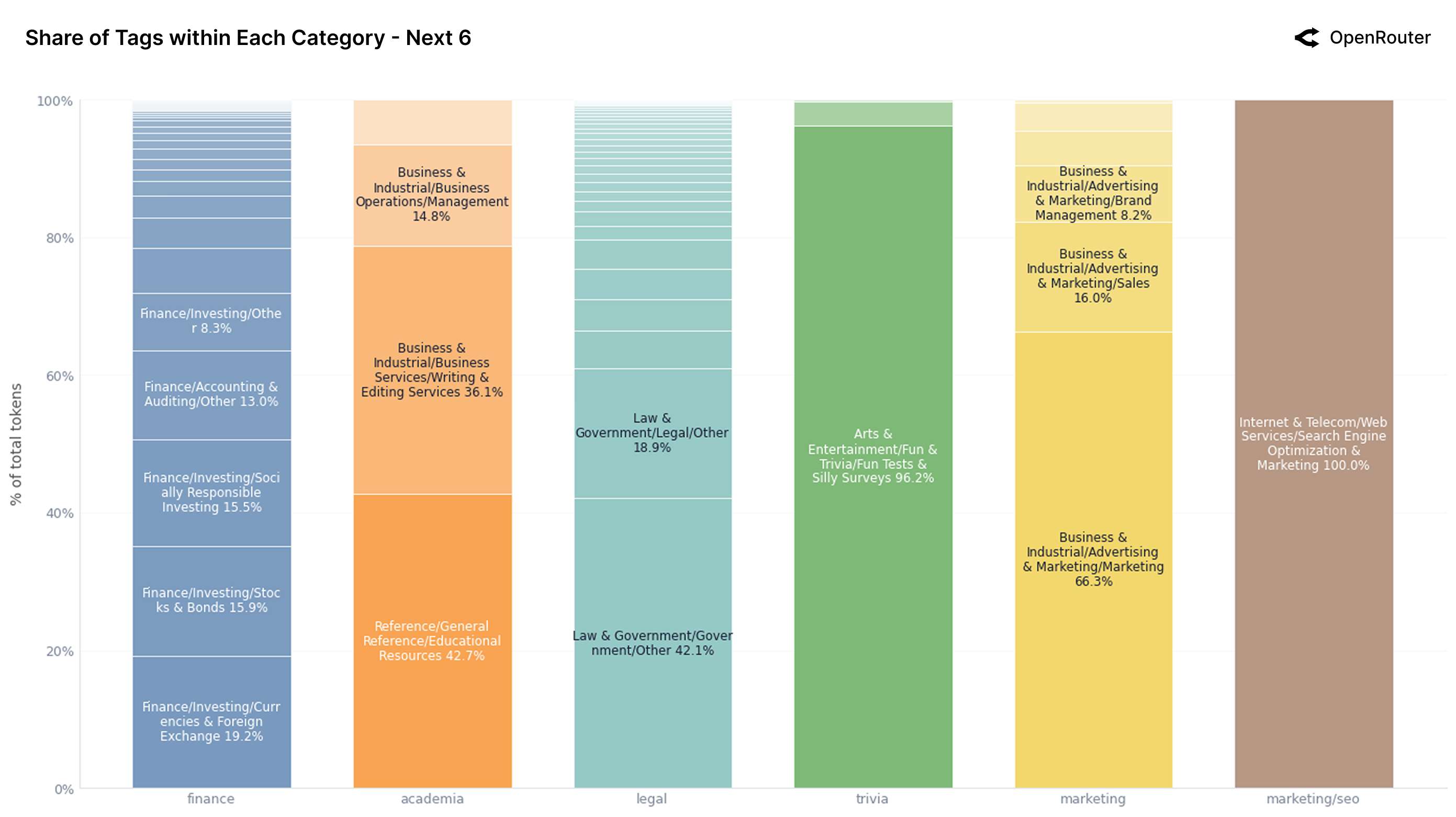}
        \caption{\textbf{Next 6 categories by token share.} Similar breakdown for secondary categories, illustrating the concentration (or lack thereof) of subtopics in each domain.}
        \label{fig:category_tags_next6}
    \end{subfigure}
    \caption{\textbf{Token share by sub-tag within each category.} The charts cover the top twelve categories overall, grouped into two panels for readability. Each column is a 100\% stacked bar of tag-level shares, revealing the internal composition of each category’s usage.}
    \label{fig:category_tag_distribution}
\end{figure}

Figure~\ref{fig:category_tag_distribution} breaks down LLM usage across the twelve most common content categories, revealing the internal sub-topic structure of each. A key takeaway is that most categories are not evenly distributed: they are dominated by one or two recurring use patterns, often reflecting concentrated user intent or alignment with LLM strengths.

Among the highest-volume categories, \textbf{roleplay} stands out for its consistency and specialization. Nearly 60\% of roleplay tokens fall under \emph{Games/Roleplaying Games}, suggesting that users treat LLMs less as casual chatbots and more as structured roleplaying or character engines. This is further reinforced by the presence of \emph{Writers Resources} (15.6\%) and \emph{Adult} content (15.4\%), pointing to a blend of interactive fiction, scenario generation, and personal fantasy. Contrary to assumptions that roleplay is mostly informal dialogue, the data show a well-defined and replicable genre-based use case.

\textbf{Programming} is similarly skewed, with over two-thirds of traffic labeled as \emph{Programming/Other}. This signals the broad and general-purpose nature of code-related prompts: users are not narrowly focused on specific tools or languages but are asking LLMs for everything from logic debugging to script drafting. That said, \emph{Development Tools} (26.4\%) and small shares from scripting languages indicate emerging specialization. This fragmentation highlights an opportunity for model builders to improve tagging or training around structured programming workflows.

Beyond the dominant categories of roleplay and programming, the remaining domains represent a diverse but lower-volume tail of LLM usage. While individually smaller, they reveal important patterns about how users interact with models across specialized and emerging tasks. For example, \textbf{translation}, \textbf{science}, and \textbf{health} show relatively flat internal structure. In translation, usage is nearly evenly split between \emph{Foreign Language Resources} (51.1\%) and \emph{Other}, suggesting diffuse needs: multilingual lookup and rephrasing, rather than sustained document-level translation. Science is dominated by a single tag, \emph{Machine Learning \& AI} (80.4\%), indicating that most scientific queries are meta-AI questions rather than general STEM topics like physics or biology. This reflects either user interest or model strengths skewed toward self-referential inquiry. 

Health, in contrast, is the most fragmented of the top categories, with no sub-tag exceeding 25\%. Tokens are spread across medical research, counseling services, treatment guidance, and diagnostic lookups. This diversity highlights the domain’s complexity, but also the challenge of modeling it safely: LLMs must span high variance user intent, often in sensitive contexts, without clear concentration in a single use case.

What links these long-tail categories is their broadness: users turn to LLMs for exploratory, lightly structured, or assistance-seeking interactions, but without the focused workflows seen in programming or personal assistants. Taken together, these secondary categories may not dominate volume, but they hint at latent demand. They signal that LLMs are being used at the fringes of many fields from translation to medical guidance to AI introspection and that as models improve in domain robustness and tooling integration, we may see these scattered intents converge into clearer, higher-volume applications.

By contrast, \textbf{finance}, \textbf{academia}, and \textbf{legal} are much more diffuse. Finance spreads its volume across foreign exchange, socially responsible investing, and audit/accounting: no single tag breaks 20\%. Legal shows similar entropy, with usage split between \emph{Government/Other} (43.0\%) and \emph{Legal/Other} (17.8\%). This fragmentation may reflect the complexity of these domains, or simply the lack of targeted LLM workflows for them compared to more mature categories like coding and chat.

The data suggest that real-world LLM usage is not uniformly exploratory: it clusters tightly around a small set of repeatable, high-volume tasks. Roleplay, programming, and personal assistance each exhibit clear structure and dominant tags. Science, health, and legal domains, by contrast, are more diffuse and likely under-optimized. These internal distributions can guide model design, domain-specific fine-tuning, and application-level interfaces particularly in tailoring LLMs to user goals.

\subsection{Author-Level Insights by Category}
\begin{figure*}[htbp]
  \centering
  \begin{subfigure}{\textwidth}
    \includegraphics[width=\textwidth]{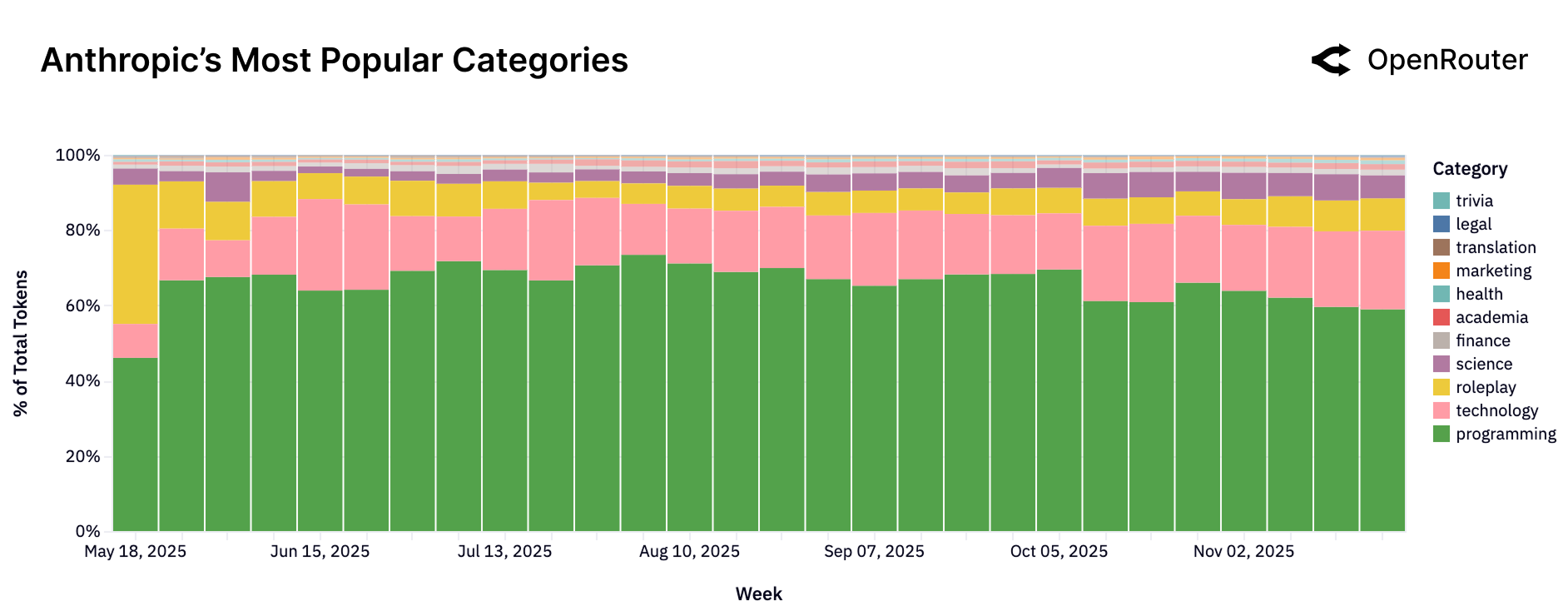}
    \caption{\textbf{Anthropic.} Predominantly used for programming and technology tasks (over 80\%), with minimal roleplay usage.}
    \label{fig:anthropic_top_tags}
  \end{subfigure}\vspace{6pt}
  \begin{subfigure}{\textwidth}
    \includegraphics[width=\textwidth]{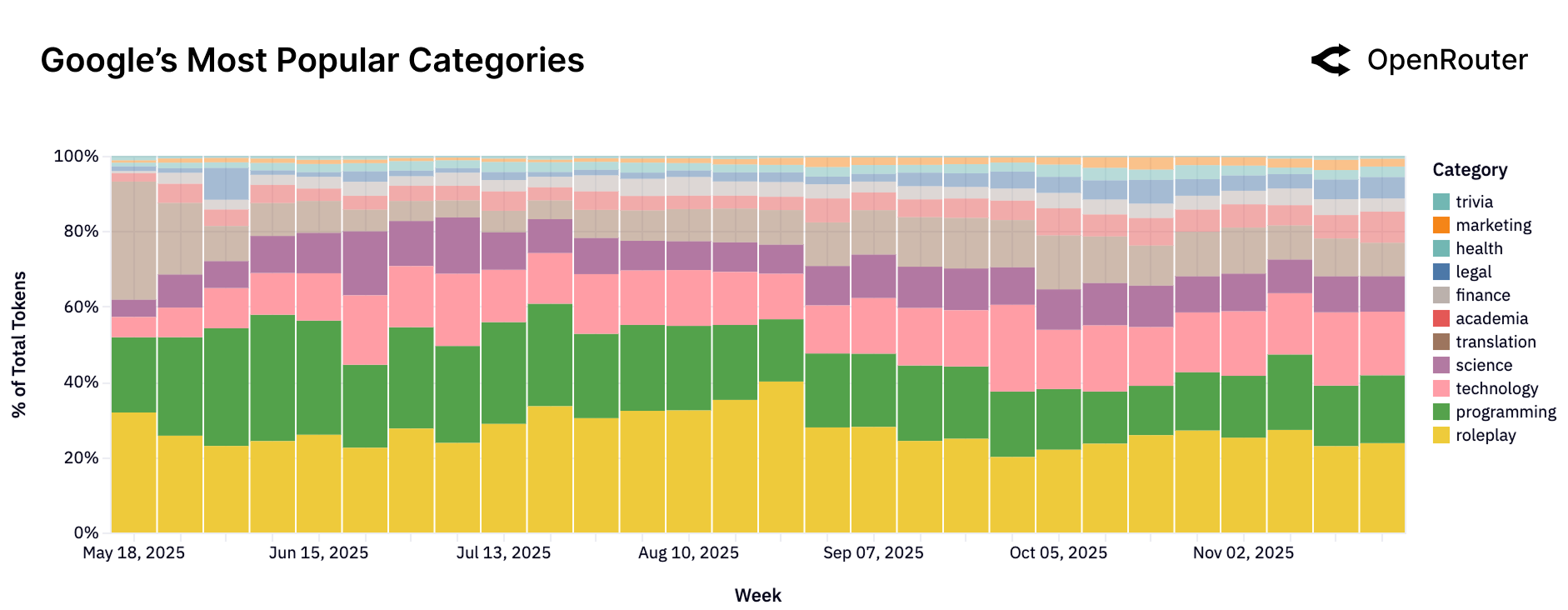}
    \caption{\textbf{Google.} A broad usage composition spanning legal, science, technology, and some general knowledge queries.}
    \label{fig:google_top_tags}
  \end{subfigure}\vspace{6pt}
  \begin{subfigure}{\textwidth}
    \includegraphics[width=\textwidth]{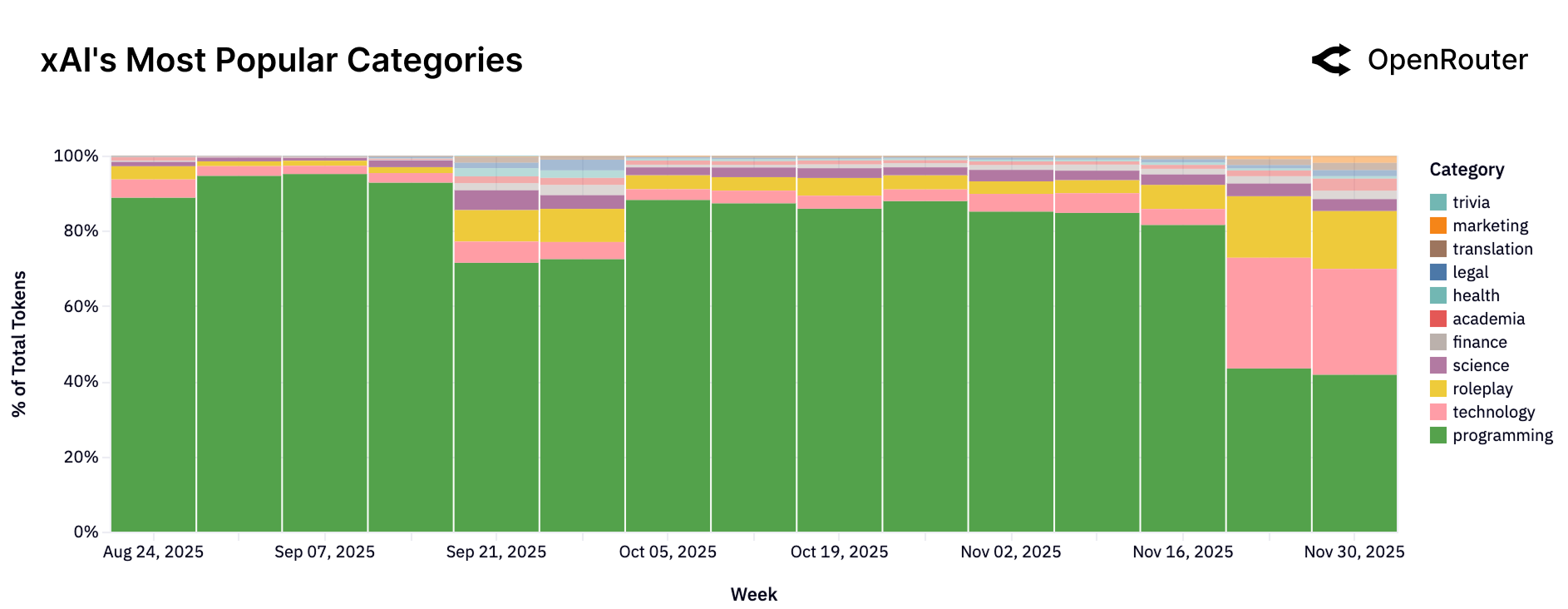}
    \caption{\textbf{xAI.} Token usage heavily centered on programming, with technology, roleplay, and academia emerging more prominently in late November.}
    \label{fig:xai_top_tags}
  \end{subfigure}\vspace{8pt}
  \caption{\textbf{Top content categories for major model providers (by share of provider’s token usage).} Each bar illustrates how a provider’s usage is distributed across categories, highlighting specialization and changes over time.}
  \label{fig:model_tag_comparison}
\end{figure*}

\begin{figure*}[htbp]
  \centering
  \begin{subfigure}[t]{0.98\textwidth}
    \includegraphics[width=\textwidth]{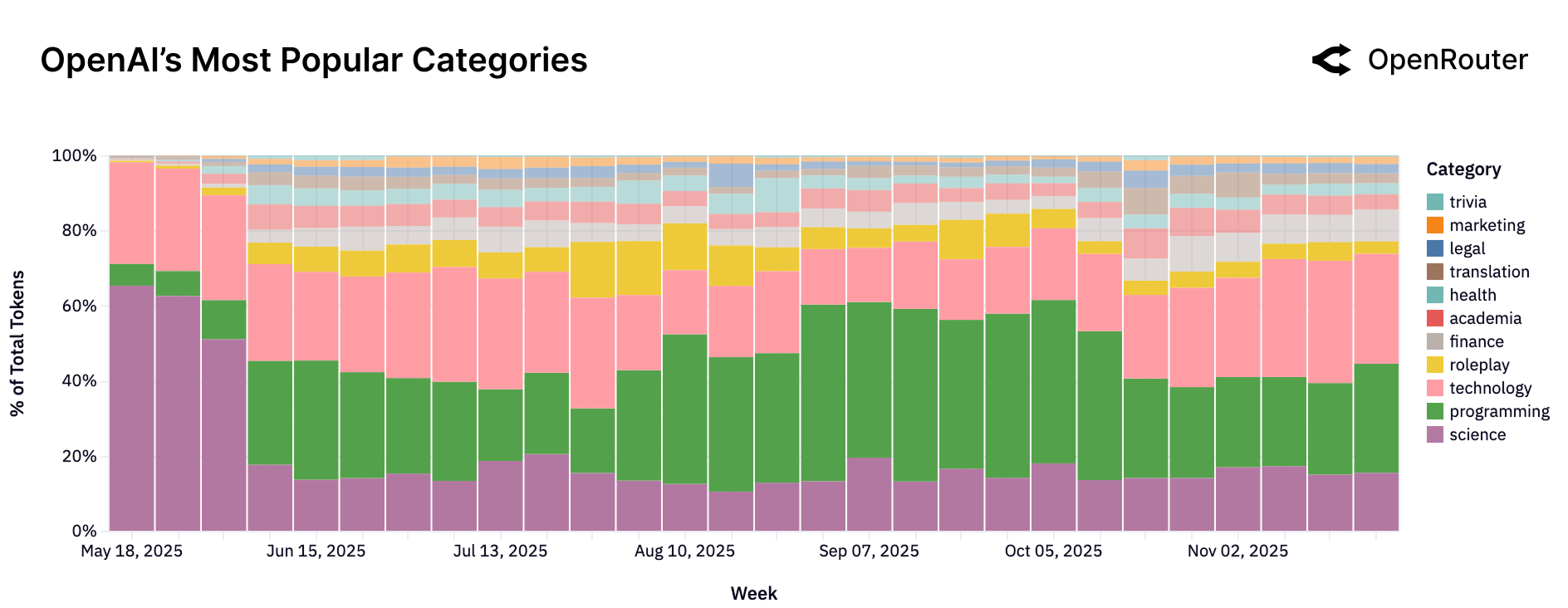}
    \caption{\textbf{OpenAI.} Shifting toward programming and technology tasks over time, with roleplay and casual chat decreasing significantly.}
    \label{fig:openai_top_tags}
  \end{subfigure}

  \begin{subfigure}[t]{0.98\textwidth}
    \includegraphics[width=\textwidth]{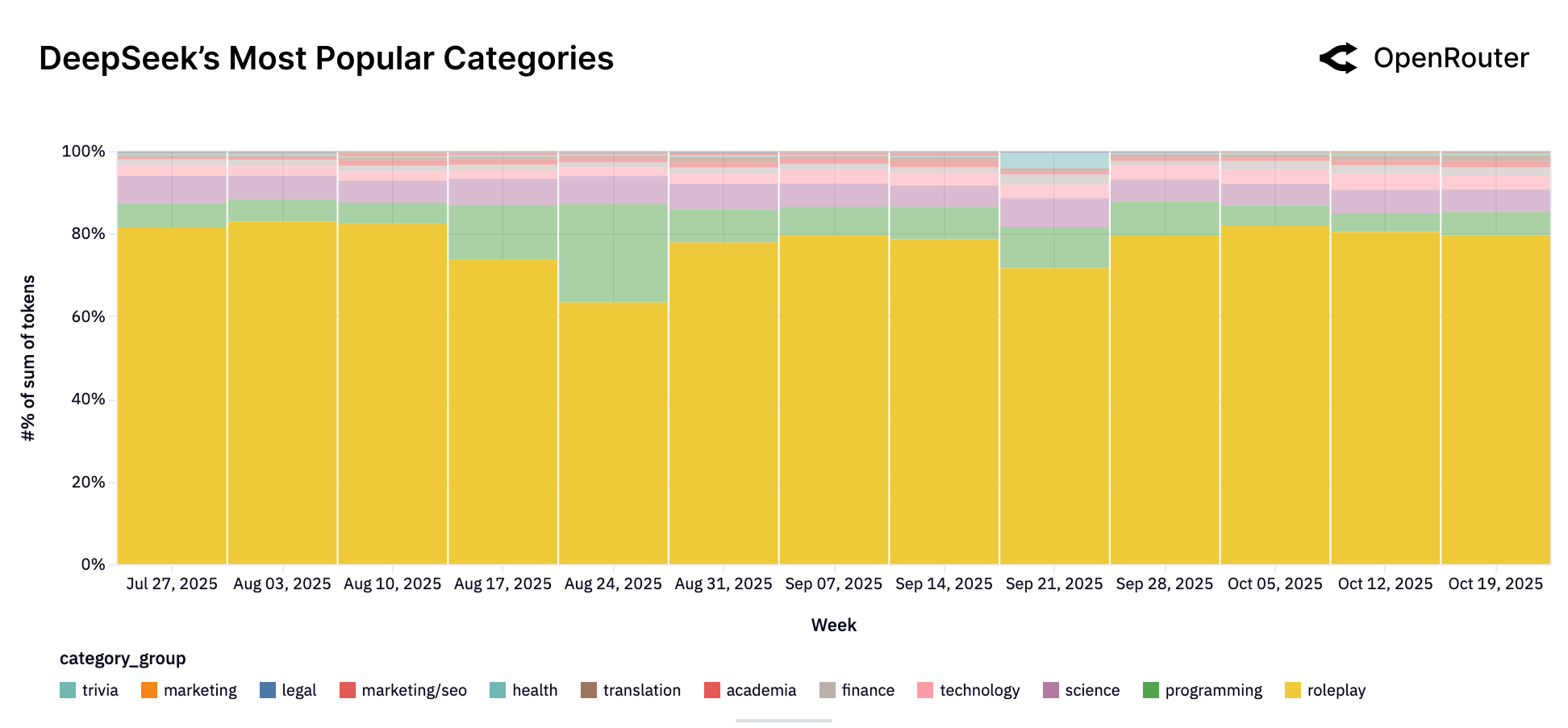}
    \caption{\textbf{DeepSeek.} Usage dominated by roleplay and casual interaction.}
    \label{fig:deepseek_top_tags}
  \end{subfigure}\vspace{6pt}

  \begin{subfigure}[t]{0.98\textwidth}
    \includegraphics[width=\textwidth]{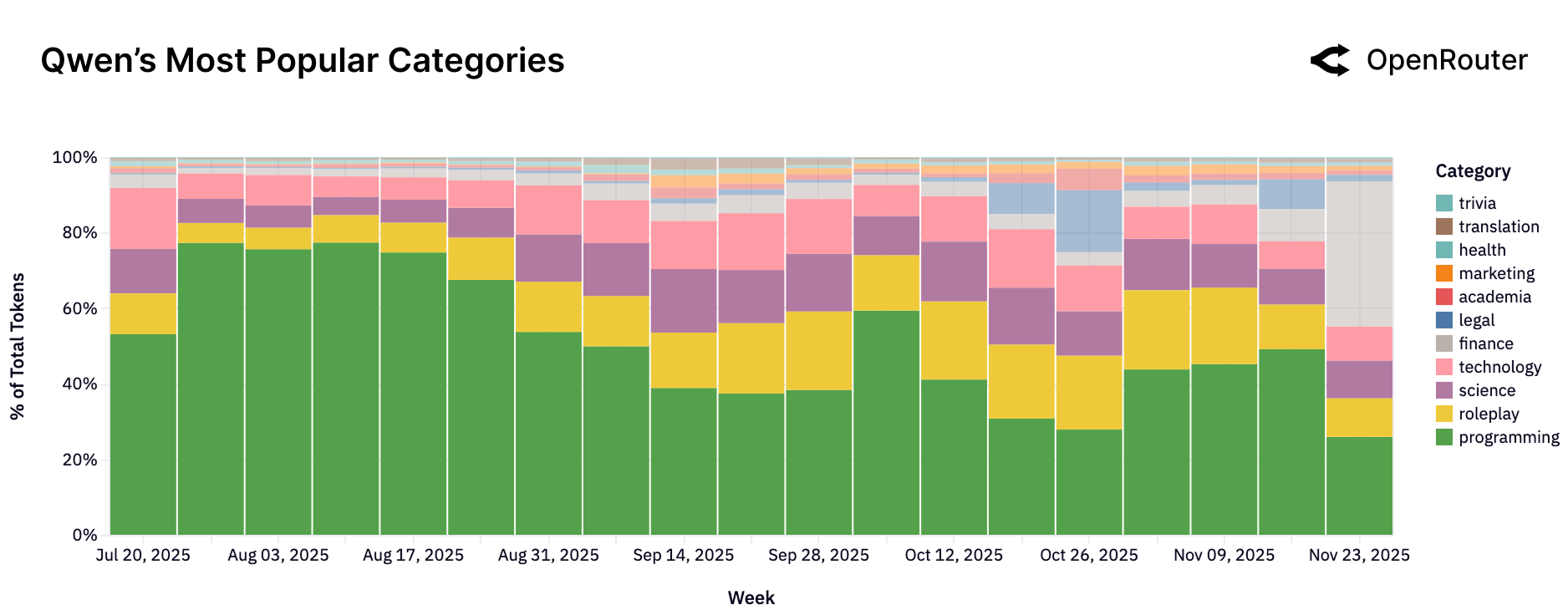}
    \caption{\textbf{Qwen.} Strong concentration in programming tasks, with roleplay and science categories fluctuating over time.}
    \label{fig:qwen_top_tags}
  \end{subfigure}

  \caption{\textbf{Top content categories by share of provider token usage.} Continuation of Figure~\ref{fig:model_tag_comparison}.}
  \label{fig:xai_deepseek_qwen_comparison}
\end{figure*}

Different model authors are utilized in different usage patterns. Figures~\ref{fig:anthropic_top_tags}–\ref{fig:openai_top_tags} show the distribution of content categories for three major model families (Anthropic’s Claude, Google’s models, and OpenAI’s GPT series). Each bar represents 100\% of that provider’s token usage, broken down by top tags.

Anthropic’s Claude (Figure~\ref{fig:anthropic_top_tags}) is heavily skewed toward \textbf{Programming} + \textbf{Technology} uses, which together exceed 80\% of its usage. Roleplay and general Q\&A are only a small sliver. This confirms Claude’s positioning as a model optimized for complex reasoning, coding, and structured tasks; developers and enterprises appear to use Claude mainly as a coding assistant and problem solver. 

Google’s model usage (Figure~\ref{fig:google_top_tags}) is more diverse. We see notable segments for \textbf{Translation}, \textbf{Science}, \textbf{Technology}, and some \textbf{General Knowledge}. For instance, ~5\% of Google’s usage was legal or policy content, and another ~10\% science-related. It might hint at Gemini’s broad training focus. Compared to others, Google has relatively less and, in fact, declining coding share by late 2025 (down to roughly 18\%) and a broader tail of categories. This suggests Google’s models are being used more as general-purpose information engines.

xAI’s usage profile (Figure~\ref{fig:xai_top_tags}) is distinct from the other providers. For most of the period, usage is overwhelmingly concentrated in \textbf{Programming}, often exceeding eighty percent of all tokens. Only in late November does the distribution broaden, with noticeable gains in \textbf{Technology}, \textbf{Roleplay}, and \textbf{Academia}. This sharp shift aligns with the timing of xAI’s model being distributed at no cost through select consumer applications, which likely introduced a large influx of non-developer traffic. The result is a usage composition that blends an early, developer-heavy core with a sudden wave of general-purpose engagement, suggesting that xAI’s adoption path is being shaped both by technical users and by episodic surges tied to promotional availability.

OpenAI’s usage profile (Figure~\ref{fig:openai_top_tags}) has shifted markedly through 2025. Earlier in the year, science tasks accounted for more than half of all OpenAI tokens; by late 2025, that share had declined to under 15\%. Meanwhile, programming and technology-related usage now comprise more than half of total volume (29\% each), reflecting deeper integration into developer workflows, productivity tools, and professional applications. OpenAI’s usage composition now sits between Anthropic’s tightly focused profile and Google’s more diffuse distribution, suggesting a broad base of utility with growing tilt toward high-value, structured tasks. 

As shown in Figure~\ref{fig:xai_deepseek_qwen_comparison}, DeepSeek and Qwen exhibit usage patterns that diverge considerably from the other model families discussed earlier. DeepSeek’s token distribution is dominated by roleplay, casual chat, and entertainment-oriented interaction, often accounting for more than two thirds of its total usage. Only a small fraction of activity falls into structured tasks such as programming or science. This pattern reflects DeepSeek’s strong consumer orientation and its positioning as a high-engagement conversational model. Notably, DeepSeek displays a modest but steady increase in programming-related usage toward late summer, suggesting incremental adoption in lightweight development workflows.

Qwen, by contrast, presents an almost inverted profile. Across the entire period shown in Figure~\ref{fig:xai_deepseek_qwen_comparison}, programming consistently represents 40-60 percent of all tokens, signaling a clear emphasis on technical and developer tasks. Compared with Anthropic’s more stable engineering-heavy composition, Qwen demonstrates higher volatility across adjacent categories such as science, technology, and roleplay. These week-to-week shifts imply a heterogeneous user base and rapid iteration in applied use cases. A noticeable rise in roleplay usage during September and October, followed by a contraction in November, hints at evolving user behavior or adjustments in downstream application routing.

\textbf{In summary}, each provider shows a distinct profile aligned with its strategic focus. The differences highlight why no single model or provider covers all use cases optimally; it also underscores the potential benefits of multi-model ecosystem.

\section{Geography: How LLM Usage Differs Across Regions}
\label{sec:geography}

Global LLM usage exhibits pronounced regional variation. By examining geographic breakdowns, we can infer how local usage and spend shape LLM usage patterns. While figures below reflect OpenRouter’s user base,  they offer one snapshot of regional engagements. 

\subsection{Regional Distribution of Usage}

\begin{figure*}[htbp]
\centering
\includegraphics[width=\textwidth]{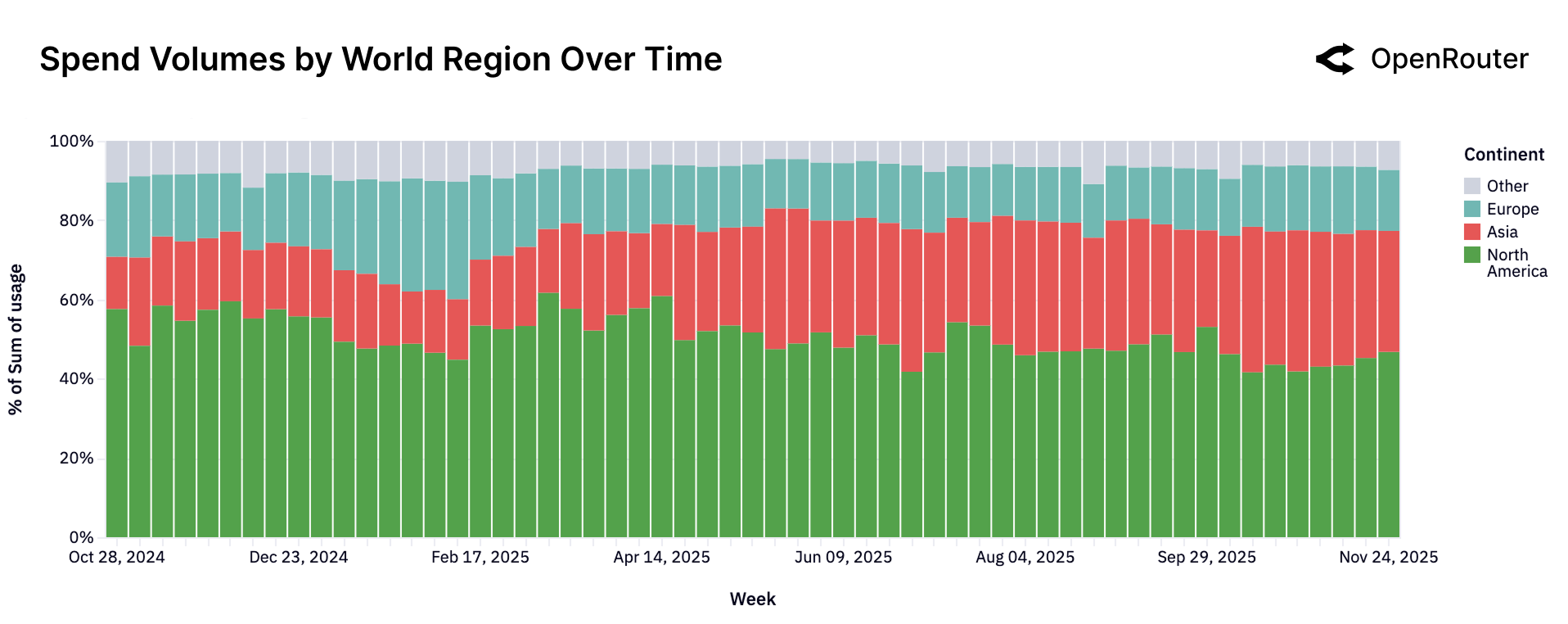}
\caption{\textbf{Spend volumes by world region over time.} Weekly share of global usage attributed to each continent.}
\label{fig:spend_by_region}
\end{figure*}

The distribution of spend, as shown in Figure~\ref{fig:spend_by_region}, underscores the increasingly global nature of AI inference market. North America, while still the single largest region, now accounts for less than half of total spend for most of the observed period. Europe shows a stable and durable contribution. Its relative share of weekly spend remains consistent throughout the timeline, typically occupying a band between the mid-teens and low twenties. A notable development is the rise of Asia not only as a producer of frontier models but also as a rapidly expanding consumer. In the earliest weeks of the dataset, Asia represented roughly thirteen percent of global spend. Over time, this share more than doubled, reaching approximately 31\% in the most recent period.

\subsection{Language Distribution}
\label{subsec:language_distribution}

As shown in Table~\ref{fig:language_distribution}, English dominates usage, accounting for more than 80\% of all tokens. This reflects both the prevalence of English-language models and the developer-centric skew of OpenRouter’s user base. However, other languages particularly Chinese, Russian, and Spanish, make up a meaningful tail. Simplified Chinese alone accounts for nearly 5\% of global tokens, suggesting sustained engagement by users in bilingual or Chinese-first environments, especially given the growth of Chinese OSS models like DeepSeek and Qwen.

\begin{table}[ht]
\centering
\caption{\textbf{Token volume by language.} Languages are based on detected prompt language across all OpenRouter traffic.}
\label{fig:language_distribution}
\begin{tabular}{lc}
\toprule
\textbf{Language} & \textbf{Token Share (\%)} \\
\midrule
English              & 82.87 \\
Chinese (Simplified) & 4.95  \\
Russian              & 2.47  \\
Spanish              & 1.43  \\
Thai                 & 1.03  \\
Other (combined)     & 7.25  \\
\bottomrule
\end{tabular}
\end{table}

For model builders and infrastructure operators, cross-regional usability, across languages, compliance regimes, and deployment settings, is becoming table stakes in a world where LLM adoption is simultaneously global and locally optimized.

\section{Analysis of LLM User Retention}
\label{sec:retention-dynamics}

\subsection{The Cinderella ``Glass Slipper'' Phenomenon}

\begin{figure}[htbp]
    \centering
    \begin{subfigure}[b]{0.48\textwidth}
        \centering
        \includegraphics[width=\textwidth]{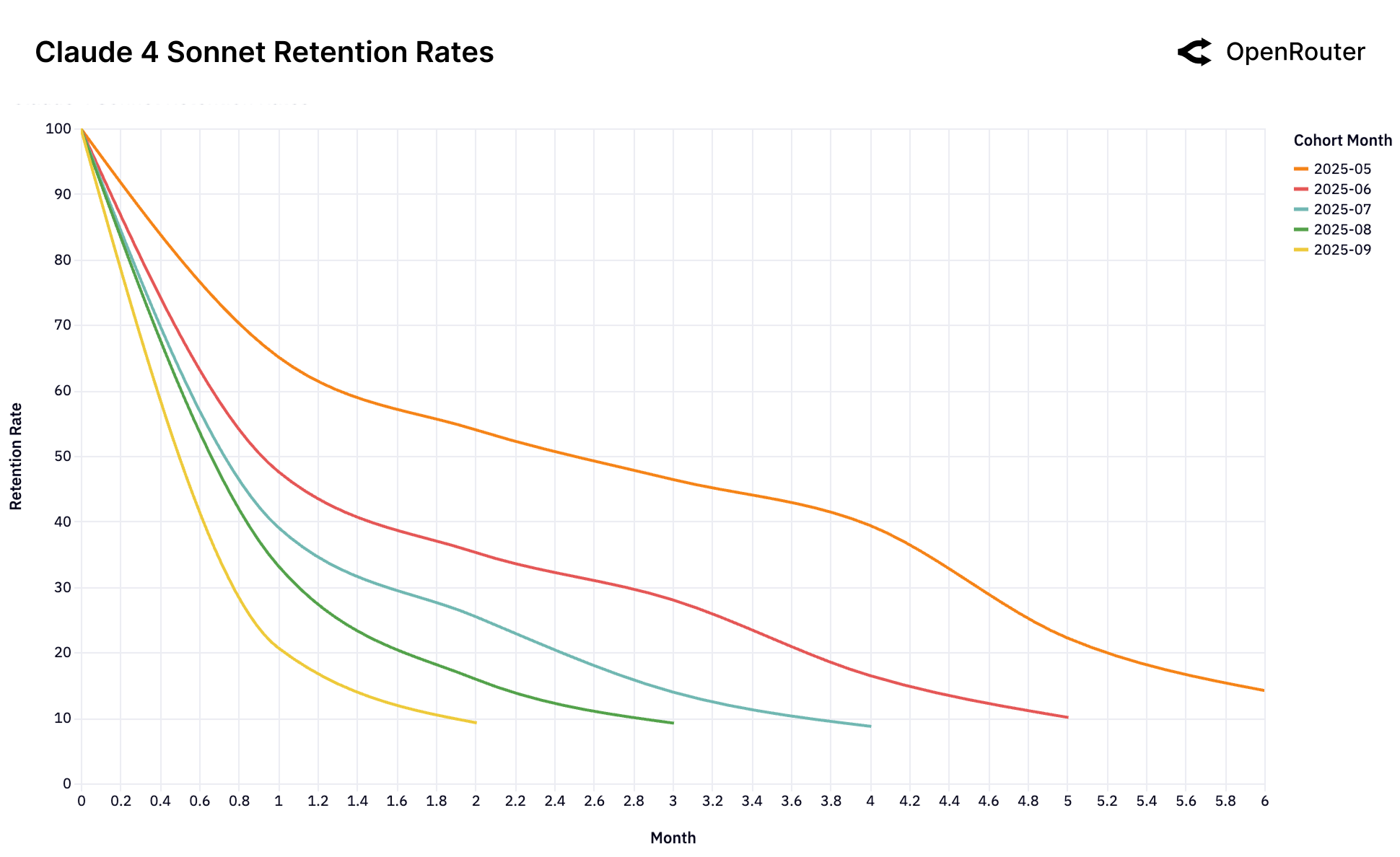}
        \caption{Claude 4 Sonnet}
        \label{fig:claude_4_sonnet}
    \end{subfigure}
    \hfill
    \begin{subfigure}[b]{0.48\textwidth}
        \centering
        \includegraphics[width=\textwidth]{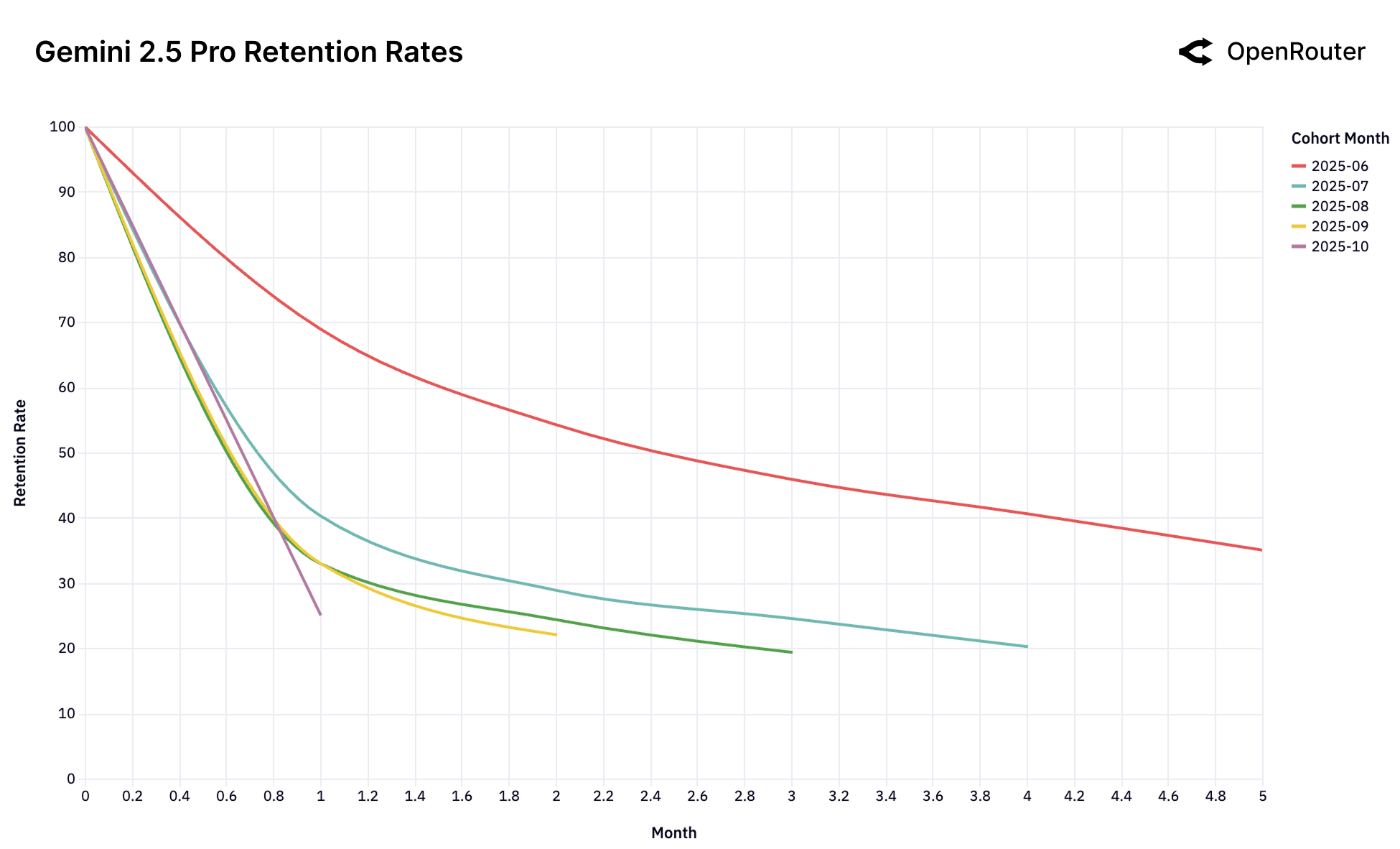}
        \caption{Gemini 2.5 Pro}
        \label{fig:gemini_2_5_pro}
    \end{subfigure}
    
    \begin{subfigure}[b]{0.48\textwidth}
        \centering
        \includegraphics[width=\textwidth]{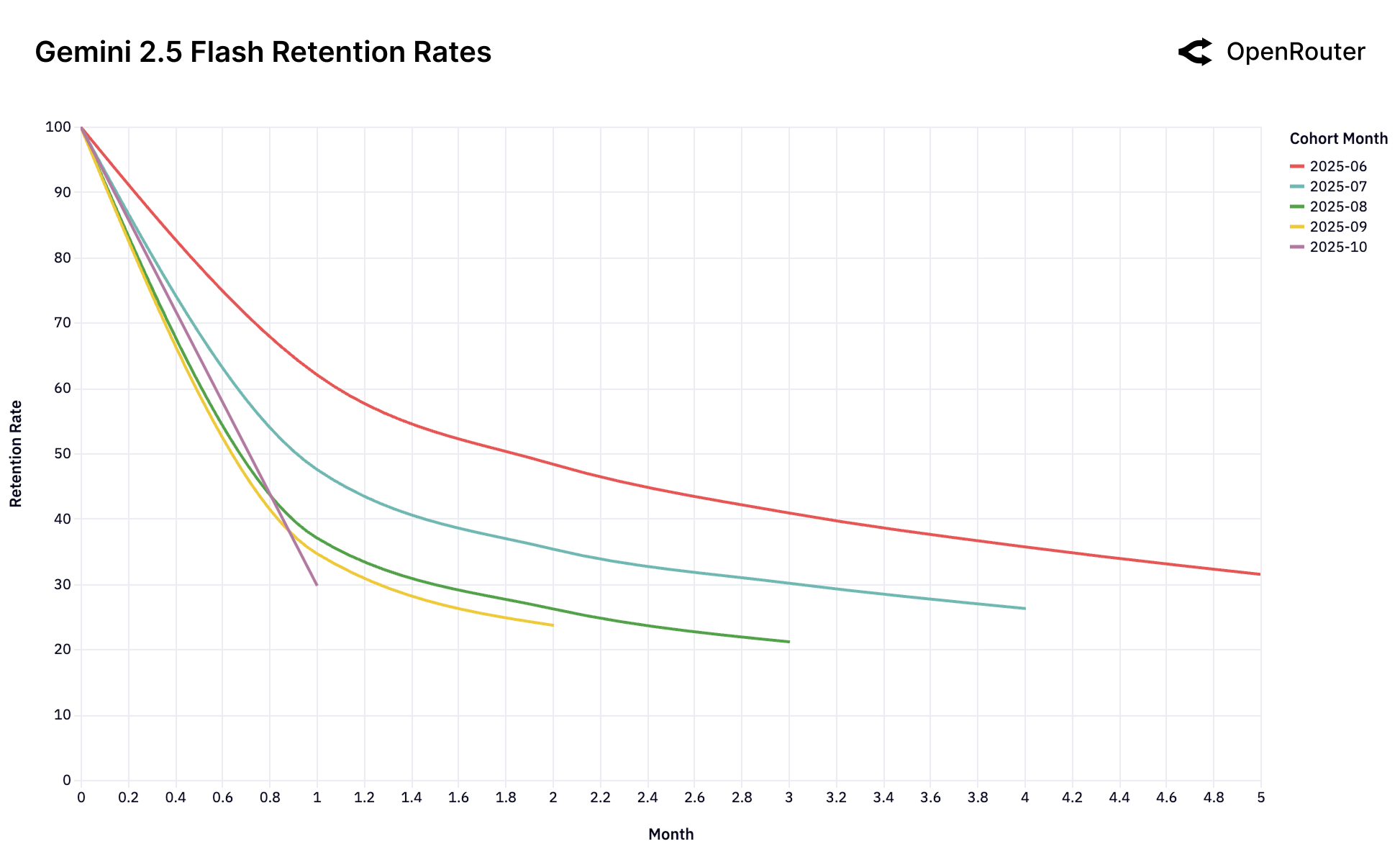}
        \caption{Gemini 2.5 Flash}
        \label{fig:gemini_2_5_flash}
    \end{subfigure}
    \hfill
    \begin{subfigure}[b]{0.48\textwidth}
        \centering
        \includegraphics[width=\textwidth]{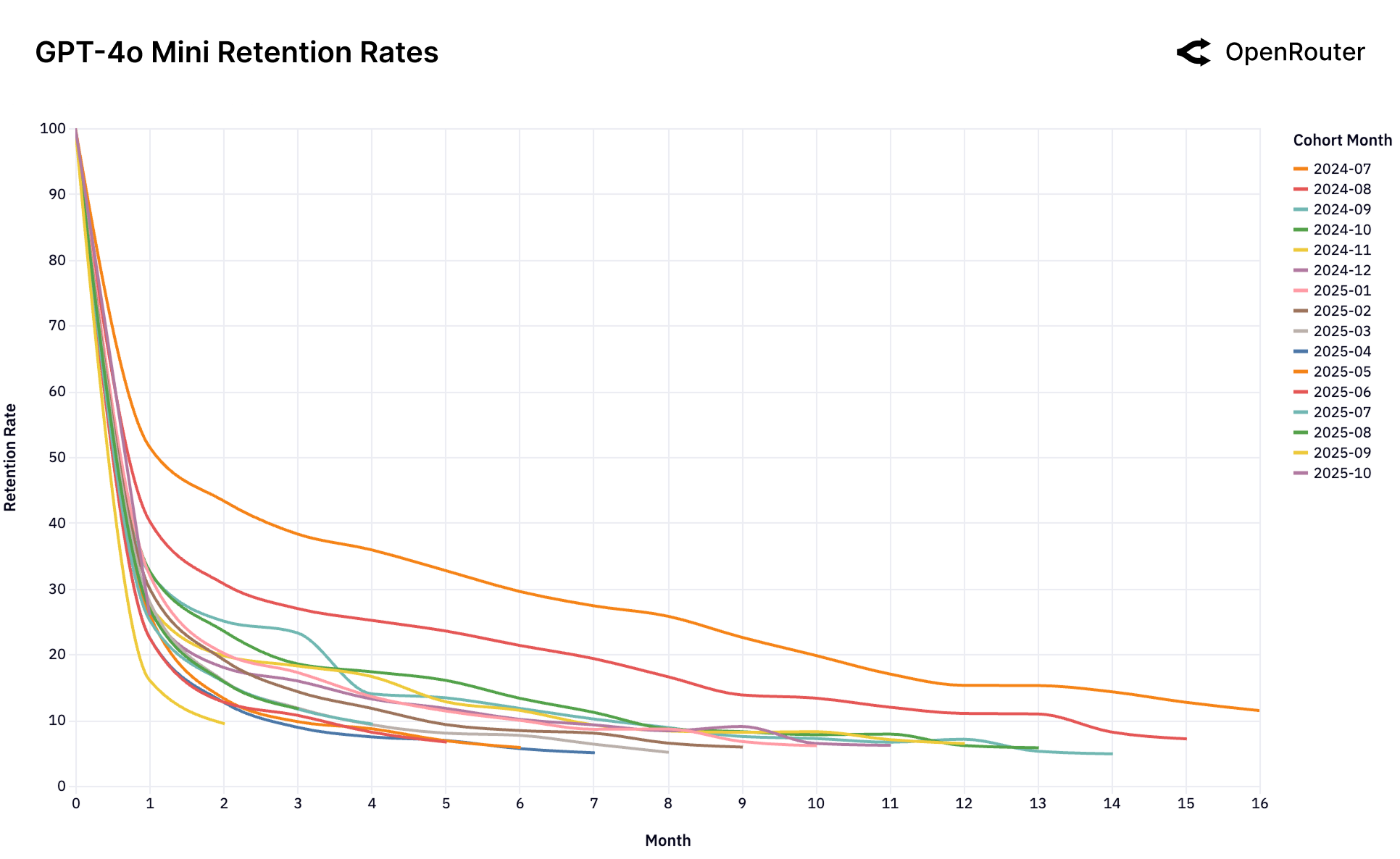}
        \caption{OpenAI GPT-4o Mini}
        \label{fig:gpt_4o_mini}
    \end{subfigure}

    \begin{subfigure}[b]{0.48\textwidth}
        \centering
        \includegraphics[width=\textwidth]{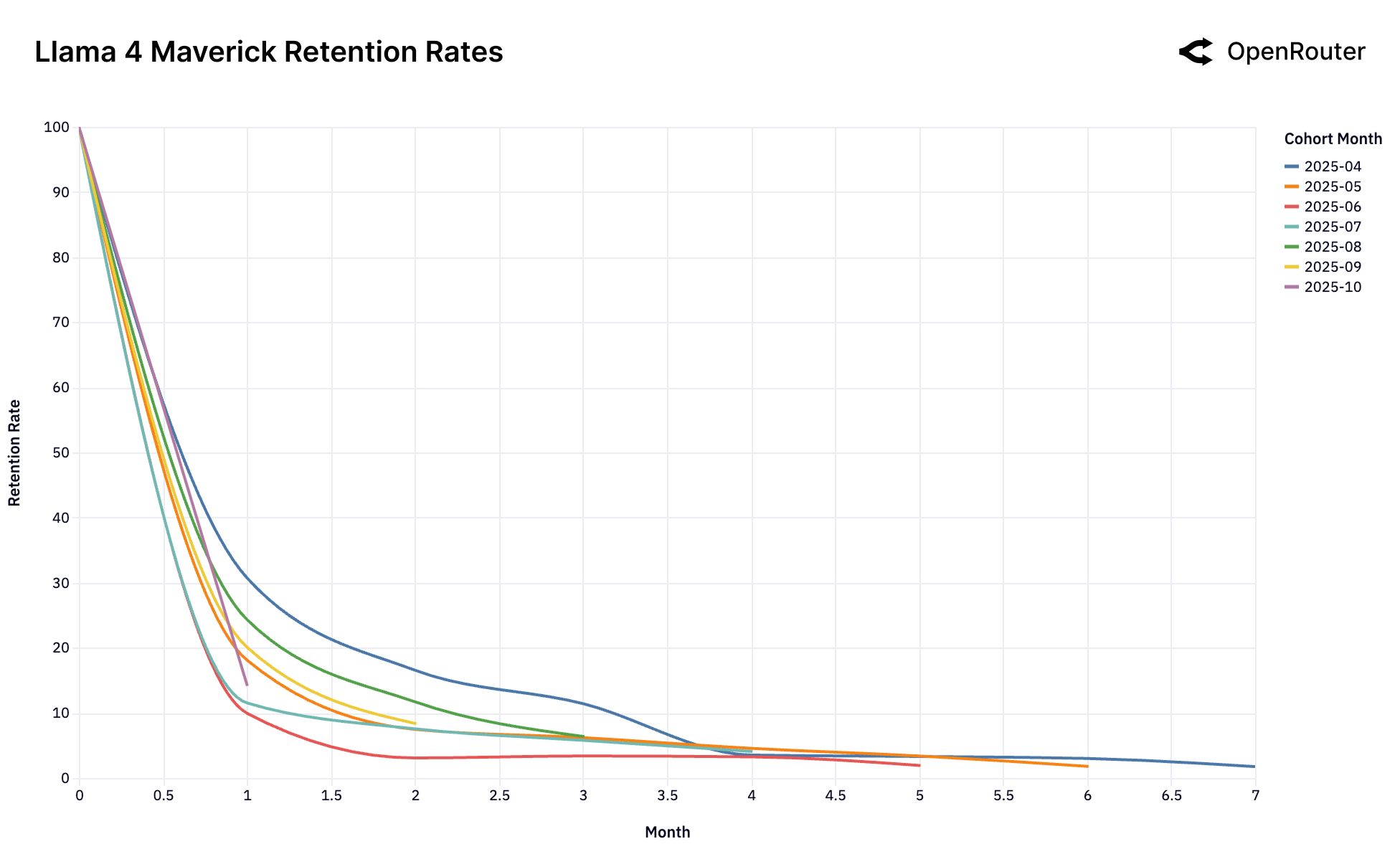}
        \caption{Llama 4 Maverick}
        \label{fig:llama_4_maverick_retention}
    \end{subfigure}
    \hfill
    \begin{subfigure}[b]{0.48\textwidth}
        \centering
        \includegraphics[width=\textwidth]{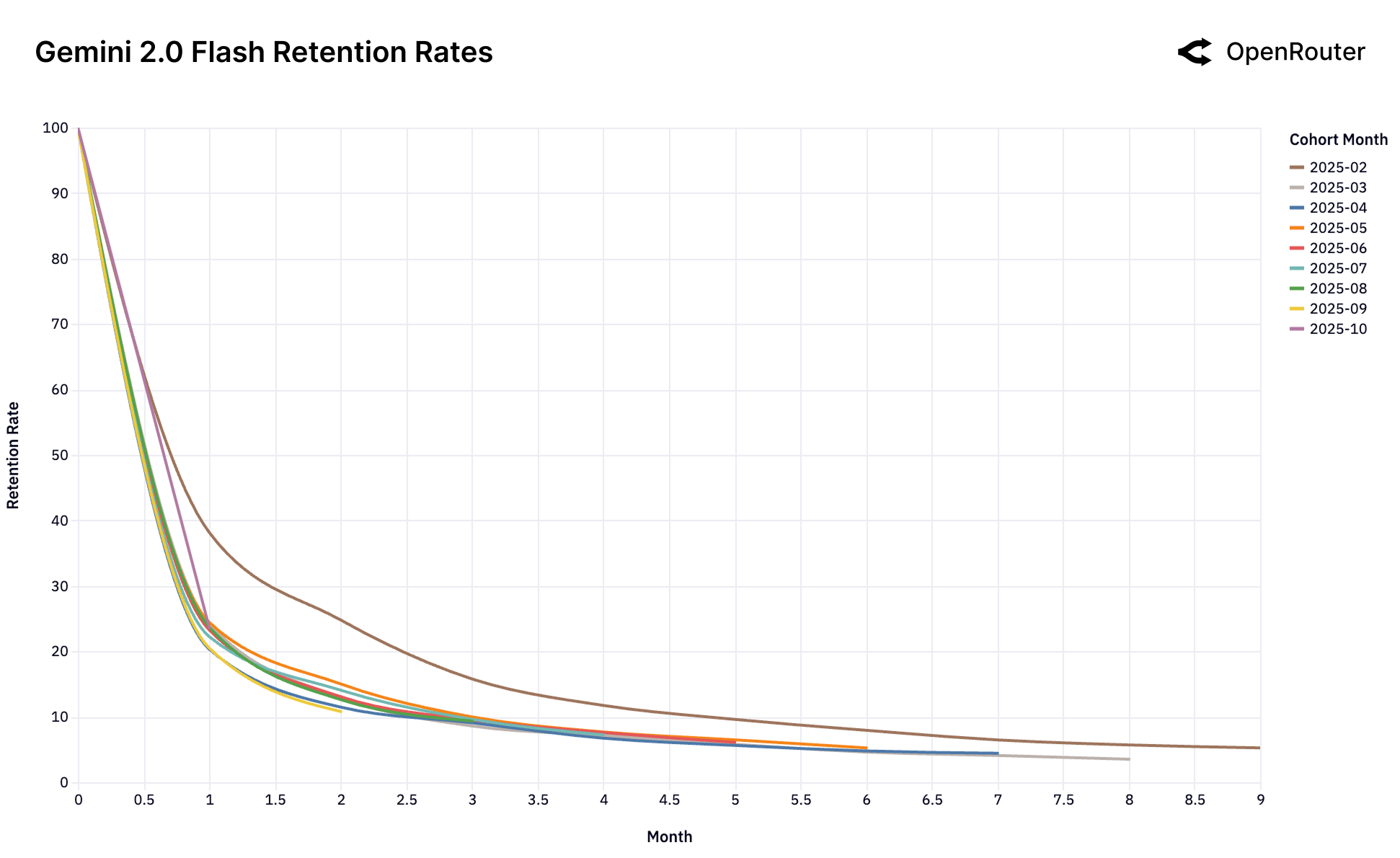}
        \caption{Gemini 2.0 Flash}
        \label{fig:gemini_2_flash}
    \end{subfigure}

    \begin{subfigure}[b]{0.48\textwidth}
        \centering
        \includegraphics[width=\textwidth]{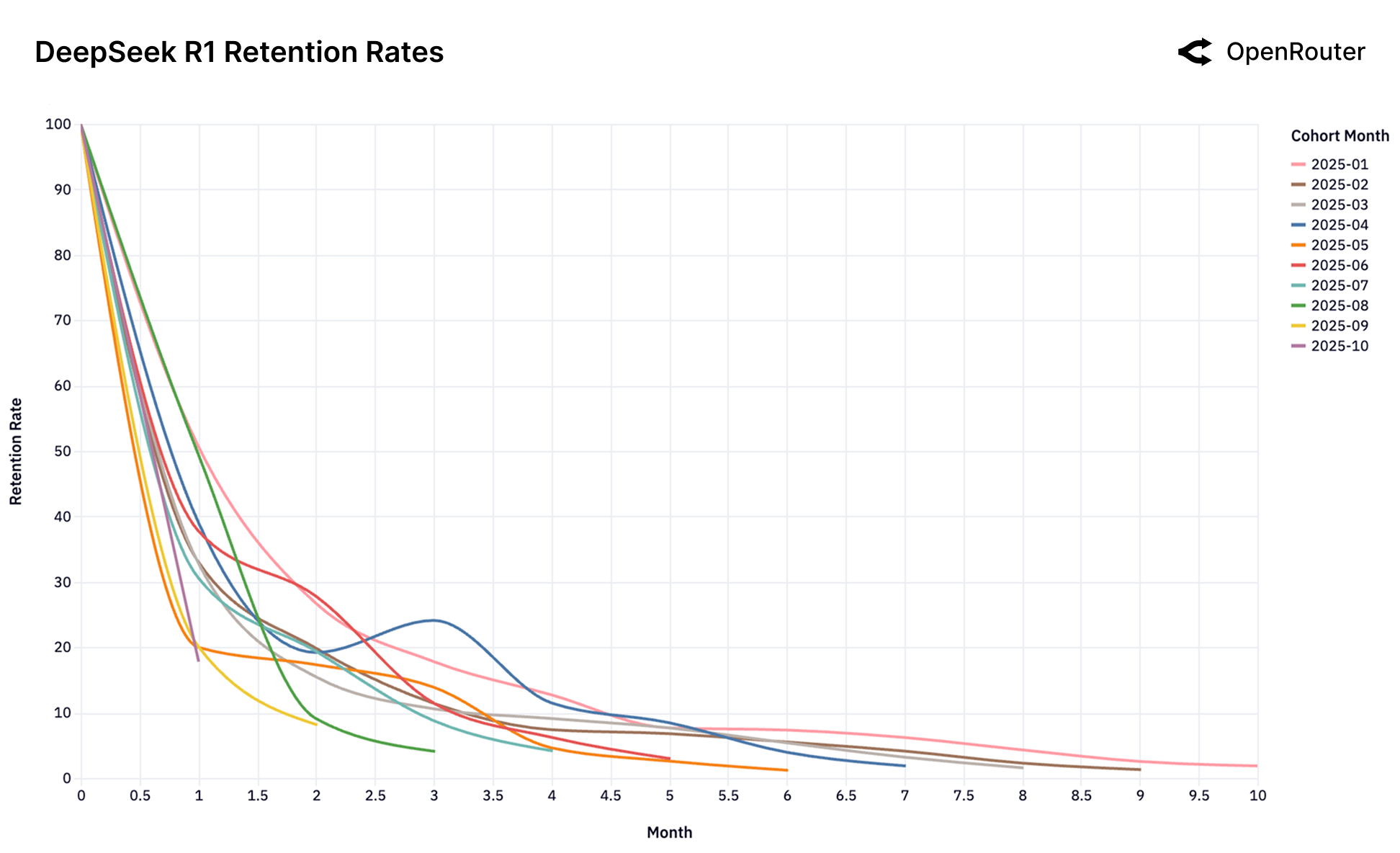}
        \caption{DeepSeek R1}
        \label{fig:deepseek_r1}
    \end{subfigure}
    \hfill
    \begin{subfigure}[b]{0.48\textwidth}
        \centering
        \includegraphics[width=\textwidth]{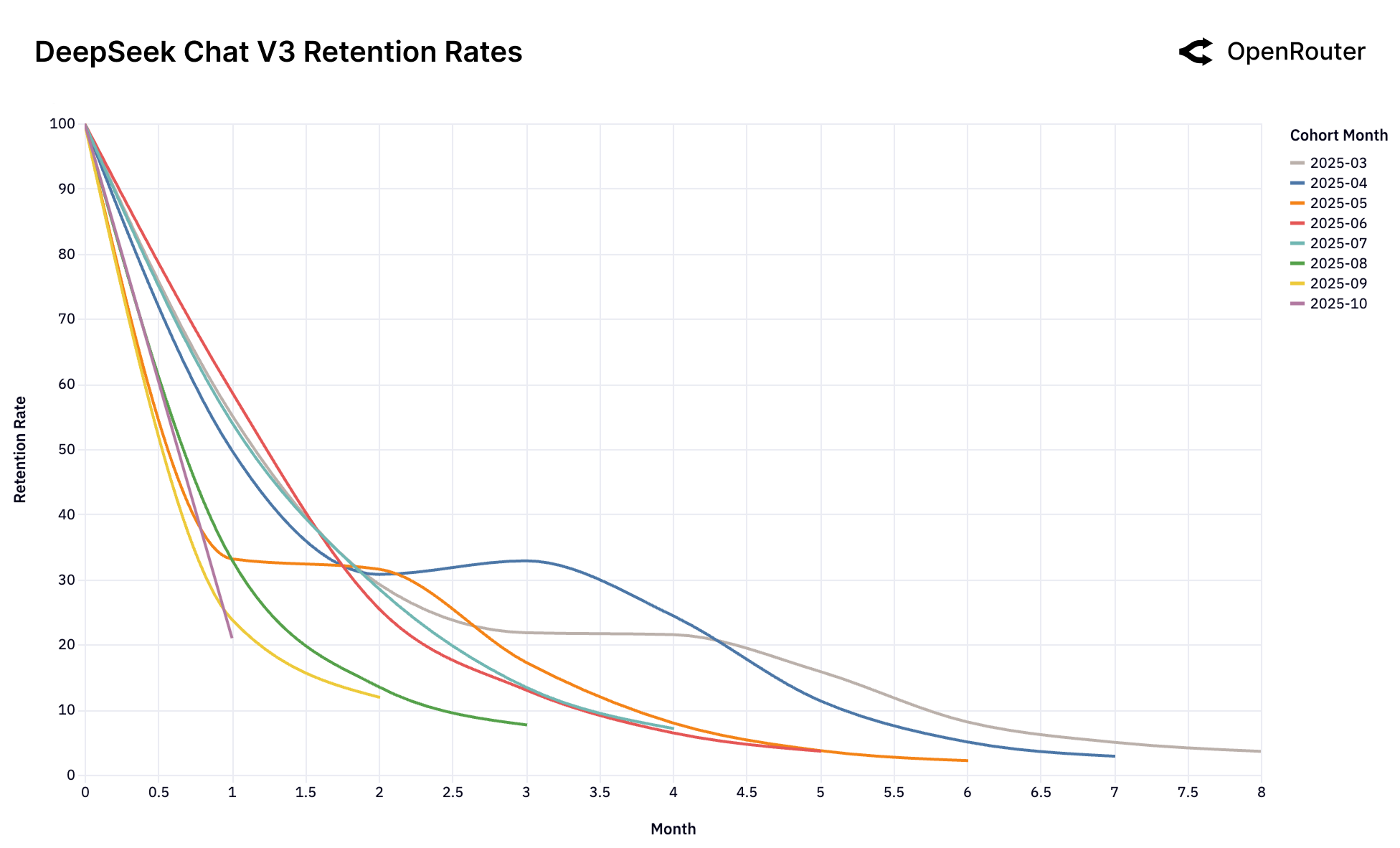}
        \caption{DeepSeek Chat V3-0324}
        \label{fig:deepseek_v3}
    \end{subfigure}
    
    \caption{Cohort Retention Rates. Retention is measured as \textit{activity retention}, where users are counted if they return in subsequent months, even after periods of inactivity; as a result, curves may exhibit small non-monotonic bumps.}
    \label{fig:retention_grid}
\end{figure}

This collection of retention charts (Figure~\ref{fig:retention_grid}) captures the dynamics of the LLM user market across leading models. At first glance, the data is dominated by high churn and rapid cohort decay. Yet beneath this volatility lies a subtler and more consequential signal: a small set of early user cohorts exhibits durable retention over time. We term these \textit{foundational cohorts}. 

These cohorts are not merely early adopters; they represent users whose workloads have achieved a deep and persistent \textit{workload--model fit}. Once established, this fit creates both economic and cognitive inertia that resists substitution, even as newer models emerge.  

We introduce the Cinderella \textbf{Glass Slipper effect} as a framework to describe this phenomenon. The hypothesis posits that in a rapidly evolving AI ecosystem, there exists a latent distribution of high-value workloads that remain unsolved across successive model generations. Each new frontier model is effectively ``tried on'' against these open problems. When a newly released model happens to match a previously unmet technical and economic constraint, it achieves the precise fit --- the metaphorical ``glass slipper.'' 

For the developers or organizations whose workloads finally ``fit,'' this alignment creates strong lock-in effects. Their systems, data pipelines, and user experiences become anchored to the model that solved their problem first. As costs decline and reliability increases, the incentive to re-platform diminishes sharply. Conversely, workloads that do not find such a fit remain exploratory, migrating from one model to another in search of their own solution.

Empirically, this pattern is observable in the June 2025 cohort of \texttt{Gemini 2.5 Pro} (Figure~\ref{fig:gemini_2_5_pro}) and the May 2025 cohort of \texttt{Claude 4 Sonnet} (Figure~\ref{fig:claude_4_sonnet}), which retain approximately 40\% of users at Month 5, substantially higher than later cohorts. These cohorts appear to correspond to specific technical breakthroughs (e.g., reasoning fidelity or tool-use stability) that finally enabled previously impossible workloads.

\begin{itemize}
    \item \textbf{First-to-Solve as Durable Advantage.} 
    The classical first-mover advantage gains significance when a model is the first to \textit{solve} a critical workload. Early adopters embed the model across pipelines, infrastructure, and user behaviors, resulting in high switching friction. This creates a stable equilibrium in which the model retains its foundational cohort even as newer alternatives emerge.

    \item \textbf{Retention as an Indicator of Capability Inflection.} 
    Cohort-level retention patterns serve as empirical signals of model differentiation. Persistent retention in one or more early cohorts indicates a meaningful capability inflection --- a workload class that transitions from infeasible to possible. Absence of such patterns suggests capability parity and limited depth of differentiation.

    \item \textbf{Temporal Constraints of the Frontier Window.} 
    The competitive landscape imposes a narrow temporal window in which a model can capture foundational users. As successive models close the capability gap, the probability of forming new foundational cohorts declines sharply. The ``Cinderella'' moment, where model and workload align precisely, is thus transient but decisive for long-term adoption dynamics.
\end{itemize}

In all, rapid capability shifts in foundation models necessitate a redefinition of user retention. Each new model generation introduces a brief opportunity to solve previously unmet workloads. When such alignment occurs, the affected users form \textit{foundational cohorts}: segments whose retention trajectories remain stable despite subsequent model introductions.

\paragraph{The Dominant Launch Anomaly.} The \texttt{OpenAI GPT-4o Mini} chart shows this phenomenon in its extreme. A single foundational cohort (July 2024, orange line) established a dominant, sticky workload-model fit at launch. All subsequent cohorts, which arrived \textit{after} this fit was established and the market had moved on, behave identically: they churn and cluster at the bottom. This suggests the window to establish this foundational fit is singular and occurs only at the moment a model is perceived as ``frontier."

\paragraph{The Consequence of No-Fit.} The \texttt{Gemini 2.0 Flash} and \texttt{Llama 4 Maverick} charts showcase a cautionary tale of what happens when this initial fit is never established. Unlike the other models, there is no high-performing foundational cohort. Every single cohort performs identically poorly. This suggests that the models were never perceived as a "frontier" for a high-value, sticky workload. It launched directly into the \textit{good enough} market and thus failed to lock in any user base. Similarly, the chaotic charts for \texttt{DeepSeek}, despite overwhelming success overall, struggle to establish a stable, foundational cohort.

\paragraph{Boomerang Effect.}  The DeepSeek models (Figures~\ref{fig:retention_grid}g and \ref{fig:retention_grid}h) introduce a more complex pattern. Their retention curves display a highly unusual anomaly: \textit{resurrection jumps}. Unlike typical, monotonically decreasing retention, several DeepSeek cohorts show a distinct rise in retention after an initial period of churn (e.g., DeepSeek R1's April 2025 cohort around Month 3, and DeepSeek Chat V3-0324's July 2025 cohort around Month 2). This indicates that some churned users are returning to the model. This "boomerang effect" suggests these users return to DeepSeek, after trying alternatives and confirming through competitive testing that DeepSeek provides an optimal, and often better fit for their specific workload due to a potential combination of specialized technical performance, cost-efficiency, or other unique features.

\paragraph{Implications.} The \emph{Glass Slipper} phenomenon reframes retention not as an outcome but as a lens for understanding capability breakthroughs. Foundational cohorts are the fingerprints of real technical progress: they mark where an AI model has crossed from novelty into necessity. For builders and investors alike, identifying these cohorts early may be the single most predictive signal of enduring model--market advantage.

\section{Cost vs. Usage Dynamics}
\label{sec:pricing-usage}

The cost of using a model is a key factor influencing user behavior. In this section, we focus on how different AI workload categories distribute across the cost–usage landscape. By examining where categories cluster on log–log cost vs usage plots, we identify patterns in how workloads concentrate in low-cost, high-volume regions versus high-cost, specialized segments. We also reference similarities to the Jevon's paradox effects, in the sense that lower-cost categories often correspond to higher aggregate usage, though we do not attempt to formally analyze the paradox or causality.

\subsection{Analysis of AI Workload Segmentation by Categories}

\begin{figure}[htbp]
    \centering
    \includegraphics[width=0.9\textwidth]{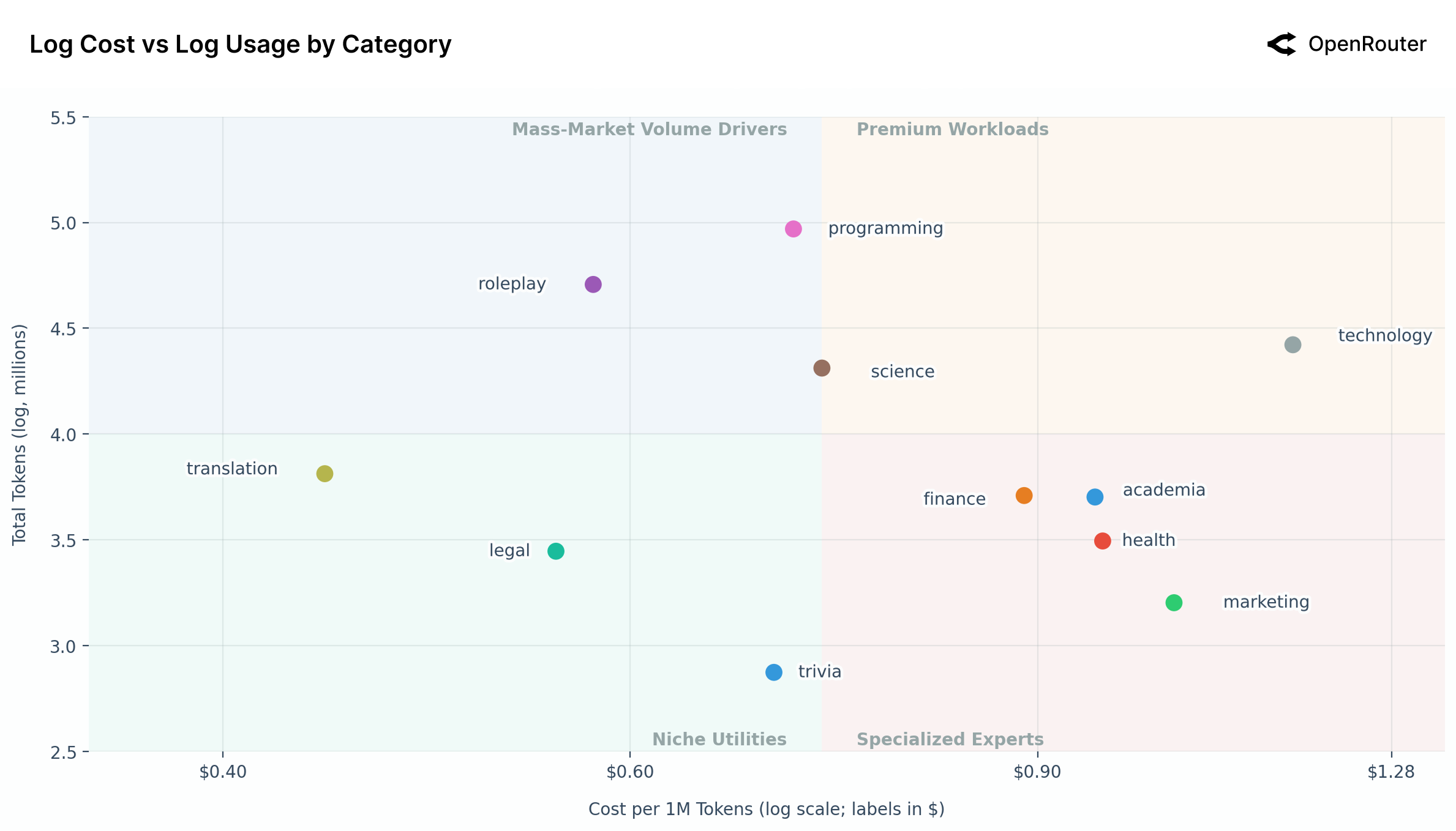}
    \caption{Log Cost vs Log Usage by Category}
    \label{fig:usage_cost}
\end{figure}

The provided scatter plot, shown in Figure~\ref{fig:usage_cost}, reveals a distinct segmentation of AI use cases, mapping them based on their aggregate usage volume (Total Tokens) against their unit cost (Cost per 1M Tokens). A critical preliminary observation is that both axes are logarithmic. This logarithmic scaling signifies that small visual distances on the chart correspond to substantial multiplicative differences in real-world volume and cost.

The chart is bisected by a vertical line at the median cost of \textbf{\$0.73 per 1M tokens}, effectively creating a four-quadrant framework to simplify the AI market across categories.

Note that these end costs differ from advertised list prices. High-frequency workloads benefit from caching, which drives down realized spend and produces materially lower effective prices than those publicly listed. The cost metric shown reflects a blended rate across both prompt and completion tokens, providing a more accurate view of what users actually pay in aggregate. The dataset also excludes BYOK activity to isolate standardized, platform-mediated usage and avoid distortion from custom infrastructure setups.

\paragraph{Premium Workloads (Top-Right):} This quadrant contains high-cost, high-usage applications, now including \texttt{technology} and \texttt{science}, positioned right at the intersection. These represent valuable and heavily-used professional workloads where users are willing to pay a premium for performance or specialized capabilities. \texttt{Technology} is a significant outlier, being dramatically more expensive than any other category. This suggests that \texttt{technology} as a use case (perhaps relating to complex system design or architecture) may require far more powerful and expensive models for inference, yet it maintains a high usage volume, indicating its essential nature.

\paragraph{Mass-Market Volume Drivers (Top-Left):} This quadrant is defined by high usage and a low, at-or-below-average cost. This area is dominated by two massive use cases: \texttt{roleplay}, \texttt{programming} as well as \texttt{science}.
\begin{itemize}
    \item \texttt{Programming} stands out as the "killer professional" category, demonstrating the highest usage volume while having a highly optimized, median cost.
    \item \texttt{Roleplay}'s usage volume is immense, nearly rivaling \texttt{programming}. This is a striking insight: a consumer-facing roleplay application drives a volume of engagement on par with a top-tier professional one.
\end{itemize}
The sheer scale of these two categories confirms that both professional productivity and conversational entertainment are primary, massive drivers for AI. The cost sensitivity in this quadrant is where, as previously noted, open source models have found a significant edge.

\paragraph{Specialized Experts (Bottom-Right):} This quadrant houses lower-volume, high-cost applications, including \texttt{finance}, \texttt{academia}, \texttt{health}, and \texttt{marketing}. These are high-stakes, niche professional domains. The lower aggregate volume is logical, as one might consult an AI for "health" or "finance" far less frequently than for "programming." Users are willing to pay a significant premium for these tasks, likely because the demand for accuracy, reliability, and domain-specific knowledge is extremely high.

\paragraph{Niche Utilities (Bottom-Left):} This quadrant features low-cost, low-volume tasks, including \texttt{translation}, \texttt{legal}, and \texttt{trivia}. These are functional, cost-optimized utilities. \texttt{Translation} has the highest volume within this group, while \texttt{trivia} has the lowest. Their low cost and relatively low volume suggest these tasks may be highly optimized, "solved," or commoditized, where good-enough alternative is available cheaply.

\paragraph{}As noted, the most significant outlier on this chart is \texttt{technology}. It commands the highest cost-per-token by a substantial margin while maintaining high usage. This strongly suggests a market segment with a high willingness-to-pay for high-value, complex answers (e.g., system architecture, advanced technical problem-solving). One key question is whether this high price is driven by high user value (a "demand-side" opportunity) or by a high cost-of-serving (a "supply-side" challenge), as these queries may require the most powerful frontier models. The "play" to be had in \texttt{technology} is to service this high-value market. A provider who can serve this segment, perhaps through highly, optimized, specialist models, could potentially capture a market with higher margins.

\subsection{Effective Cost vs Usage of AI Models.}

\begin{figure}[htbp]
  \centering
  \includegraphics[width=\linewidth]{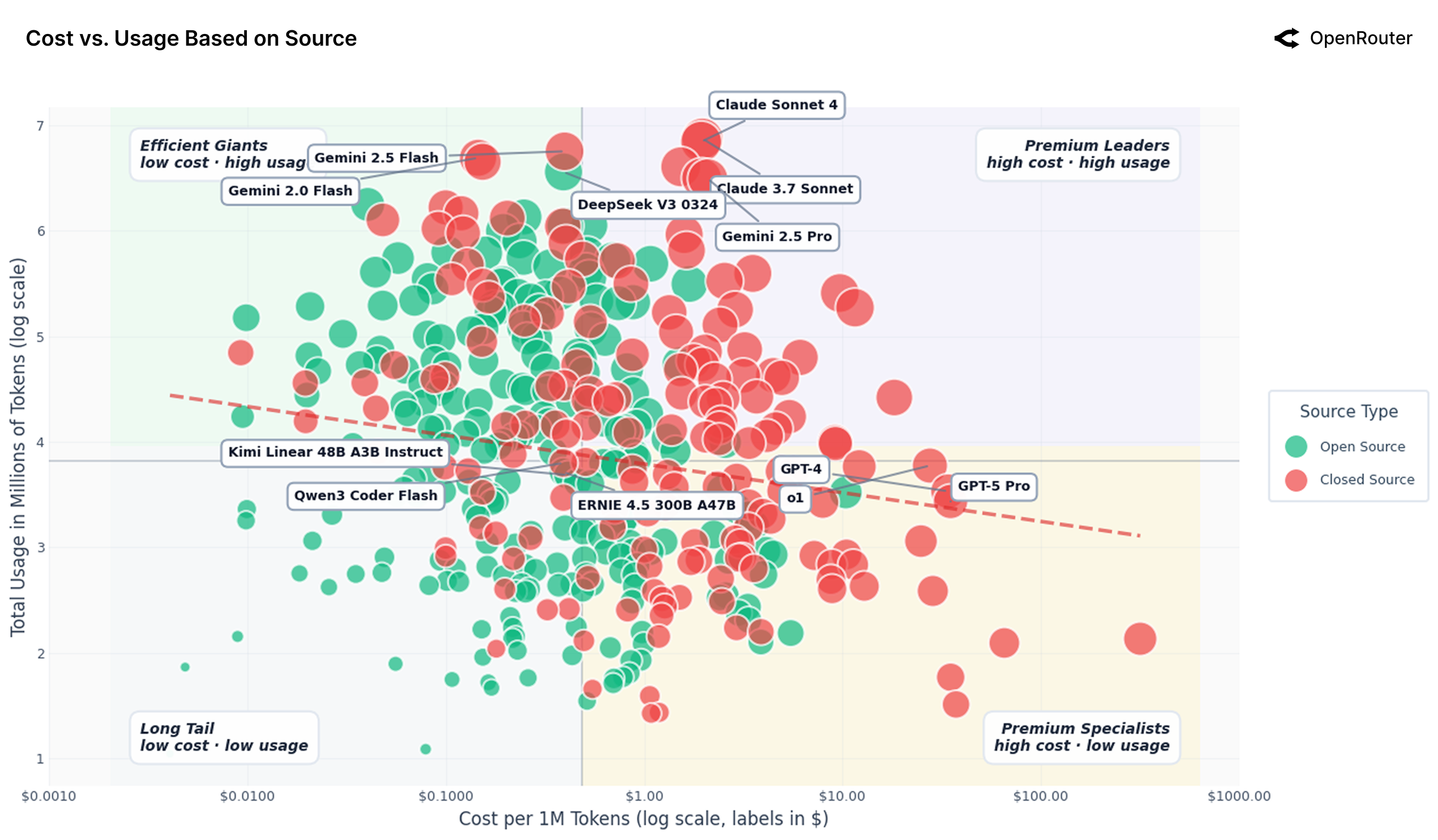}
  \caption{\textbf{Open vs. closed source model landscape: cost vs. usage (log–log scale).} 
  Each point represents a model provided on OpenRouter, colored by source type. 
  Closed source models cluster toward the high-cost, high-usage quadrant, while open source models dominate the low-cost, high-volume region. 
  The dashed trendline is nearly flat, showing limited correlation between cost and total usage. Note: the metric reflects a blended average across prompt and completion tokens, and effective prices are often lower than list rates due to caching. BYOK activity is excluded}
  \label{fig:close-usage-oss-vs-prop}
\end{figure}

\textbf{Figure~\ref{fig:close-usage-oss-vs-prop}} maps model usage against cost per 1M tokens (log–log scale), revealing weak overall correlation. The x-axis maps out the nominal values for convenience. The trendline is nearly flat, indicating that demand is relatively price-inelastic; a 10\% decrease in price corresponds to only about a 0.5–0.7\% increase in usage. Yet the dispersion across the chart is substantial, reflecting strong market segmentation. 
Two distinct regimes appear: proprietary models from OpenAI and Anthropic occupy the high-cost, high-usage zone, while open models like DeepSeek, Mistral, and Qwen populate the low-cost, high-volume zone. 
This pattern supports a simple heuristic: \textbf{closed source models capture high value tasks, while open source models capture high volume lower value tasks.} The weak price elasticity indicates that even drastic cost differences do not fully shift demand; proprietary providers retain pricing power for mission-critical applications, while open ecosystems absorb volume from cost-sensitive users.

\begin{figure}[htbp]
  \centering
  \includegraphics[width=\linewidth]{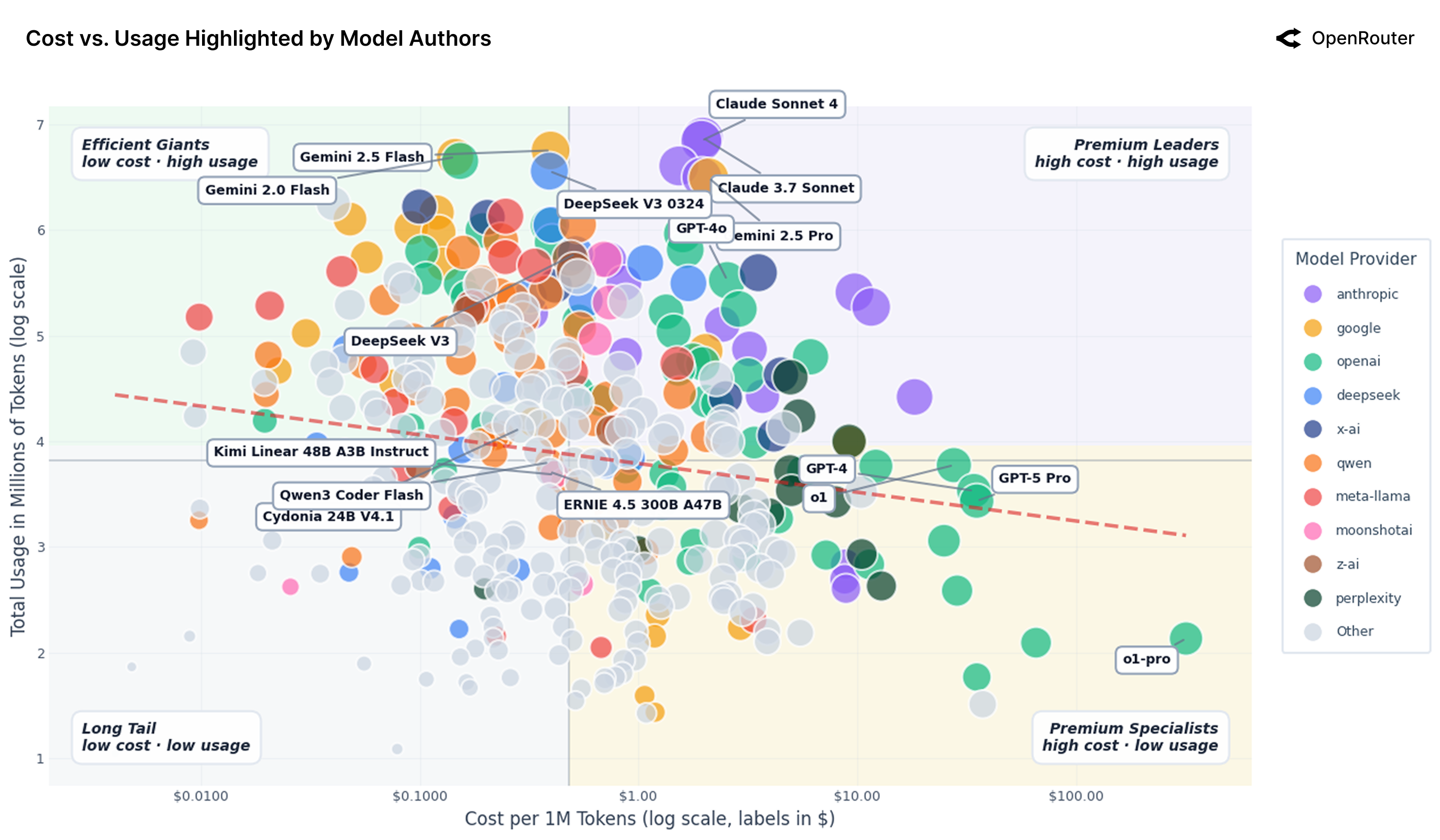}
  \caption{\textbf{AI model market map: cost vs. usage (log–log scale).} Similar to the above graph but each point is colored by model provider.}
  \label{fig:cost-usage}
\end{figure}

\begin{table}[htbp]
\centering
\small
\setlength{\tabcolsep}{3pt}       
\renewcommand{\arraystretch}{1.1} 
\begin{tabular}{l l r r p{0.3\linewidth}}
\toprule
\textbf{Segment} & \textbf{Model} & \textbf{Price per 1M} & \textbf{Usage (log)} & \textbf{Takeaway} \\
\midrule
Efficient giants & google/gemini-2.0-flash & \$0.147 & 6.68 & Low price and strong distribution make it a default high-volume workhorse \\
Efficient giants & deepseek/deepseek-v3-0324 & \$0.394 & 6.55 & Competitive quality at bargain cost drives massive adoption \\
Premium leaders & anthropic/claude-3.7-sonnet & \$1.963 & 6.87 & Very high usage despite premium price, signaling preference for quality and reliability \\
Premium leaders & anthropic/claude-sonnet-4 & \$1.937 & 6.84 & Enterprise workloads appear price-inelastic for trusted frontier models \\
Long tail & qwen/qwen-2-7b-instruct & \$0.052 & 2.91 & Rock-bottom pricing but limited reach, likely due to weaker model-market fit \\
Long tail & ibm/granite-4.0-micro & \$0.036 & 2.95 & Cheap yet niche, used mainly in limited settings \\
Premium specialists & openai/gpt-4 & \$34.068 & 3.53 & High cost and moderate usage, reserved for the most demanding tasks \\
Premium specialists & openai/gpt-5-pro & \$34.965 & 3.42 & Ultra-premium model with focused, high-stakes workloads. Still early in adoption given recent release. \\
\bottomrule
\end{tabular}
\caption{\textbf{Example models by segment.} Values sampled from the updated dataset. The market-level regression remains nearly flat, yet segment-level behavior differs sharply.}
\label{tab:segments}
\end{table}

\paragraph{Now let's zoom in on specific model authors within the same map.}
\textbf{Figure~\ref{fig:cost-usage}} is similar to the prior figure but displays the model authors. Four usage–cost archetypes emerge. \textit{Premium leaders}, such as Anthropic's Claude 3.7 Sonnet and Claude Sonnet 4, command costs around \$2 per 1M tokens and still reach high usage, suggesting users are willing to pay for superior reasoning and reliability at scale. \textit{Efficient giants}, like Google’s Gemini 2.0 Flash and DeepSeek V3 0324, pair strong performance with prices below \$0.40 per 1M tokens and achieve similar usage levels, making them attractive defaults for high-volume or long-context workloads. \textit{Long tail} models, including Qwen 2 7B Instruct and IBM Granite 4.0 Micro, are priced at just a few cents per 1M tokens yet sit around $10^{2.9}$ in total usage, reflecting constraints from weaker performance, limited visibility, or fewer integrations. Finally, \textit{premium specialists}, such as OpenAI’s GPT-4 and GPT-5 Pro, occupy the high-cost, low-usage quadrant: at roughly \$35 per 1M tokens and usage near $10^{3.4}$, they are used sparingly for niche, high-stakes workloads where output quality matters far more than marginal token cost.

Overall, the scatterplot highlights that pricing power in the LLM market is not uniform. While cheaper models can drive scale through efficiency and integration, premium offerings still command strong demand where stakes are high. This fragmentation suggests that the market has not yet commoditized, and that differentiation, whether through latency, context length, or output quality, remains a source of strategic advantage.

These observations suggest the following:
\begin{itemize}
    \item At a macro level, \textbf{demand is inelastic}, but this masks different micro behaviors. Enterprises with mission-critical tasks will pay high prices (so these models see high usage). On the other hand, hobbyists and dev pipelines are very cost-sensitive and flock to cheaper models (leading to large usage for efficient models).
    \item There is some evidence of \textbf{Jevons Paradox}: making some models very cheap (and fast) led to people using them for far more tasks, ultimately consuming more total tokens. We see this in the efficient giants group: as cost per token dropped, those models got integrated everywhere and total consumption soared (people run longer contexts, more iterations, etc.).
    \item  \textbf{Quality and capabilities often trump cost:} The heavy usage of expensive models (Claude Sonnet series, GPT-4) indicates that if a model is significantly better or has a trust advantage, users will bear higher costs. Often these models are integrated in workflows where the cost is negligible relative to the value of what they produce (e.g., code that saves an hour of developer time is worth far more than a few dollars of API calls).
    \item Conversely, simply being cheap isn’t enough, a model must also be \textbf{differentiable and sufficiently capable}. Many open models priced near zero still because they are just good enough but dont find a \textit{workload-model fit} or are not quite reliable, so developers hesitate to integrate them deeply.
\end{itemize}

From an operator’s standpoint, several strategic patterns emerge. Providers like Google have leaned heavily into tiered offerings (most notably with Gemini Flash and Pro) explicitly trading off speed, cost, and capability. This tiering enables market segmentation by price sensitivity and task criticality: lightweight tasks are routed to cheaper, faster models; premium models serve complex or latency-tolerant workloads. Optimizing for use cases and reliability is often as impactful as "cutting" price. A faster, purpose-built model may be preferred over a cheaper but unpredictable one, especially in production settings. This shifts focus from cost-per-token to cost-per-successful-outcome. \textbf{The relatively flat demand elasticity suggests LLMs are not yet a commodity—many users are willing to pay a premium for quality, capabilities, or stability.} Differentiation still holds value, particularly when task outcomes matter more than marginal token savings.

\section{Discussion}
This empirical study offers a data-driven perspective on how LLMs are actually being used, highlighting several themes that nuance the conventional wisdom about AI deployment:

\textbf{1. A Multi-Model Ecosystem.} Our analysis shows that no single model dominates all usage. Instead, we observe a rich \textbf{multi-model ecosystem} with both closed and open models capturing significant shares. For example, even though OpenAI and Anthropic models lead in many programming and knowledge tasks, open source models like DeepSeek and Qwen collectively served a large portion of total tokens (sometimes over 30\%). This suggests the future of LLM usage is likely model-agnostic and heterogeneous. For developers, this means maintaining flexibility, integrating multiple models and choosing the best for each job, rather than betting everything on one model’s supremacy. For model providers, it underscores that competition can come from unexpected places (e.g., a community model might erode part of your market unless you continuously improve and differentiate).

\textbf{2. Usage Diversity Beyond Productivity.} A surprising finding is the sheer volume of \textit{roleplay and entertainment-oriented usage}. Over half of open source model usage was for roleplay and storytelling. Even on proprietary platforms, a non-trivial fraction of early ChatGPT use was casual and creative before professional use cases grew. This counters an assumption that LLMs are mostly used for writing code, emails, or summaries. In reality, many users engage with these models for companionship or exploration. This has important implications. It highlights a substantial opportunity for consumer-facing applications that merge narrative design, emotional engagement, and interactivity. It suggests new frontiers for personalization—agents that evolve personalities, remember preferences, or sustain long-form interactions. It also redefines model evaluation metrics: success may depend less on factual accuracy and more on consistency, coherence, and the ability to sustain engaging dialog. Finally, it opens a pathway for crossovers between AI and entertainment IP, with potential in interactive storytelling, gaming, and creator-driven virtual characters.

\textbf{3. Agents vs Humans: The Rise of Agentic Inference.}  
LLM usage is shifting from single-turn interactions to \textit{agentic inference}, where models plan, reason, and execute across multiple steps. Rather than producing one-off responses, they now coordinate tool calls, access external data, and iteratively refine outputs to achieve a goal. Early evidence shows rising multi-step queries and chained tool use that we proxy to agentic use. As this paradigm expands, evaluation will move from language quality to task completion and efficiency. The next competitive frontier is how effectively models can \textit{perform sustained reasoning}, a shift that may ultimately redefine what agentic inference at scale means in practice.

\textbf{4. Geographic Outlook.}  LLM usage is becoming increasingly \textit{global and decentralized}, with rapid growth beyond North America. Asia’s share of total token demand has risen from about 13\% to 31\%, reflecting stronger enterprise adoption and innovation. Meanwhile, \textit{China has emerged as a major force}, not only through domestic consumption but also by producing globally competitive models. The broader takeaway: \textit{LLMs must be globally useful} performing well across languages, contexts, and markets. The next phase of competition will hinge on cultural adaptability and multilingual capability, not just model scale.

\textbf{5. Cost vs. Usage Dynamics.}  
The LLM market does not seem to behave like a commodity just yet: price alone explains little about usage. Users balance cost with reasoning quality, reliability, and breadth of capability. Closed models continue to capture high-value, revenue-linked workloads, while open models dominate lower-cost and high-volume tasks. This creates a dynamic equilibrium—one defined less by stability and more by constant pressure from below.  Open source models continuously push the \textit{efficient frontier}, especially in reasoning and coding domains (e.g. Kimi K2 Thinking) where rapid iteration and OSS innovations narrow the performance gap. Each improvement in open models compresses the pricing power of proprietary systems, forcing them to justify premiums through superior integration, consistency, and enterprise support. The resulting competition is fast-moving, asymmetric, and continuously shifting. Over time, as quality convergence accelerates, \textit{price elasticity is likely to increase}, turning what was once a differentiated market into a more fluid one.

\textbf{6. Retention and the Cinderella Glass Slipper Phenomenon.} As foundation models advance in leaps, not steps, retention has become the true measure of defensibility. Each breakthrough creates a fleeting launch window where a model can “fit” a high-value workload perfectly (the Cinderella Glass Slipper moment) and once users find that fit, they stay. In this paradigm, product-market fit equals workload-model fit: being the first to solve a real pain point drives deep, sticky adoption as users build workflows and habits around that capability. Switching then becomes costly, both technically and behaviorally. For builders and investors, the signal to watch isn’t growth but retention curves, namely, the formation of foundational cohorts who stay through model updates. In an increasingly fast-moving market, capturing these important unmet needs early determines who endures after the next capability leap.

\paragraph{}Together, LLMs are becoming an essential computational substrate for reasoning-like tasks across domains, from programming to creative writing. As models continue to advance and deployment expands, having accurate insights on real-world usage dynamics will be crucial for making informed decisions. Ways in which people use LLMs do not always align with expectations and vary significantly country by country, state by state, use case by use case. By observing usage at scale, we can ground our understanding of LLM impact in reality, ensuring that subsequent developments, be they technical improvements, product features, or regulations, are aligned with actual usage patterns and needs. We hope this work serves as a foundation for more empirical studies and that it encourages the AI community to continuously measure and learn from real-world usage as we build the next generation of frontier models.

\section{Limitations} 
This study reflects patterns observed on a single platform, namely OpenRouter, and over a finite time window, offering only a partial view of the broader ecosystem. Certain dimensions, such as enterprise usage, locally hosted deployments, or closed internal systems, remain outside the scope of our data. Moreover, several of our data analyses rely on \textit{proxy measures}: for instance, identifying agentic inference through multi-step or tool-invocation calls, or inferring user geography from billing rather than verified location data. As such, the results should be interpreted as indicative behavioral patterns rather than definitive measurements of underlying phenomena.

\section{Conclusion}

This study offers an empirical view of how large language models are becoming embedded in the world’s computational infrastructure. They are now integral to workflows, applications, and agentic systems, transforming how information is generated, mediated, and consumed.

The past year catalyzed a step change in how the field conceives \textit{reasoning}. The emergence of \textit{o1}-class models normalized extended deliberation and tool use, shifting evaluation beyond single-shot benchmarks toward process-based metrics, latency-cost tradeoffs, and success-on-task under orchestration. Reasoning has become a measure of how effectively models can plan and verify to deliver more reliable outcomes.

The data show that the LLM ecosystem is structurally plural. No single model or provider dominates; instead, users select systems along multiple axes such as capability, latency, price, and trust depending on context. This heterogeneity is not a transient phase but a fundamental property of the market. It promotes rapid iteration and reduces systemic dependence on any one model or stack.

Inference itself is also changing. The rise of multi-step and tool-linked interactions signals a shift from static completion to dynamic orchestration. Users are chaining models, APIs, and tools to accomplish compound objectives, giving rise to what can be described as \textit{agentic inference}. There are many reasons to believe that agentic inference will exceed, if it hasn't already, human inference.

Geographically, the landscape is becoming more distributed. Asian share of usage continues to expand, China specifically has emerged as both a model developer and exporter, illustrated by the rise of players like Moonshot AI, DeepSeek, and Qwen. The success of non-Western open-weight models shows that LLMs are truly global computational resource.

In effect, \textit{o1} did not end competition. Far from that. It expanded the design space. The field is moving toward systems thinking instead of monolithic bets, toward instrumentation instead of intuition, and toward empirical usage analytics instead of leaderboard deltas. If the past year demonstrated that agentic inference is viable at scale, the next will focus on operational excellence: measuring real task completion, reducing variance under distribution shifts, and aligning model behavior with the practical demands of production-scale workloads.

\section*{References}

\makeatletter
\renewcommand\@biblabel[1]{} 
\renewenvironment{thebibliography}[1]
     {\list{\@biblabel{\@arabic\c@enumiv}}%
           {\settowidth\labelwidth{\@biblabel{#1}}%
            \leftmargin\labelwidth
            \advance\leftmargin\labelsep
            \itemindent-\leftmargin
            \usecounter{enumiv}%
            \let\p@enumiv\@empty
            \renewcommand\theenumiv{\@arabic\c@enumiv}}%
      \sloppy\clubpenalty4000\widowpenalty4000}
     {\endlist}
\makeatother

\section*{Contributions}

This work was made possible by the foundational platform, infrastructure, datasets, and technical vision developed by the OpenRouter team. In particular, Alex Atallah, Chris Clark, Louis Vichy provided the engineering groundwork and architectural direction that enabled the explorations undertaken in this study. Justin Summerville contributed fundamental support across implementation, testing, and experimental refinement. Additional contributions included launch support from Natwar Maheshwari and design edits from Julian Thayn.

Malika Aubakirova (a16z) served as the lead author, responsible for experiment design, implementation, data analysis, and full preparation of the paper. Anjney Midha provided strategic guidance and shaped the overarching framing and direction.

Early exploratory experimentation and system setup were supported by Abhi Desai during his internship at a16z. Rajko Radovanovic and Tyler Burkett provided targeted technical insights and practical assistance that strengthened several critical components of the work.

All contributors participated in discussions, provided feedback, and reviewed the final manuscript.

\section*{Appendix}

\subsection*{Category Sub-Composition Details}
\begin{figure*}[htbp]
    \centering
    \begin{subfigure}[t]{0.8\textwidth}
        \includegraphics[width=\linewidth]{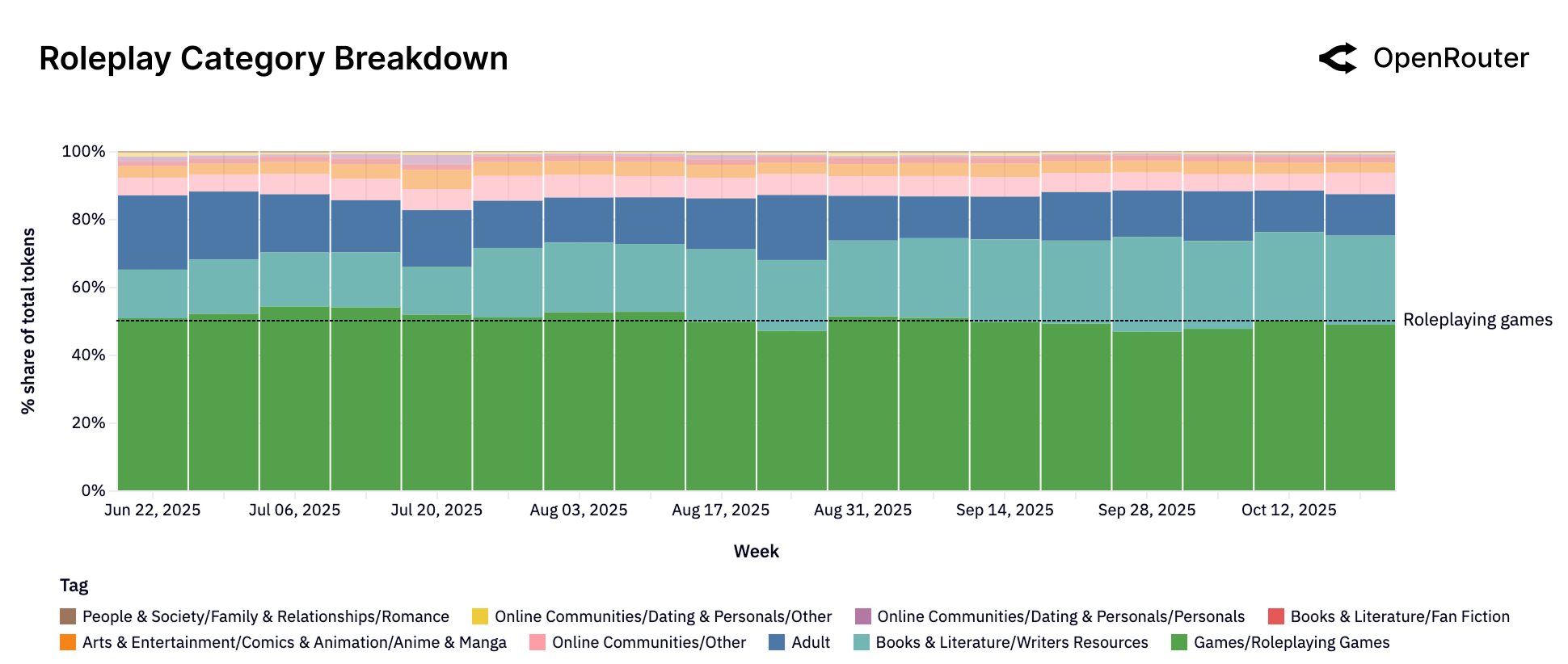}
        \caption{\textbf{Roleplay (sub-tags).} Tokens partition into \textit{Role-Playing Game} scenarios (58\%) and other creative dialogue (persona chat, narrative co-writing, etc.).}
        \label{fig:roleplay_category_breakdown}
    \end{subfigure}\hfill
    \begin{subfigure}[t]{0.80\textwidth}
        \includegraphics[width=\linewidth]{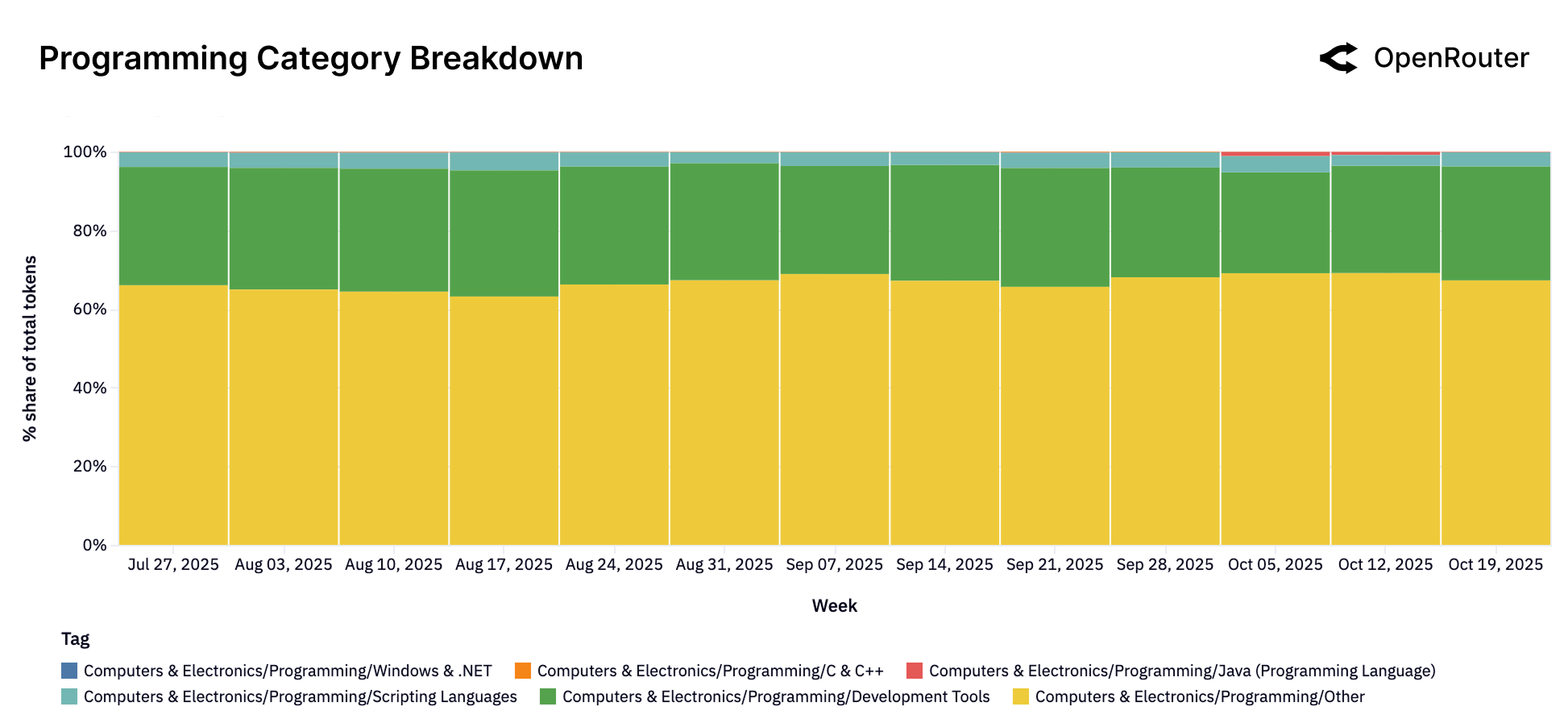}
        \caption{\textbf{Programming (sub-tags).} General coding tasks form the majority (no single specific domain dominates), with smaller shares for web dev, data science, etc., indicating broad use across programming topics.}
        \label{fig:programming_category_breakdown}
    \end{subfigure}    
    \begin{subfigure}[t]{0.79\textwidth}
        \includegraphics[width=\linewidth]{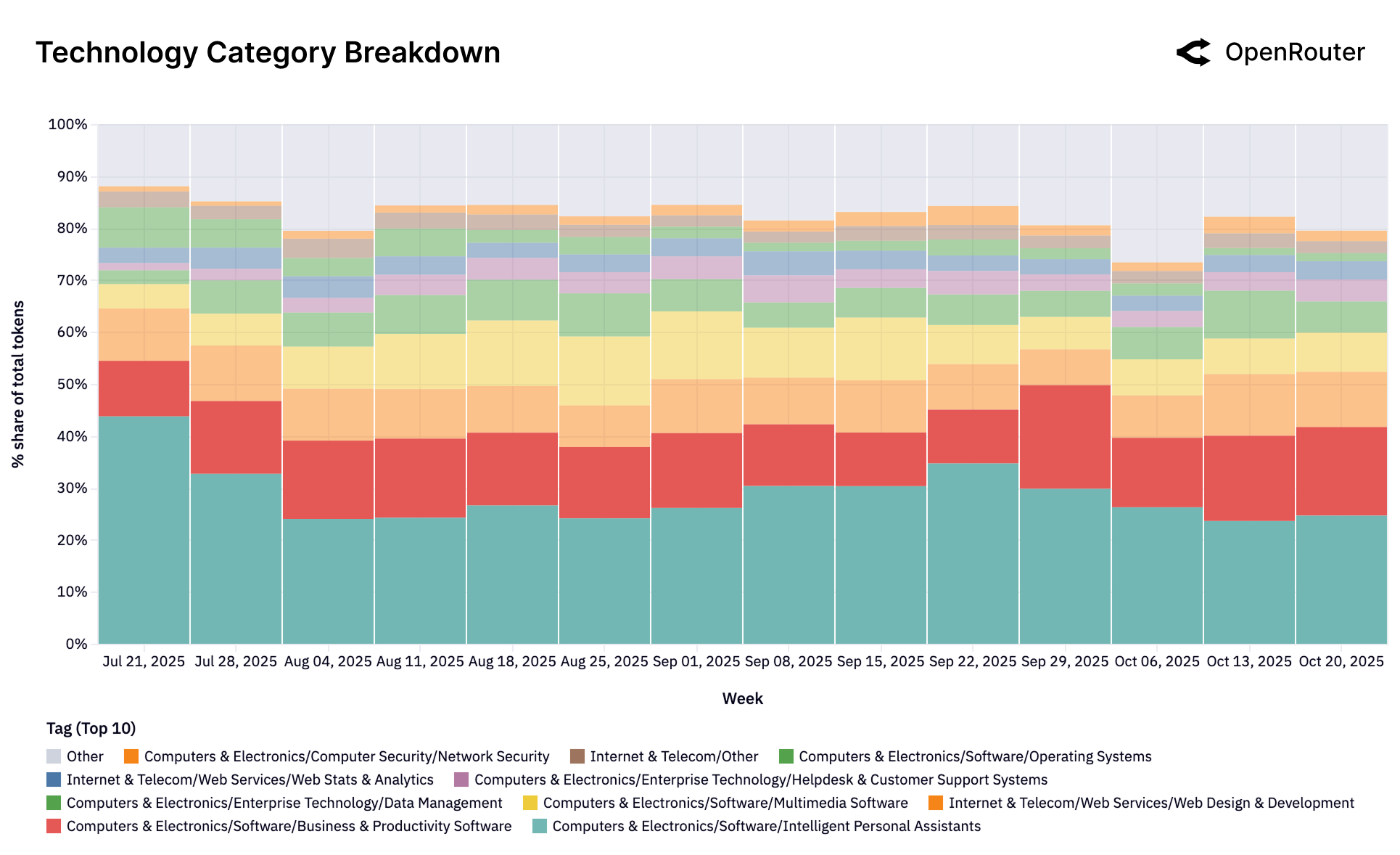}
        \caption{\textbf{Technology (sub-tags).} Dominated by \textit{Intelligent Assistants} and \textit{Productivity Software} use cases (combined ~65\%), followed by IT support and consumer electronics queries.}
        \label{fig:technology_category_breakdown}
    \end{subfigure}\hfill
    \caption{\textbf{Category sub-composition for major domains.} All three domains (roleplay, technology, programming) exhibit distinct internal patterns.}
    \label{fig:category_breakdowns_combined}
\end{figure*}

\clearpage


\begin{thebibliography}{99}

\bibitem{anthropic2025}
R. Appel, J. Zhao, C. Noll, O. K. Cheche, and W. E. Brown Jr.
Anthropic economic index report: Uneven geographic and enterprise AI adoption.
\textit{arXiv preprint arXiv:2511.15080}, 2025.
URL \url{https://arxiv.org/abs/2511.15080}.

\bibitem{openai2025}
A. Chatterji, T. Cunningham, D. J. Deming, Z. Hitzig, C. Ong, C. Y. Shan, and K. Wadman.
How people use chatgpt.
\textit{NBER Working Paper 34255}, 2025.
URL \url{https://cdn.openai.com/pdf/a253471f-8260-40c6-a2cc-aa93fe9f142e/economic-research-chatgpt-usage-paper.pdf}.

\bibitem{zhao2024}
W. Zhao, X. Ren, J. Hessel, C. Cardie, Y. Choi, and Y. Deng.
WildChat: 1M ChatGPT interaction logs in the wild.
\textit{arXiv preprint arXiv:2405.01470}, 2024.
URL \url{https://arxiv.org/abs/2405.01470}.

\bibitem{openai2024o1}
OpenAI.
OpenAI o1 system card.
\textit{arXiv preprint arXiv:2412.16720}, 2024.
URL \url{https://arxiv.org/abs/2412.16720}.

\bibitem{chiang2024}
W. L. Chiang, L. Zheng, Y. Sheng, A. N. Angelopoulos, T. Li, D. Li, H. Zhang, B. Zhu, M. Jordan, J. Gonzalez, and I. Stoica.
Chatbot Arena: An open platform for evaluating LLMs by human preference.
\textit{arXiv preprint arXiv:2403.04132}, 2024.
URL \url{https://arxiv.org/abs/2403.04132}.

\bibitem{wei2022}
J. Wei, X. Wang, D. Schuurmans, M. Bosma, E. H. Chi, F. Xia, Q. Le, and D. Zhou.
Chain-of-thought prompting elicits reasoning in large language models.
\textit{Advances in Neural Information Processing Systems}, 35:24824–24837, 2022.
URL \url{https://proceedings.neurips.cc/paper_files/paper/2022/hash/9d5609613524ecf4f15af0f7b31abca4-Abstract-Conference.html}.

\bibitem{yao2023}
S. Yao, J. Zhao, D. Yu, N. Du, I. Shafran, K. Narasimhan, and Y. Cao.
ReAct: Synergizing reasoning and acting in language models.
\textit{International Conference on Learning Representations (ICLR)}, 2023.
URL \url{https://arxiv.org/abs/2210.03629}.

\bibitem{dubey2024}
A. Grattafiori, A. Dubey, A. Jauhri, A. Pandey, A. Kadian, A. Al-Dahle, A. Letman, A. Mathur, A. Schelten, A. Yang, A. Fan, et al.
The Llama 3 Herd of Models.
\textit{arXiv preprint arXiv:2407.21783}, 2024.
URL \url{https://arxiv.org/abs/2407.21783}.

\bibitem{deepseek2024}
DeepSeek-AI, A. Liu, B. Feng, B. Xue, B. Wang, B. Wu, C. Lu, C. Zhao, C. Deng, C. Zhang, et al.
DeepSeek-V3 technical report.
\textit{arXiv preprint arXiv:2412.19437}, 2024.
URL \url{https://arxiv.org/abs/2412.19437}.

\end{thebibliography}
\end{document}